\documentclass{article}

\usepackage{arxiv}

\usepackage[utf8]{inputenc} 
\usepackage[T1]{fontenc}    
\usepackage{hyperref}       
\usepackage{url}            
\usepackage{booktabs}       
\usepackage{amsfonts}       
\usepackage{nicefrac}       
\usepackage{microtype}      
\usepackage{lipsum}
\usepackage{graphicx}
\usepackage{amsmath}
\usepackage{subfigure}
\usepackage{multirow}
\usepackage{algorithm}
\usepackage{algorithmic}
\DeclareMathOperator{\sgn}{sgn}

\title{Dynamic Mode Decomposition in Adaptive Mesh Refinement and Coarsening Simulations}

\author{
    Gabriel F. Barros \\
  Dept. of Civil Engineering\\
  COPPE/Federal University of Rio de Janeiro \\
  P.O. Box 68506, RJ 21945-970, Rio de Janeiro, Brazil \\
  \texttt{gabriel.barros@coc.ufrj.br} \\
     \And
  Malú Grave \\
  Dept. of Civil Engineering\\
  COPPE/Federal University of Rio de Janeiro \\
  P.O. Box 68506, RJ 21945-970, Rio de Janeiro, Brazil \\
  \texttt{malugrave@nacad.ufrj.br} \\
   \And
   Alex Viguerie \\
  Department of Mathematics\\
  Gran Sasso Science Institute\\
  Viale Francesco Crispi 7, L'Aquila, AQ 67100, Italy \\
  \texttt{alexander.viguerie@gssi.it} \\
  \And
 Alessandro Reali \\
  Dipartimento di Ingegneria Civile ed Architettura\\
  Università di Pavia \\
  Via Ferrata 3, Pavia, PV 27100, Italy \\
  \texttt{alereali@unipv.it} \\
   \And
 Alvaro L.G.A. Coutinho \\
  Dept. of Civil Engineering\\
  COPPE/Federal University of Rio de Janeiro \\
  P.O. Box 68506, RJ 21945-970, Rio de Janeiro, Brazil \\
  \texttt{alvaro@nacad.ufrj.br} \\
}

\begin{document}

\maketitle

\begin{abstract}
Dynamic Mode Decomposition (DMD) is a powerful data-driven method used to extract spatio-temporal coherent structures that dictate a given dynamical system. The method consists of stacking collected temporal snapshots into a matrix and mapping the nonlinear dynamics using a linear operator. The standard procedure considers that snapshots possess the same dimensionality for all the observable data. However, this often does not occur in numerical simulations with adaptive mesh refinement/coarsening schemes (AMR/C). This paper proposes a strategy to enable DMD to extract features from observations with different mesh topologies and dimensions, such as those found in AMR/C simulations. For this purpose, the adaptive snapshots are projected onto the same reference function space, enabling the use of snapshot-based methods such as DMD. The present strategy is applied to challenging AMR/C simulations: a continuous diffusion-reaction epidemiological model for COVID-19, a density-driven gravity current simulation, and a bubble rising problem. We also evaluate the DMD efficiency to reconstruct the dynamics and some relevant quantities of interest. In particular, for the SEIRD model and the bubble rising problem, we evaluate DMD's ability to extrapolate in time (short-time future estimates).
\end{abstract}

\section{Introduction}
\par 
Data-driven methods are currently revolutionizing the modeling, prediction, and control of complex systems. Increasingly, researchers are considering data-driven approaches for a diverse range of complex systems, such as turbulent flows, climate sciences, epidemiology, finance, robotics, and many other different applications \cite{Brunton2019book}. Even with the availability of better hardware and advances in techniques and algorithms, numerical simulations of these systems are still resource-demanding: strong nonlinearities, multiple scales, and large dimensionalities are typical examples of complexities found in modern applications. With the assembly of modern mathematical methods, unprecedented data availability, and increasing computational resources, previously complex, challenging problems can now be tackled within the new research field entitled scientific machine learning (SciML).
\par 
  SciML is a core component of artificial intelligence and computational technology that can be trained, with scientific data, to augment or automate human skills \cite{osti_1478744}. This emerging research area aims at the opportunities and challenges in the context of complex applications across science and engineering, and other interdisciplinary fields. A wide range of SciML methods can be categorized regarding the type of information available, and their intended use \cite{Brunton2020}. In this study, we focus on Dynamic Mode Decomposition (DMD), an unsupervised SciML method that can extract the most dynamically relevant low-rank structures from large-dimensional data observed in dynamical systems. DMD can be applied to both numerical \cite{Schmid2010} and experimental data \cite{Schmid2011}. 
\par 
The standard DMD procedure for numerical simulations consists of stacking the $m$ snapshots (discrete solutions in space for a given time step) in columns to create a matrix and map the dynamics using a linear operator. The DMD procedure assumes that the $m$ collected snapshots have spatial dimensionality $n$, where $n \gg m$, such that the snapshots matrix has dimension $n \times m$. This occurs when snapshots are obtained from experimental data with sensors in fixed positions or numerical simulations considering a fixed mesh. However, in many situations, this is not always achievable. For numerical simulations using Adaptive Mesh Refinement/Coarsening (AMR/C), for instance, spatial adaptivity leads to solutions computed in meshes that constantly change in time. Adaptive meshes lead to a different number of nodes, nodal coordinates and numbering, and mesh topologies. In the present study, we develop a strategy to project all the snapshots of a given simulation with different dimensionality onto a reference target mesh with minor accuracy loss, enabling the use of any SVD-based data-driven technique such as DMD. 
\par 
This paper is structured as follows: Section \ref{sec:fem_dmd} describes the relation between the discretization of PDEs in space and time and a dynamical system. This section introduces Dynamic Mode Decomposition, our method of choice for short-time future estimates and extrapolation. Section \ref{sec:methodology} describes our strategy to deal with simulations that consider AMR/C in their evolution (e.g., in the case of dimensionality of the output vector, as well as of mesh topology and/or node numbering, varying in time). In Section \ref{sec:numerical} we describe the numerical applications in this study: the use of DMD on a continuous SEIRD model for COVID-19 and two fluid dynamics problems, a density-driven gravity current, and a bubble rising problem. We show efficiency and accuracy results for the signal reconstruction. Moreover, future time step predictions using DMD are evaluated for the SEIRD model and the bubble rising problem. In Section \ref{sec:conclusions}, we draw our final remarks and conclusions.


    
    
\section{Numerical Methods and Dynamic Mode Decomposition}
\label{sec:fem_dmd}
\par 
Solving partial differential equations (PDEs) using fast, accurate, reliable, and robust methods is crucial for many industrial and scientific applications. Several methods (such as finite elements, finite differences, and many others) are responsible for approximating the infinite-dimensional PDEs into finite-dimensional spaces. The discretization of these equations allows the process to be automated. In the present study, we focus on using the finite element method for spatial discretization of the PDEs. That is, consider a generic transient parametric PDE such as

\begin{equation}
    \dfrac{\partial \mathbf{u}}{\partial t} + \mathcal{N}(\mathbf{u}; \mathbf{\zeta}) = \mathbf{f}, \text{ \hspace{2cm} in $\Omega \times (0, T]$,}
    \label{eq:strong}
\end{equation}
where $\mathcal{N}$ is a nonlinear operator, $\mathbf{\zeta}$ is a vector of parameters (e.g., diffusion, density, viscosity, etc.), $\mathbf{f}=\mathbf{f}(t;\zeta)$ is a given function and the solution $\mathbf{u}$ is a function of spatial coordinates $\mathbf{x}$, temporal coordinates $t$, and parameters $\zeta$ such that $\mathbf{u} = \mathbf{u}(\mathbf{x},t;\zeta)$. The equation is equipped with boundary and initial conditions
\begin{equation}
    \begin{aligned}
        \mathbf{u} &= \mathbf{g} \text{ \hspace{4cm} on $\Gamma_D \times (0, T]$,} \\
        \nabla \mathbf{u} \cdot \mathbf{n} &= \mathbf{h} \text{ \hspace{4cm} on $\Gamma_N \times (0, T]$,} \\
        \mathbf{u}(\mathbf{x},0;\zeta) &= \mathbf{u}_0(\mathbf{x};\zeta) \text{ \hspace{3.85cm} on $\tilde{\Omega}$},
    \end{aligned}
\end{equation}
where $\mathbf{g}$ and $\mathbf{h}$ are the Dirichlet and Neumann boundary conditions, respectively, $T$ is the final time, and $\mathbf{u}_0$ is the initial condition for $\mathbf{u}$. The domain $\Omega \subset \mathbb{R}^{nsd}$ is bounded by Lipschitz continuous boundaries $\Gamma_N \cup \Gamma_D = \Gamma \subset \mathbb{R}^{nsd - 1}$, and $\mathbf{n}$ is the unit outward normal to $\Gamma_N$. The union of boundaries and domain is represented as $\tilde{\Omega}$. The standard finite element method consists of discretizing  $\tilde{\Omega}$ into a mesh composed of nodes and elements. Each element has its domain $\Omega^e \subset \mathbb{R}^{nsd}$ and boundary $\Gamma^e \subset \mathbb{R}^{nsd-1}$. The weak form of the system can be obtained by integrating Eq. (\ref{eq:strong}) in its strong form against a weighting function $\mathbf{w} \in H^1(\Omega)$, where $H^1(\Omega)$ is the Sobolev space of the square-integrable functions with an integrable first weak derivative, and applying the divergence theorem. Being $P^k(\Omega^e)$ the space of polynomials of degree equal or less than $k$ over $\Omega^e$, the function spaces are defined as: 
\begin{flalign}
    \label{eq:spaces}
    S_t^h &= \{\mathbf{u}^h(\cdot,t) \in H^1(\Omega) ~|~ \mathbf{u}^h(\cdot,t)|_{\Omega_e}\in P^k(\Omega^e), \forall e\}, \\
    W^h &= \{\mathbf{w}^h\in H^1(\Omega) ~|~ \mathbf{w}^h|_{\Omega_e}\in P^k(\Omega^e), \forall e\}.
\end{flalign}
Therefore, the semi-discrete finite element formulation for Eq. (\ref{eq:strong})
is: find $\mathbf{u}^h(t) \in S_t$ such that $\forall \mathbf{w}^h \in W^h$:
\begin{equation}
    \bigg(\dfrac{\partial \mathbf{u}^h}{\partial t}, \mathbf{w}^h\bigg) + (\mathcal{N}(\mathbf{u}^h; \mathbf{\zeta}), \mathbf{w}^h) = (\mathbf{f}^h, \mathbf{w}^h),
    \label{eq:weak}
\end{equation}
where the $\mathcal{L}^2$ inner product over the domain $\Omega$ is indicated by $(\cdot, \cdot)$. The weak form given by Eq. \eqref{eq:weak} naturally accommodates several finite element formulations, from Galerkin to Variational Multiscale methods \cite{hughes,rasthofer,ahmed2017,codina2018,bazilevs2013computational}. After the temporal discretization of Eq. \eqref{eq:weak}, the equation can be translated into a discrete-time dynamic system. In this system, the state vector $\mathbf{u}^h$ at the time instant $k+1$ can be written such that:
\begin{equation}
    \mathbf{u}^h_{k+1} = \mathcal{F}(\mathbf{u}^h_k),
    \label{eq:dynamical}
\end{equation}
where $\mathcal{F}$ represents the discrete-time flow map of the system and incorporates information regarding the parameters $\zeta$, mesh size, solver tolerances, etc. In the present study, we consider that the measurements of the system are the state vectors themselves, that is, $\mathbf{u}^h_k$. Analyzing the evolution in time of a discretized PDE as a dynamical system is a key concept for introducing Dynamic Mode Decomposition (DMD).

DMD is an equation-free, data-driven method that provides accurate assessments of the dominant structures in a given complex system \cite{Kutz2016book}. Differently from the Proper Orthogonal Decomposition (POD) \cite{Lumley1967, Berkooz1993} that represents data in terms of spatial modes, DMD provides a decomposition of data into spatio-temporal modes that correlate the data across spatial features and also associates them to unique temporal Fourier modes. The main idea of DMD is to efficiently compute the regression of linear/nonlinear terms to a least-square linear dynamics approximation from experimental or numerical observable data. Despite its first appearance in the fluid dynamics context \cite{Rowley2009,Schmid2010}, DMD has been used in many other applications such as epidemiology \cite{Proctor2015}, biomechanics \cite{Calmet2020}, urban mobility \cite{Alla2020}, climate \cite{Kutz2016} and aeroelasticity \cite{Fonzi2020}, especially in structure extraction from data and control-oriented methods. 

\par 
We can now apply DMD on the dynamical system described in Eq. (\ref{eq:dynamical}). Consider a dataset $\mathbf{Y}^h$ containing the observations in time of the dynamical system $\mathbf{u}^h_k$ for $k = 0, 1, \dotsc, m$, where $m+1$ is the total number of observations. The dataset
\begin{equation}
                \mathbf{Y}^h = \left[
                \begin{array}{cccc}
                \vrule & \vrule &        & \vrule \\
                \mathbf{u}^h_0  & \mathbf{u}^h_1  & \ldots & \mathbf{u}^h_m    \\
                \vrule & \vrule &        & \vrule
                \end{array}
                \right]
\end{equation}
can be split into two datasets $\mathbf{Y}^h_1 = [\mathbf{u}^h_0 \dotsc \mathbf{u}^h_{m-1}]  \in \mathbb{R}^{n \times m}$ and $\mathbf{Y}^h_2 = [\mathbf{u}^h_1 \dotsc \mathbf{u}^h_m]  \in \mathbb{R}^{n \times m}$. 
DMD consists on finding the best fit approximation of the linear mapping $\mathbf{A}$ that transforms dataset $\mathbf{Y}^h_1$ into dataset $\mathbf{Y}^h_2$, that is, 
\begin{equation}
    \mathbf{Y}^h_2 = \mathbf{A}\mathbf{Y}^h_1.
\end{equation} 
The computation of $\mathbf{A}$ can be done as $\mathbf{A} = \mathbf{Y}^h_2\mathbf{Y}^{h\dagger}_1$, where $\mathbf{Y}_1^{h\dagger}$ is the Moore-Penrose pseudoinverse of $\mathbf{Y}^h_1$. However, we avoid the computation of the full matrix $\mathbf{A}$ since $\mathbf{A}$ is a $n \times n$ matrix. Also, the computation of the full Moore-Penrose pseudoinverse is not advisable due to its ill-conditioning. Instead, we can compute the SVD of $\mathbf{Y}^h_1$ as 
\begin{equation}
\label{eq:SVD}
    \mathbf{Y}_1^h = \mathbf{U\Sigma V}^T,
\end{equation}
where  $\mathbf{U} \in \mathbb{R}^{n \times m}$ and $\mathbf{V} \in \mathbb{R}^{m \times m}$ are the left and right singular vectors and $\mathbf{\Sigma} \in \mathbb{R}^{m \times m}$ is a diagonal matrix with real, non-negative, and decreasing entries named singular values. The singular values $\sigma_0 \geq \sigma_1 \geq \sigma_2 \geq \dots \geq \sigma_{m-1}$ are hierarchical and can be interpreted in terms of how much the singular vectors influence the original matrix $\mathbf{Y}^h_1$. For the DMD procedure, considering the Eckart-Young Theorem \cite{Eckart1936}, the optimal low-rank update approximation matrix $\mathbf{Y}_1^h$, when subjected to a truncation rank $r$, can be written as    
\begin{equation}
    \mathbf{Y}_1^h \approx \mathbf{\tilde{Y}}^h_1 = \mathbf{U}_r\mathbf{\Sigma}_r\mathbf{V}^T_r,
\end{equation}
where $\mathbf{U}_r \in \mathbb{R}^{n \times r}$ is a matrix containing the first $r$ columns of $\mathbf{U}$, $\mathbf{V}_r \in \mathbb{R}^{m \times r}$ contains the first $r$ columns of $\mathbf{V}$, and $\mathbf{\Sigma}_r \in \mathbb{R}^{r \times r}$ is the diagonal matrix containing the first $r$ singular values. The pseudoinverse can be approximated as
\begin{equation}
    \mathbf{Y}^{h\dagger}_1 \approx \mathbf{\tilde{Y}}^{h\dagger}_1= \mathbf{V}_r\mathbf{\Sigma}_r^{-1}\mathbf{U}_r^T
\end{equation}
and, instead of computing $\mathbf{A}\in \mathbb{R}^{n \times n}$, we can obtain $\mathbf{\tilde{A}}$, a $r \times r$ projection of $\mathbf{A}$ as,
\begin{equation}
    \mathbf{\tilde{A}} = \mathbf{U}_r^T\mathbf{A}\mathbf{U}_r =  \mathbf{U}_r^T\mathbf{Y}^h_2\mathbf{V}_r\mathbf{\Sigma}^{-1}_r.
\end{equation}
\par 
Note that $\mathbf{\tilde{A}}$ is unitarily similar to $\mathbf{A}$. Further mathematical details regarding the optimization problem (the best-fitting matrix $\mathbf{\tilde{A}}$) and the influence of the Eckart-Young Theorem on constraints of the problem can be found in \cite{heas2020lowrank}. Now we can compute the eigendecomposition of $\mathbf{\tilde{A}}$:
\begin{equation}
    \mathbf{\tilde{A}W = W\Lambda},
\end{equation}
where $\mathbf{\Lambda}$ is a diagonal matrix containing the discrete eigenvalues $\lambda_j$ and the matrix $\mathbf{W}$ contains the eigenvectors $\mathbf{\phi}_j$ of $\mathbf{\tilde{A}}$. The DMD basis can be written as:
\begin{center}
	\begin{equation}
	\mathbf{\Psi} = \mathbf{Y}^h_2\mathbf{V}_r\mathbf{\Sigma}^{-1}_r\mathbf{W},
	\end{equation}
\end{center}
and the signal reconstruction as:

\begin{equation}\label{DMDExpansion}
    \mathbf{u}^h(t)  \approx  \tilde{\mathbf{u}}^h(t) =  \mathbf{\Psi}\exp(\mathbf{\Omega}_{eig}t) \mathbf{b},
\end{equation}


being $\mathbf{b}$ the vector containing the projected initial conditions such that $\mathbf{b} = \mathbf{\Psi^{\dagger}}\mathbf{u}^h_0$, and $\mathbf{\Omega}_{eig}$ is a diagonal matrix whose entries are the continuous
eigenvalues $\omega_i = \ln(\lambda_i)/\Delta t_o$, where $\Delta t_o$ is the time step size between the outputs. In the present study $\Delta t_o = j \Delta t$, where $j = 1, 2, \dots, m$, being $\Delta t$ the time step size used in the temporal integration of the PDEs. For instance, if one chooses to output the solutions once every two time steps, the time step size between the two observations will be two times larger than the time step size used to compute time integration, that is,  $\Delta t_o = 2 \Delta t$. It is important to mention that the snapshots sampling frequency affects directly the DMD's ability to capture the dynamics. For lower dominant frequencies, a larger $\Delta t_o$ is more adequate, while smaller values of $\Delta t_o$ are required for capturing rapid dynamics \cite{Schmid2010}. The form of \eqref{DMDExpansion} can be regarded as a generalization of the Sturm-Liouville expansion for a differential problem:
\begin{align}\label{SturmLiouville}
    u(t) &= \sum_{i=0}^{\infty} b_i \psi_i e^{\omega_i t},
\end{align}
where $\psi_i$ and $\omega_i$ are the $i$-th Sturm-Liouville eigenfunctions and eigenvalues for a given differential operator.
\par 
DMD can be seen as a dimensionality reduction method due to its inherent ability to extract the $r$ most relevant dynamical modes, where $r$ is often much smaller than the snapshot matrix rank $m$. However, a strategy to determine the number of relevant modes is not straightforward, and is an active topic in DMD research \cite{Taira2017}. Even though the choice of $r$ for DMD may require some trial and error, some techniques can be used to find a good starting point. A hard threshold technique \cite{Kutz2016book} consists of choosing $r$ such that:
\begin{equation}
	\label{eq:relative_energy}
	\kappa = 1 - \dfrac{\sum_{i=1}^{r}\sigma_{i}^{2}}{\sum_{i=1}^{m}\sigma_{i}^{2}} \leq \tau,
\end{equation}    
where $\tau$ is a tolerance threshold, set, e.g., to $10^{-6}$. This method implies that more than $100(1-\tau)\%$ of the variance in the data is retained by the approximation. For the case where DMD is used on experimental (or numerical but noisy) data, more sophisticated solutions are presented in the literature \cite{Gavish2014,Donoho1995}.

\par Another important consideration when using DMD is the SVD algorithm. The SVD can represent a significant part of the computational effort, meaning that improvements in the SVD performance lead to significant CPU time gains. For many SVD-based methods (such as DMD), there is no need to compute the SVD for the whole matrix, since the method aims at extracting the first $r$ dominant structures in the matrix. Many algorithms are designed in this direction to make this computation more efficient. One important contribution in this direction is seen in \cite{Sirovich1987}, where the method of snapshots was proposed, paving the way to more algorithms and techniques. In this paper, we employ the randomized SVD (rSVD) algorithm \cite{Halko2011, Erichson2019}, a non-deterministic algorithm able to compute the near-optimal low-rank approximation of a given large dataset with good efficiency.


\section{DMD on Adapted Meshes}
\label{sec:methodology}
Regarding numerical methods to approximate PDEs, the use of finer meshes in finite element discretizations usually leads to more accurate solutions. On the other hand, reducing the number of equations $n_{eq}$ in the nonlinear system is crucial for efficiency, especially considering that the optimal computational complexity of a single physics transient finite element simulation is $\mathcal{O}(n_{eq}^{\frac{4}{3}})$ \cite{Burstedde2010}. The duality between the two statements describes a well-known trade-off between accuracy and efficiency in the finite element context. Despite a considerable research effort in the past decades, strategies to generate tailored meshes to maximize the accuracy while minimizing the computational effort are still an open research topic. Milestones addressing this subject are finite element \textit{a posteriori} error estimators/indicators  \cite{ainsworth2011posteriori}, techniques such as adaptation, interpolation \cite{carey1997computational, Lohner1995} and projection \cite{Carey2001, Farrell2011, CodinaIJNME2017}, as well as libraries and frameworks containing automated versions of AMR/C techniques \cite{libmesh, AlnaesBlechta2015a}.  
\par
\begin{figure}
    \centering
    \includegraphics[width=0.7\linewidth]{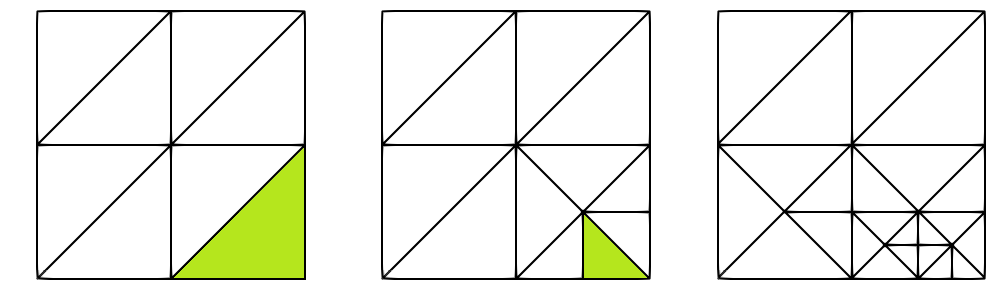}
    \caption{Illustration of a mesh refinement procedure. A local \textit{a posteriori} error estimator or indicator flags an element for refinement (in green) using the solution computed in the mesh on the left. The mesh is refined (or coarsened) according to the flagged elements, and the process can be restarted until a given criterion is met (error level, element size, maximum number of elements, etc.). Note that the initial and the final mesh differ in the number of degrees of freedom and topology.}
    \label{fig:mesh_ref}
\end{figure}

 
The general structure of the AMR/C scheme is given in Algorithm 1 and illustrated in Figure \ref{fig:mesh_ref}. Three criteria are fundamental in an AMR/C algorithm: remeshing, flagging, and stopping. The remeshing criterion defines whether the computed solution at a given time step requires remeshing driven by global \textit{a posteriori} error estimators (or indicators) and/or by calling the refinement/coarsening procedure at every $j$ time steps. Next, all mesh elements are visited and flagged for refinement or coarsening. The element flagging criterion is often represented by local \textit{a posteriori} error estimators or indicators, i.e. flux-jumps of the solution gradient. In possession of the flagged elements, the remeshing algorithm is invoked. Finally, the previous mesh solution must be projected or interpolated into the target mesh. This is done by the projection/interpolation algorithm. The whole process is repeated until the stopping criterion is achieved. This criterion could be error equidistribution (until a certain threshold), a given element size, the maximum number of elements in the mesh, or the number of refinement levels. Note that the meshes generated in Algorithm 1, guided by the error estimation procedure, have different dimensions (number of degrees of freedom) and different topologies, which poses difficulties for DMD (or, in fact, for any snapshot-based method).
\begin{algorithm}
	\caption{Adaptive Mesh Refinement/Coarsening Algorithm}
	\begin{algorithmic}
	    \STATE INPUT: Finite element solution computed at a given time instant $t_i$.
	    \STATE OUTPUT: Adapted finite element solution.
	    \vspace{0.1cm}
		\STATE - Definition of a remeshing criterion: global \textit{a posteriori} error estimator/indicator, remeshing at every $j$ time steps, etc.
		\FOR{each computed solution} 
            \IF{one or more remeshing criteria are met}
                \WHILE{one or more stopping criteria are not met}
                \STATE - Compute local error estimators/indicators (or another strategy) to identify the elements requiring refinement or coarsening and flag them.
                \STATE - Call the refinement/coarsening procedure to generate a target mesh for the flagged elements. Different strategies for mesh refinement/coarsening lead to distinct meshes.
                \STATE - Project or interpolate the solution computed on instant $t_i$ onto the target mesh
                \ENDWHILE
            \ENDIF
        \ENDFOR
		\end{algorithmic}
	\label{algdmd}
\end{algorithm}

\par 
The structure of the exact DMD algorithm relies on the fact that the measurements of the state vectors, i.e., $\mathbf{u}^h_0 \dotsc \mathbf{u}^h_m$ have the same dimensionality. This structure could be exemplified in numerical experiments as fixed discretizations in space, i.e., fixed meshes or static sensors in experimental data. However, finite element simulations equipped with AMR/C strategies provide solutions in different function spaces, depending on the mesh used to compute the solution on a given time step. Spatial adaptivity on transient finite element simulations leads to meshes with a different number of nodes, numbering, and nodal coordinates. It can also lead to different mesh topologies and structures. For that reasons, snapshots obtained by AMR/C simulations cannot be stacked in columns to construct the snapshot matrix. Even if one considers an AMR/C strategy that restricts the adaptive meshes to preserve the dimensionality of the snapshots, the difference between the nodal coordinates of the various meshes will lead to misleading dynamics captured by DMD. In the present study, we circumvent this issue by considering the $\mathcal{L}^2-$projection \cite{CodinaIJNME2017} of the numerical simulation results for different meshes into a reference target mesh. Recent work on reducing these projection costs for related reduced-order modeling techniques (though not DMD) may be found in \cite{HALE2021113723}. The projection or interpolation of numerical solutions between finite element meshes is a well-known computational mechanics subject. Many issues regarding boundary conditions, data visualization, or coupling arise from this kind of problem. The choice of a proper method to successfully project functions in different finite element spaces is not a trivial task since conservation may not be satisfied \cite{Carey2001, Farrell2011, CodinaIJNME2017}. This issue is not addressed in the present study since the mesh projection occurs as a post-processing phase after computing the AMR/C solutions. Therefore it cannot lead to cumulative errors. Furthermore, we are also careful to choose reference target meshes with characteristic length equivalent with the existent in the AMR/C meshes to avoid any major accuracy losses. For the sake of generality, we consider the $\mathcal{L}^2-$projection as our method of choice, and it can be defined as follows \cite{Thompson2015}. Assuming that the solution on the donor mesh $\mathbf{u}^h(\mathbf{x}) = \sum_{j=1}^n \mathbf{u}_j\phi_j(\mathbf{x})$  to be projected onto the target mesh must satisfy the orthogonality condition, 
\begin{equation}
    \begin{centering}
        (\mathbf{u}^h - \mathbf{u}^{proj},\mathbf{v})_{\mathcal{L}^2} = 0 \hspace{4cm} \forall \mathbf{v} \in V^{\psi}
        \label{eq:l2_norm_1}
    \end{centering}
\end{equation}
where $V^{\psi}$ is a finite-dimensional subspace of $\mathcal{L}^2(\Omega)$ defined by the target mesh and the interpolant $\mathbf{u}_{proj}$ is the optimal interpolant in the $\mathcal{L}^2-$norm for $V^{\psi}$. The orthogonal projection can be defined in terms of the following linear system
\begin{equation}
    \mathbf{M}\mathbf{u}^{proj} = \mathbf{P}\mathbf{u}
\end{equation}
where $\mathbf{M}$ is the mass matrix and $\mathbf{P}$ is the projection matrix. The mass matrix is usual in finite element computations. The $\mathbf{P}$ matrix, however, can present technical difficulties (see Appendix). 
\par
\begin{figure}[!ht]
	\begin{center}
		\subfigure[Adaptive mesh (left) and solution (right),  $||\mathbf{u}^h||_\infty = 1.000$.]{\label{fig:adapt}\includegraphics[width=0.7\linewidth]{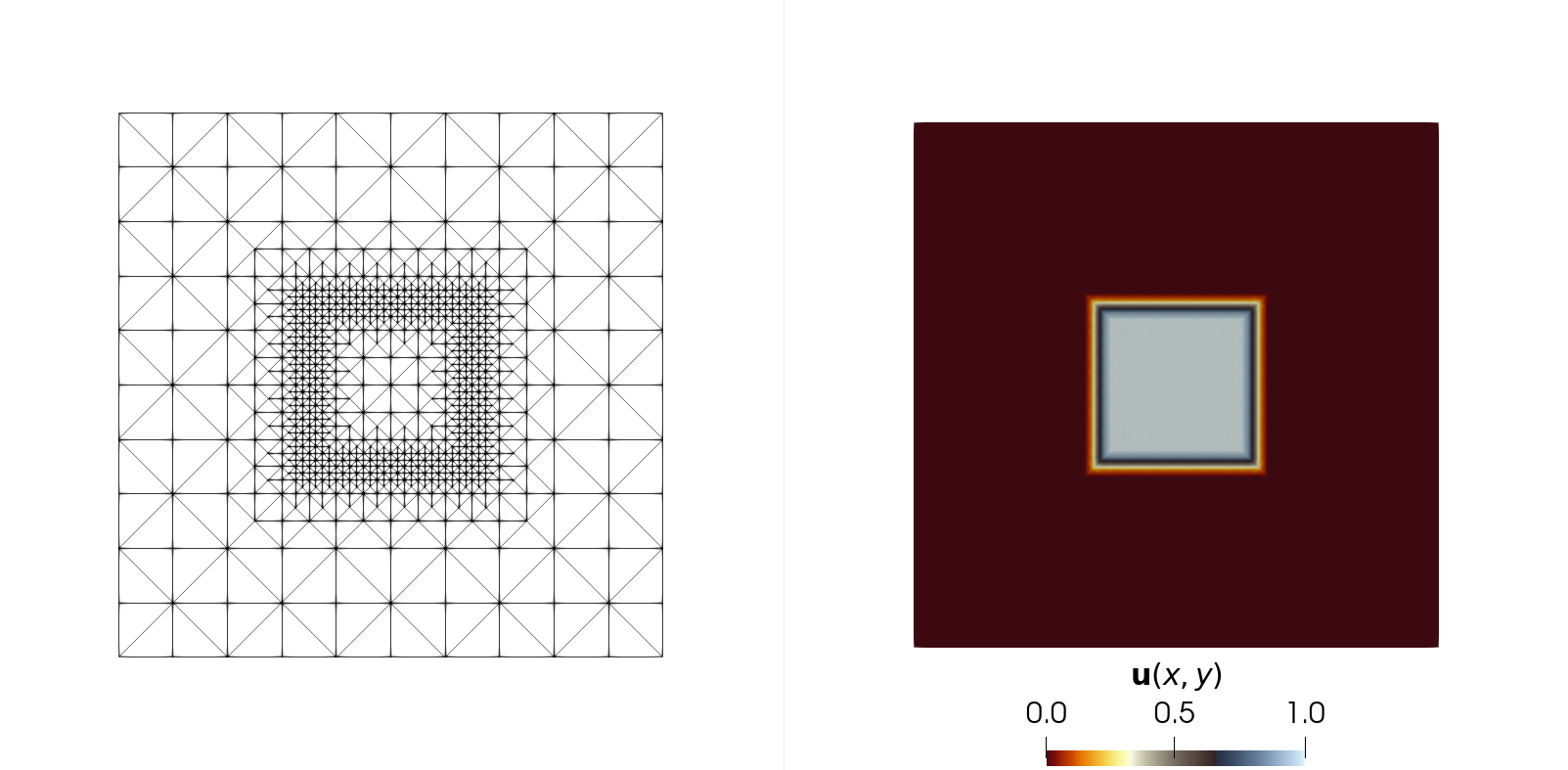}}
		\subfigure[Structured mesh (left) and projected solution (right),  $||\mathbf{u}^{proj}||_\infty = 1.000$.]{\label{fig:struct}\includegraphics[width=0.7\linewidth]{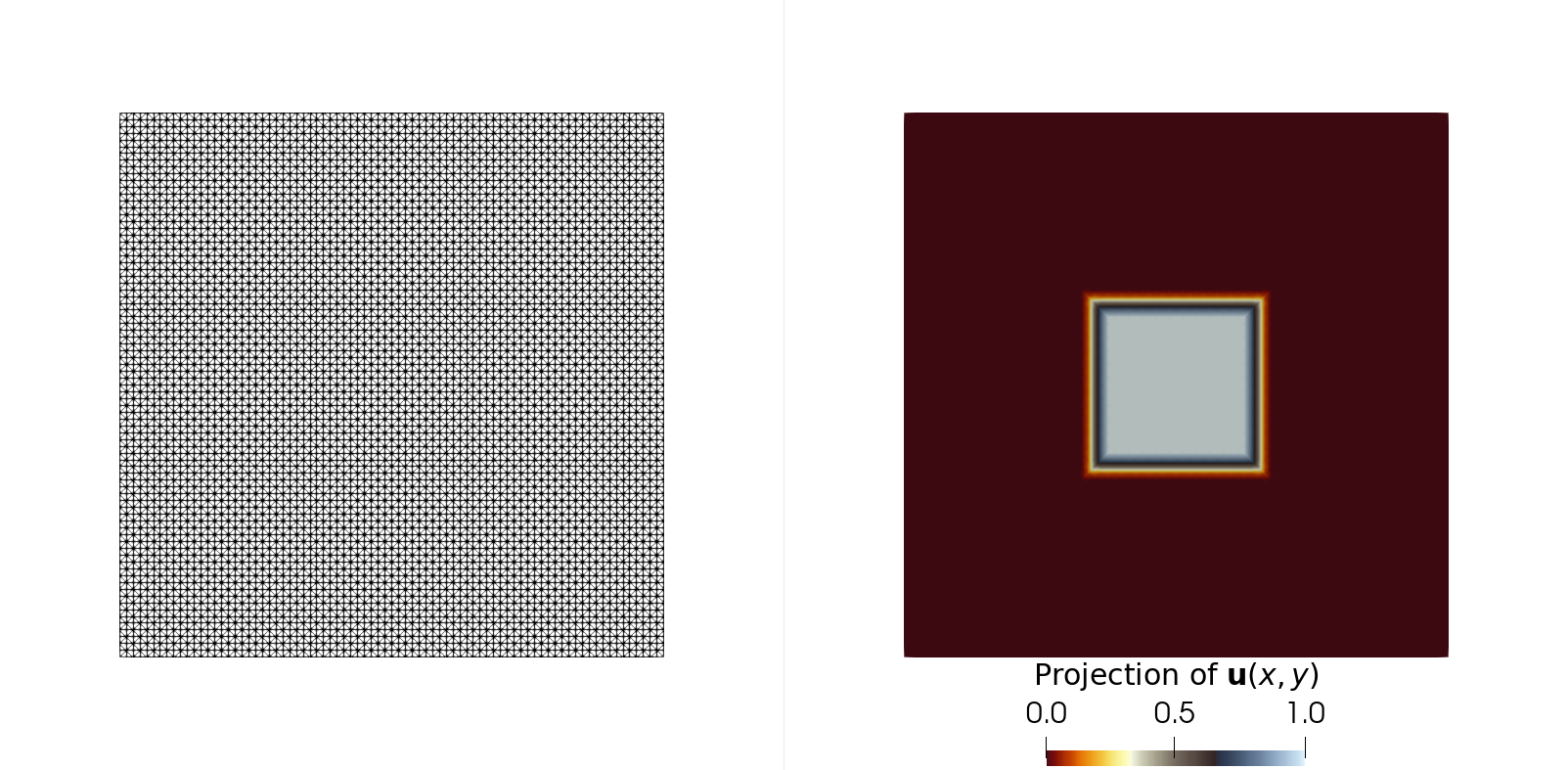}}
		\subfigure[Unstructured mesh (left) and projected solution (right),  $||\mathbf{u}^{proj}||_\infty = 0.999$.]{\label{fig:unstruct}\includegraphics[width=0.7\linewidth]{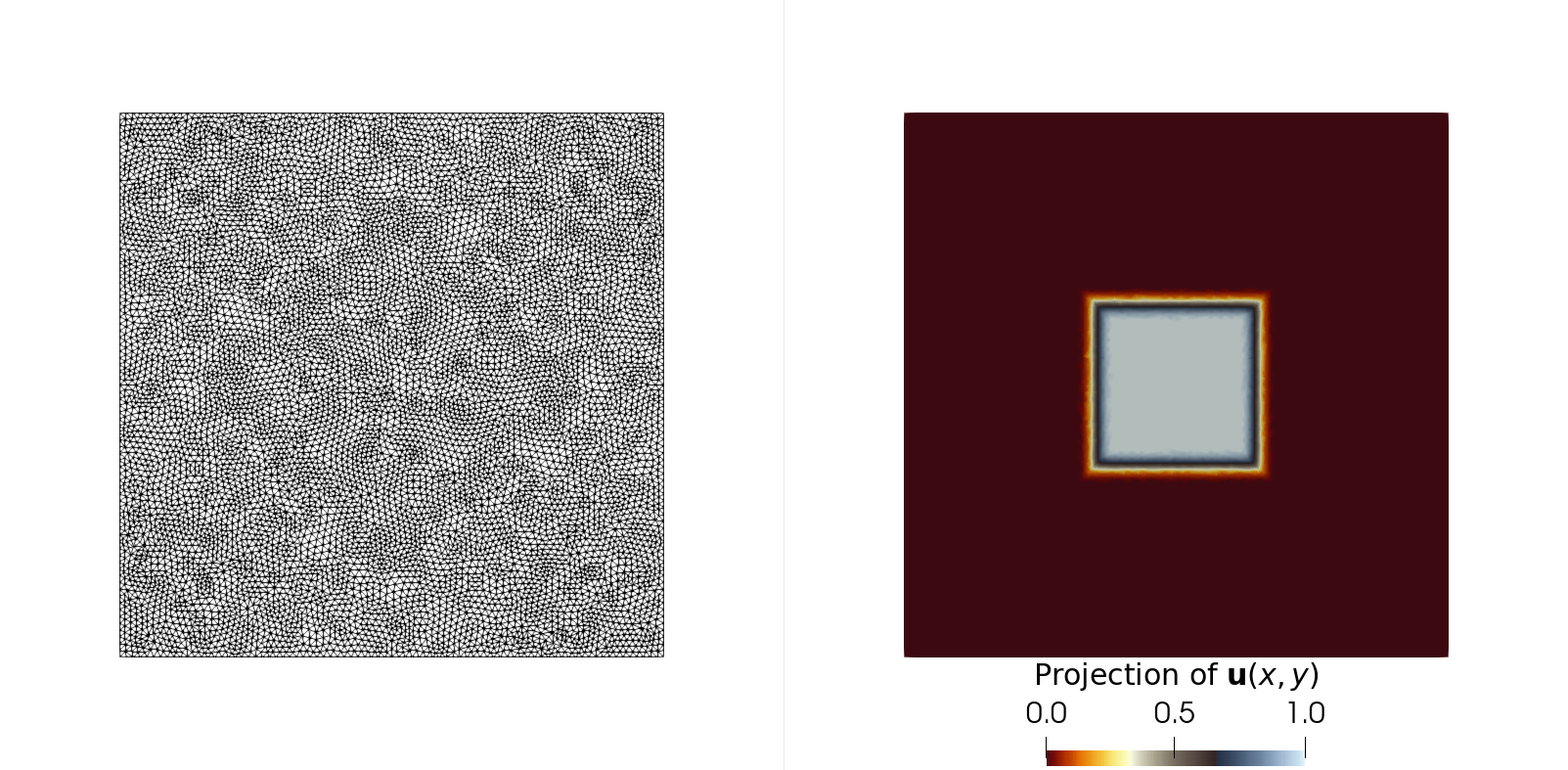}}
		\caption{Comparison of different $\mathcal{L}^2-$projection examples on structured and unstructured meshes.}
		\label{fig:l2projection}
	\end{center}
\end{figure}

 
Therefore, our strategy consists of applying the $\mathcal{L}^2-$projection onto a tailored reference target mesh capable of representing the many scales in time and space of all the snapshots. This routine is inserted in the code and invoked after the adaptive procedure in Algorithm 1 at every output time step $\Delta t_o$. 
This tailored reference mesh is described in this work as target reference mesh and should not be confused with the target mesh generated during the AMR/C procedure. The new solution (computed on the adaptive donor mesh) is projected onto the reference target mesh and exported as a simulation output. In this work, we export as output files for visualization purposes, although there is no restriction on stacking the snapshots on the snapshot matrix during the simulation runtime or dumping only $\mathbf{u}^{proj}$ on disk. We considered \texttt{Gmsh}\cite{Geuzaine2009}, an open-source robust mesh generator, as our software of choice for defining and creating the target meshes for this study. The output for each time step is the snapshot with constant dimensionality $n$ such that all nodes in space are correctly mapped and capturing the dynamics existent in the system. The procedure is summarized in Algorithm 2. This strategy is relatively simple since the $\mathcal{L}^2-$projection consists of solving a linear system where the generated matrix is a mass matrix and requires no extra outputs for storing the projected solutions since the projection can be applied right after the AMR/C code. The mass matrix is generated in the finite element context by a self-adjoint operator, enabling more efficient solvers. In terms of versatility, the $\mathcal{L}^2-$projection method is flexible because a solution obtained for a given mesh can be naturally projected onto reference target meshes with different topologies and dimensionalities. Also, since the mesh projection is a vital part of AMR/C algorithms, finite element algorithms frequently present efficient implementations of interpolation or projection techniques. Figure \ref{fig:l2projection} shows an example where a solution obtained by an adaptive mesh simulation is projected onto two meshes with different topologies.
For this example, we consider the square domain $\Omega = [-1,1]\times[-1,1]$ and the function $u$ 
defined in $\Omega$ such that
\begin{equation}
    u = \begin{cases} 1 \text{, if }(x,y)\in [-0.3,0.3]\times[-0.3,0.3], \\ 0 \text{, otherwise.} \end{cases}
\end{equation}
This function is approximated on a structured finite element mesh discretized into $10 \times 10$ cells where each cell is divided into two triangular elements. An AMR/C procedure is invoked to refine three times the transition between $\mathbf{u} = 1$ and $\mathbf{u} = 0$, creating a new mesh containing $1672$ elements and $857$ nodes. Figure \ref{fig:adapt} shows the new mesh generated after the AMR/C procedure and the function approximated by the resulting function space. Two meshes - one structured and another unstructured - are considered for projection. The structured mesh contains $80 \times 80$ cells, resulting in $12800$ triangular elements and $6561$ nodes. The element sizes of the structured mesh are similar to the smallest elements in the adaptive mesh. The unstructured mesh presents a smaller characteristic length than the structured mesh and contains $17088$ elements and $8705$ nodes. Figures \ref{fig:struct} and \ref{fig:unstruct} show the proposed meshes and the projections of the solution onto the new finite element spaces. Note that the projected solutions are fairly accurate since their infinity norm is in good agreement with the adaptive mesh solution's infinity norm.

\begin{algorithm}
	\caption{Snapshot projection}
	\begin{algorithmic}
	    \STATE INPUT: Finite element solutions considering AMR/C 
	    \STATE OUTPUT: Projected solutions onto a prescribed reference target mesh.
	    \vspace{0.1cm}
        \FOR{each observation time step $\Delta t_o$}
            \STATE 1: Apply the $\mathcal{L}^2-$projection of the adaptive solution onto the function space of the reference target mesh.
            \STATE 2: Stack the resulting snapshot into the snapshot matrix $\mathbf{Y}^h$ or output the projected solutions to disk.
        \ENDFOR
		\end{algorithmic}
	\label{algdmd}
\end{algorithm}

\textbf{\textit{Remark:}} In this study, the projection is carried out inside the finite element simulation since the projection computational cost is practically negligible in comparison with the overall time required for solving nonlinear systems of a complex numerical simulation. However, if one does not have access to the finite element simulation codes used, the reconstruction of the solution of each time step can still be done off-line. Output files of various formats contain information regarding the mesh used (such as nodal coordinates and connectivities) for visualization purposes. By properly reading these files, the solutions can be reconstructed under a finite element framework (such as \texttt{FEniCS} \cite{fenics} or \texttt{libMesh} \cite{libmesh}) or on a code developed by the user. However, this totally non-intrusive approach can significantly increase the computational cost due to several I/O operations that are often extremely low compared to computational intensive operations. Since the mesh constantly varies in an AMR/C finite element simulation, each solution obtained in the simulation has to be imported, reconstructed under its original mesh, and projected. After the projection, the user can choose to dump the solution in the disk or stack the snapshots on the snapshot matrix. For the first case, another I/O operation would be invoked. This non-intrusive strategy is especially suitable (and restricted) to be used with DMD, which is also a non-intrusive algorithm. For intrusive snapshot-based methods such as Proper Orthogonal Decomposition (POD), one needs to project finite element matrices onto the computed basis and, therefore, requires access to the code \cite{ullmann2016pod}.  


\section{Numerical Experiments}
\label{sec:numerical}
\par In this section, for the sake of generality, we apply our method in several applications, with different systems of equations, numerical formulations, mesh topologies, spatial dimensions, refinement criteria, and finite element libraries. We compare the results between the AMR/C solution, the fixed mesh solution, and the DMD results for all cases. First, we test the DMD short-time future prediction capabilities on a continuous SEIRD model for COVID-19, a nonlinear system of diffusion-reaction equations. The equations are considered using a Galerkin finite element discretization and are solved using \verb|libMesh| \cite{libmesh}, a high-performance C++ finite element library. The error estimators, refinement/coarsening strategies built-in on \verb|libMesh| can be seen on \cite{Viguerie2021, grave2020adaptive}. Also, the $\mathcal{L}^2$-projection algorithm is embedded in \texttt{libMesh}. We explore the results in one and two spatial dimensions, where the 1D case is a hypothetical example, and the 2D case describes the COVID-19 evolution in the Lombardy region in Italy \cite{viguerie2020diffusion, Viguerie2021, grave2020adaptive}. The simulation obtains the results for $44$ and $60$ days for the 1D and 2D cases, respectively. Since we want to predict $14$ days in the future, we only feed DMD with snapshot matrices containing the first $30$ and $46$ days in the 1D and 2D cases, respectively, such that the predicted results can be compared with the results obtained in the simulations. To apply DMD to the SEIRD data obtained on an AMR/C simulation, we project the adaptive solution onto a mesh with characteristic length compatible with the obtained simulation results. The discretization of the reference mesh used for the projection is as fine as the final refinement level on the adaptive meshes. Next, we consider two fluid dynamics applications where the governing equations for both cases are advection-dominated. To circumvent the LBB condition and spurious oscillations regarding dominant advection, these equations consider the residual-based variational multiscale (RBVMS) formulation \cite{hughes,rasthofer,ahmed2017,codina2018,bazilevs2013computational,Guerra2013} on a finite element discretization. We consider the use of DMD on a 2D density-driven gravity flow and a 3D bubble rising simulation. The density-driven gravity current is modeled by the coupling of the incompressible Navier-Stokes equation and the advection-diffusion equation. We consider a lock-exchange problem, where a tank is filled with two fluids of different densities, separated by a lock. The simulation starts when the lock is removed, and the difference in the densities of the fluids generates the driving forces responsible for the motion of the fluids. Unlike the other examples in this work, the implementation of this numerical test is made on the \texttt{FEniCS} v.2019.1 framework \cite{fenics}, a high-performance Python/C++ finite element library. The refinement/coarsening algorithm and $\mathcal{L}^2-$projection algorithm for this example are part of the framework. We consider an interface-tracking error indicator for the AMR/C simulations, and the mesh is refined following the bisection method \cite{Rivara1984}. DMD is considered to reconstruct the solution, and the results are compared to the fixed mesh results and the results obtained by AMR/C simulations. Finally, we extend our analysis to a 3D bubble rising case \cite{grave2020new}, a two-phase incompressible flow problem where the interface is captured by the convected level-set method. This model is implemented on \texttt{libMesh}, taking advantage of the same refinement/coarsening strategies as well as the projection algorithm used in the SEIRD numerical tests. In this example, we test the projection of the adaptive solutions onto three different meshes containing the different scales existent on the AMR/C simulation and evaluate the results. We consider DMD to predict the bubble geometry and dynamics for a short time in the future.  
\par 
For all the numerical tests proposed, we evaluate the results in terms of efficiency and accuracy. For efficiency purposes, we compute the ratio between the computational time required to run the finite element simulations and the time required to run DMD separately. We refer to this quotient as speedup. The finite element code is responsible for computing the snapshots and projecting the results onto the reference target meshes proposed, while the DMD code imports the output files, extract the snapshots, computes the approximation, and outputs the results. Also, we provide a table describing how the projection routine affects the overall computational time required for the simulations for all examples. In terms of accuracy, we evaluate the results in terms of overall relative error $\eta_F = \dfrac{||\mathbf{Y}^h - \mathbf{Y}^h_{DMD}||_F}{||\mathbf{Y}^h||_F}$ where $\mathbf{Y}^h$ is the snapshots matrix, $\mathbf{Y}^h_{DMD}$ is a matrix comprising the approximations obtained by DMD and $||\cdot||_F$ denotes the Frobenius norm. A more detailed analysis is also done in terms of relative error in time $\eta$. For that, we plot the curves of the relative errors (in terms of $\mathcal{L}^2$-norm) of each snapshot. We compute $\eta = \dfrac{||\mathbf{u_k}^h - \mathbf{u_k}^h_{DMD}||_2}{||\mathbf{u_k}^h||_2}$ for $k = 0, 1, \dots, m$ snapshots, where $||\cdot||_2$ denotes the $\mathcal{L}^2$-norm. Also, to avoid unphysical results, some quantities of interest are evaluated and compared using both simulation and approximation results. For the SEIRD model, we plot the results regarding the total population. This quantity should be constant in time according to the hypothesis of the model. For the lock-exchange simulation, we compute the mass during the simulation and the front position. Since the simulation is considered on a closed tank, the mass must be kept constant during the simulation. For the 3D bubble rising problem, we plot the quantities of interest related to the geometry (volume and sphericity) and dynamics (center of mass and rise velocity) of the bubble. In this case, we test meshes with different minimum characteristic lengths. The results are shown and discussed below.

\subsection{Continuous SEIRD model for COVID-19}

The outbreak of COVID-19 in 2020 has led to a surge in interest in the mathematical modeling of infectious diseases. This new virus is responsible for infecting millions of people worldwide and impacting the economy in an unprecedented way. Therefore, numerical simulation of the virus' dynamics may help provide short-term prediction models for forecasting the number of future cases. In this perspective, it is possible to develop strategic planning in the public health system to avoid deaths and manage patients.

Disease transmission may be modeled as \textit{compartmental models}, in which the population under study is divided into compartments and has assumptions about the nature and time rate of transfer from one compartment to another \cite{brauer2019mathematical}. Here, we work with a spatio-temporal SEIRD model, presented in \cite{viguerie2020diffusion, Viguerie2021, grave2020adaptive}, and given by, 

\begin{equation}\begin{split}
&\frac{\partial s}{\partial t} + \beta_i \left(1-\frac{A_e}{n_{pop}}\right)si + \beta_e \left(1-\frac{A_e}{n_{pop}}\right)se - \nabla \cdot (n_{pop}\nu_s\nabla s) = 0
\end{split}
\label{covid_s_dens}
\end{equation}
\begin{equation}
\begin{split}
&\frac{\partial e}{\partial t} - \beta_i \left(1-\frac{A_e}{n_{pop}}\right)si - \beta_e \left(1-\frac{A_e}{n_{pop}}\right)se + (\alpha  + \gamma_e) e - \nabla \cdot (n_{pop}\nu_e \nabla e) = 0
\end{split}
\label{covid_e_dens}
\end{equation}
\begin{equation}
\frac{\partial i}{\partial t} - \alpha e + (\gamma_i  + \delta) i - \nabla \cdot (n_{pop}\nu_i \nabla i) = 0
\label{covid_i_dens}
\end{equation}
\begin{equation}
\frac{\partial r}{\partial t}  -\gamma_e e- \gamma_i i  -\nabla \cdot (n_{pop} \nu_r \nabla r) = 0
\label{covid_r_dens}
\end{equation}
\begin{equation}
\frac{\partial d}{\partial t} - \delta i = 0.
\label{covid_d_dens}
\end{equation}

\noindent where $s(\mathbf{x}, t)$, $e(\mathbf{x}, t)$, $i(\mathbf{x}, t)$, $r(\mathbf{x}, t)$, and $d(\mathbf{x}, t)$ denote the densities of the \textit{susceptible}, \textit{exposed}, \textit{infected}, \textit{recovered}, and \textit{deceased} populations, respectively. The sum of all the compartments with the exception of $d(\mathbf{x},t)$ is represented by $n_{pop}$ which is the total living population. $A_e$ characterizes the Allee effect (persons), that takes into account the tendency of outbreaks to cluster around large populations,
$\beta_i$ and $\beta_e$ denote the transmission rates between symptomatic and susceptible individuals and asymptomatic and susceptible individuals, respectively (units days$^{-1}$),
$\alpha$ denotes the incubation period (units days$^{-1}$), $\gamma_e$ corresponds to the asymptomatic recovery rate (units days$^{-1}$), $\gamma_i$ the symptomatic recovery rate (units days$^{-1}$), $\delta$ represents the mortality rate (units days$^{-1}$), and $\nu_s$, $\nu_e$, $\nu_i$, $\nu_r$ are the diffusion parameters of the different population groups as denoted by the sub-scripted letters (units km$^2$  persons$^{-1}$  days$^{-1}$). Note that all these parameters can be considered time and space-dependent. 
We also compute the compartment $c$, the cumulative field of the $i$ compartment.

For the numerical solution of (\ref{covid_s_dens})-(\ref{covid_d_dens}), we discretize in space using a Galerkin finite element variational formulation. The resulting systems of equations are stiff, leading us to employ implicit methods for time integration. We apply the Backward Differentiation Formula (BDF2), which offers second-order accuracy while remaining unconditionally stable.  We implement the whole model in \texttt{libMesh} \cite{libmesh}. We additionally make use of AMR/C, allowing us to resolve multiple scales. One may find more details about the methods in \cite{Viguerie2021, grave2020adaptive}.

\subsubsection{Reproducing a 1D model}

\begin{figure}[htpb]
    \centering
\includegraphics[width=0.5\linewidth]{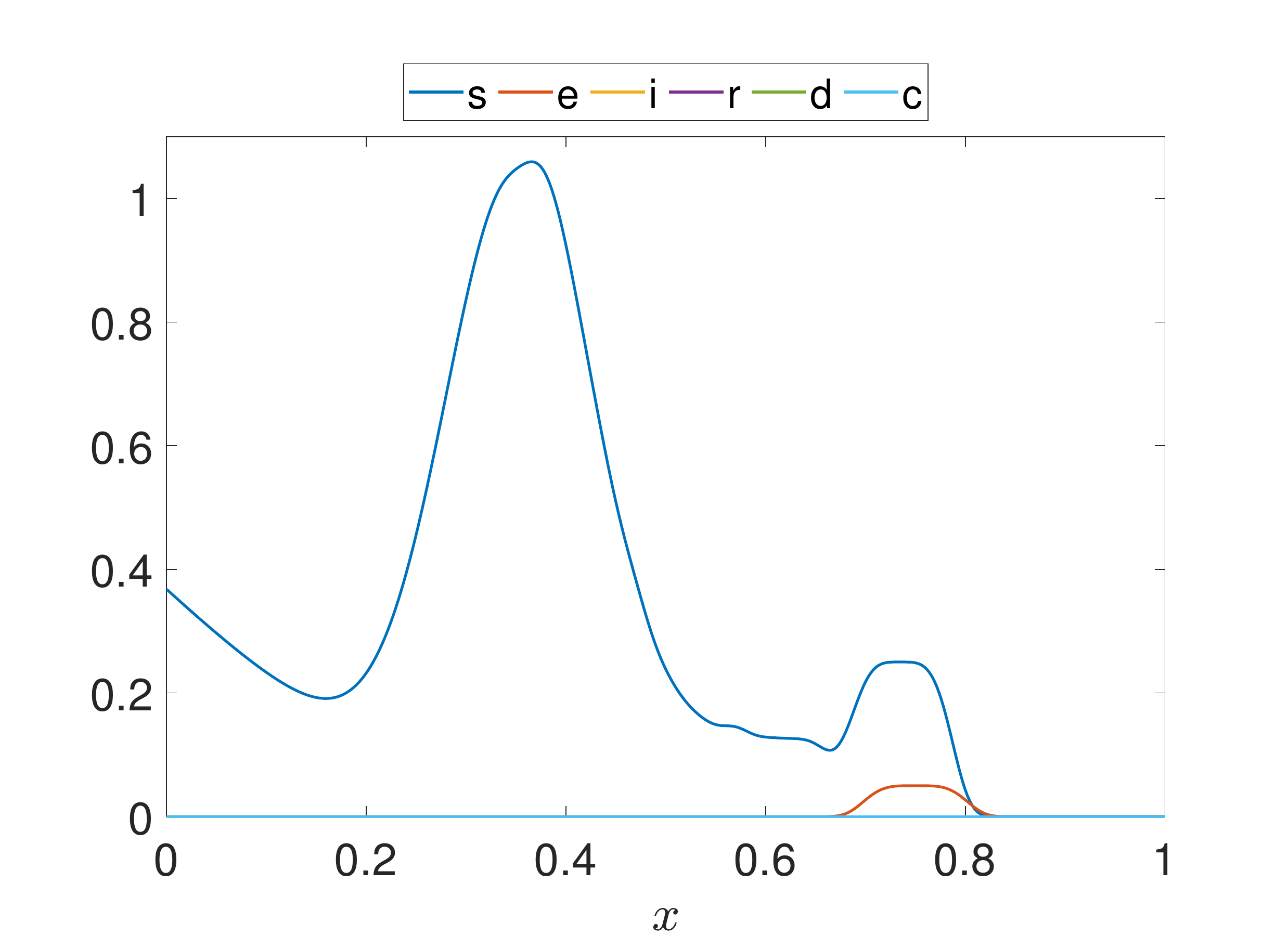}
    \caption{Initial conditions for the 1D model.}
    \label{fig:initial_conditions}
\end{figure}

First, we use a simple 1D continuous SEIRD model for COVID-19 with adaptive mesh refinement to validate the $\mathcal{L}^2-$projection and DMD. This example was introduced in \cite{Viguerie2021} and reproduced in \cite{grave2020adaptive}. Basically, we consider a 1D region $\Omega = [0,1]$ with initial conditions that represents a large population centered around $x = 0.35$ with no exposed persons and a small population centered around $x = 0.75$ with some exposed individuals, as shown in Figure \ref{fig:initial_conditions}. Thus, we set $s = s_0$ and $e = e_0$ as follows,

\begin{flalign}
    s_0 = &e^{-(x+1)^4}+e^{-\frac{(x-0.35) ^2}{10^{-2}}} \nonumber \\
    &+\frac{1}{8}\left(e^{-\frac{(x-0.62)^4}{10^{-5}}}+e^{-\frac{(x-0.52)^4}{10^{-5}}}+e^{-\frac{(x-0.42)^4}{10^{-5}}}\right) \\
    &+ \frac{1}{4}e^{-\frac{(x-0.735)^4}{10^{-5}}} \nonumber\\
    e_0= &\frac{1}{20}e^{-\frac{(x-0.75)^4}{10^{-5}}}
\end{flalign}

We further set $i_0 = 0$, $r_0 = 0$, and $d_0 = 0$. We also enforce homogeneous Neumann boundary conditions at $x = 0$ and a zero-population Dirichlet boundary condition at $x = 1$ for all model compartments. The latter represents a non-populated area at $x = 1$.

 Following \cite{Viguerie2021, grave2020adaptive} we set $\alpha = 0.09375$ days$^{-1}$, $\beta_i = \beta_e = 0.375$ days$^{-1}$persons$^{-1}$, $\delta = 0.0046875$ days$^{-1}$, $\gamma_i = 0.03125$ days$^{-1}$ and $\gamma_e = 0.125$ days$^{-1}$, $A=0$, $\nu_s = 3.75\times10^{-5}$, $\nu_e = 0.75\times 10^{-3}$, $\nu_i = 0.75\times10^{-10}$ and $\nu_r = 3.75\times 10^{-5}$ km$^2$persons$^{-1}$days$^{-1}$. The time step size is defined as $\Delta t = 0.25$ days and we consider the mesh projection and outputs at every time step, that is, $\Delta t_o = \Delta t = 0.25$.
 
 We use an adapted mesh with initially 125 elements, and after the refinement, the smallest element has a size 0.002. At the beginning of the simulation, we refine uniformly the whole domain into two levels and, after that, we apply the adaptive mesh refinement every 4 time steps. The idea is that the AMR/C strategy will keep this spatial resolution on more dynamically relevant regions while coarsening other regions in the domain. As a target reference mesh, we consider the uniformly refined mesh, such that all the domain contains the minimal spatial resolution obtained by the AMR/C simulation.
 
\begin{figure}[!ht]
	\begin{center}
		\subfigure{\includegraphics[width = \linewidth]{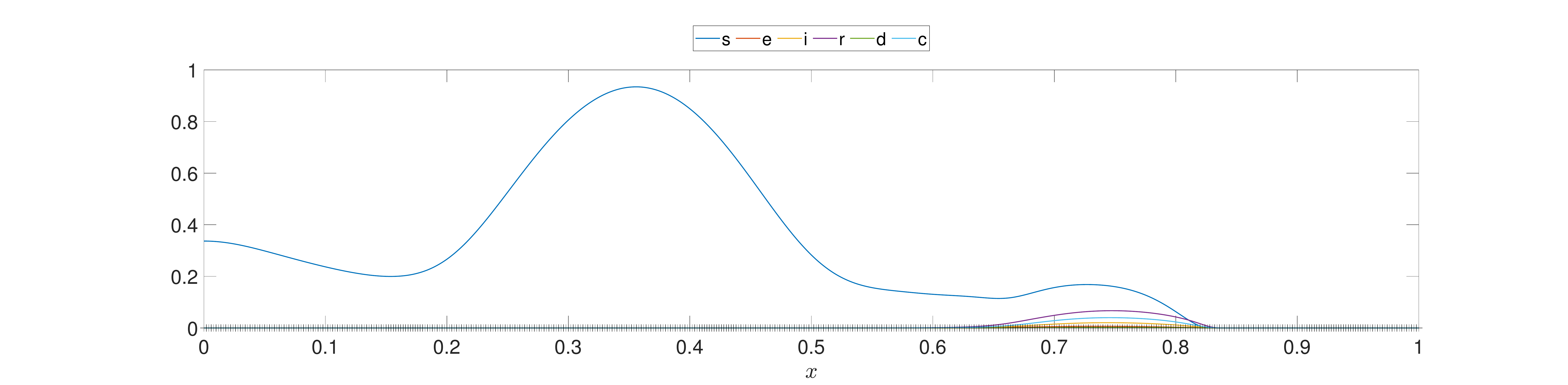}}
		\subfigure{\includegraphics[width = \linewidth]{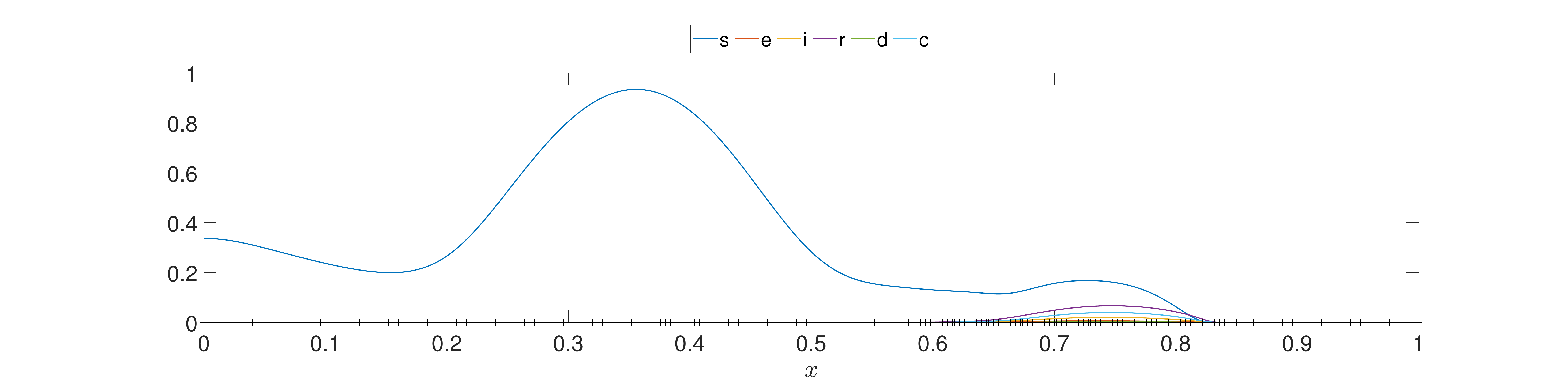}}
		\subfigure{\includegraphics[width = \linewidth]{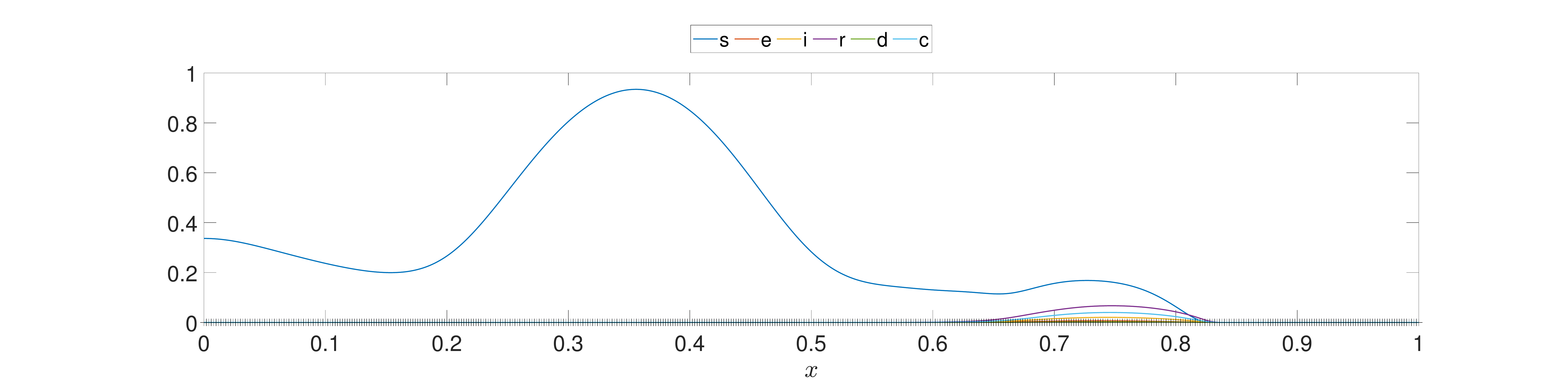}}
		\vspace{12pt}
		\caption{Solution at $t = 30$ days for the fixed mesh solution, AMR solution and the respective projection onto a reference mesh for the 1D SEIRD example. The reference mesh was built with characteristic length similar to the smaller elements in the adaptive mesh.}
		\label{fig:1D_seird_meshes}
	\end{center}
\end{figure}

The results of this simulation are validated against the results from \cite{Viguerie2021} and \cite{grave2020adaptive}. Figure \ref{fig:1D_seird_meshes} show the solution of the 1D SEIRD example at $t = 30$ days for the fixed mesh simulation, the adaptive mesh simulation, and the projection of the adaptive solution onto the target reference mesh used in the fixed mesh simulation. We observe a good agreement between the solutions. This agreement is a positive indicator that the DMD can be used on the projected meshes with little to no error compared to the DMD on a fixed mesh simulation. In terms of efficiency, Table \ref{tab:projection_1D} shows the computational effort required for the reference mesh projection embedded on the adaptive finite element code in comparison with the time used for the simulation code itself (AMR/C FEM).
 
\begin{table}[ht!]
    \centering
    \begin{tabular}{|c|c|c|}
    \hline
        Code Part               &   Absolute Time (s)   &   Relative Time ($\%$) \\
        \hline
        AMR/C FEM               &   $1262.89$          &   $98.53$ \\
        \hline
        Mesh Projection         &   $18.88$            &   $1.47$ \\
        \hline 
    \end{tabular}
    \caption{Absolute and relative time required for the projection routine in comparison with the adaptive finite element simulation code (AMR/C FEM) for the SEIRD model in the 1D case.}
    \label{tab:projection_1D}
\end{table}

\begin{figure}[!ht]
	\begin{center}
		\subfigure{\includegraphics[width = 0.49\linewidth]{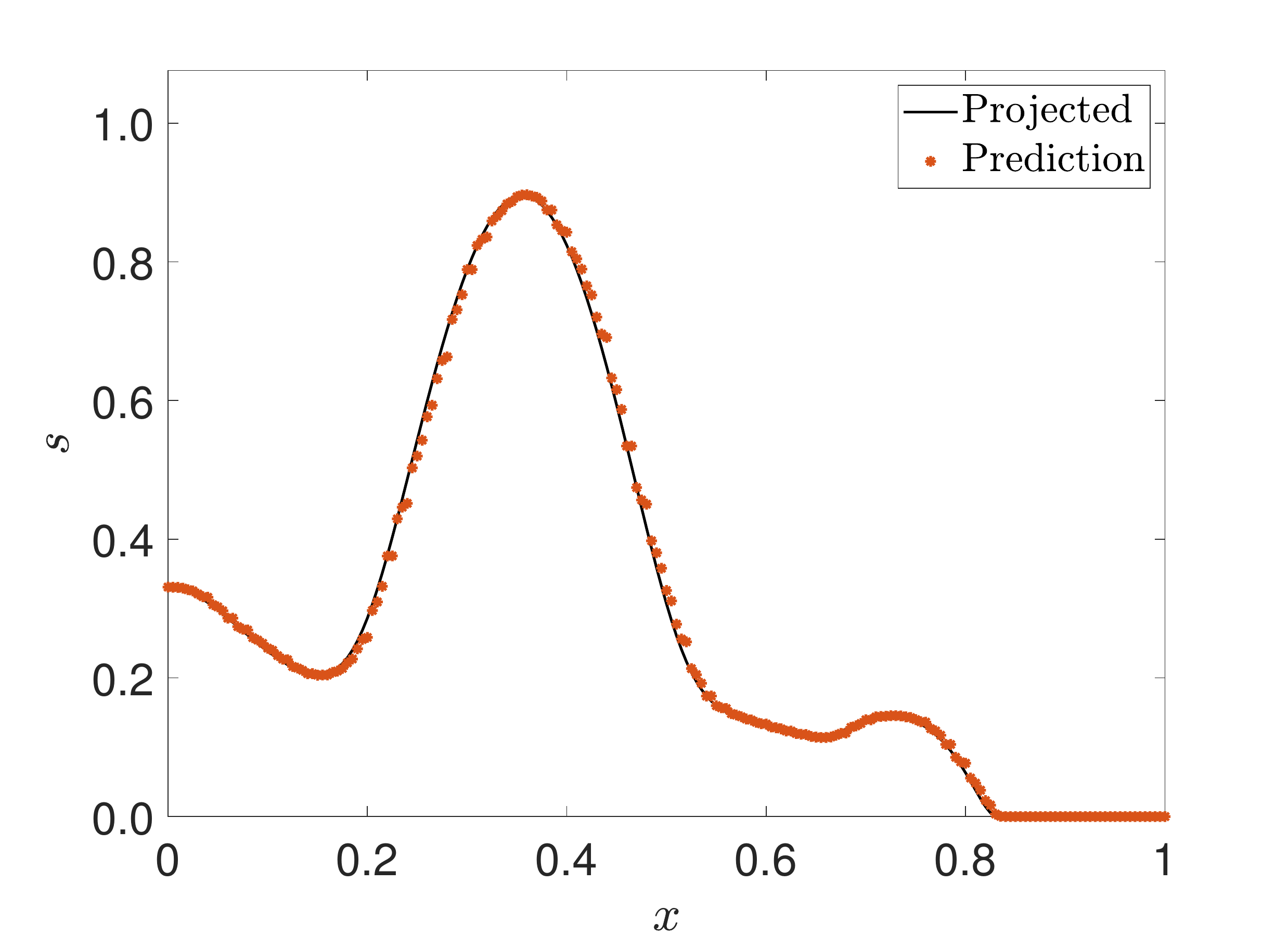}}
		\subfigure{\includegraphics[width = 0.49\linewidth]{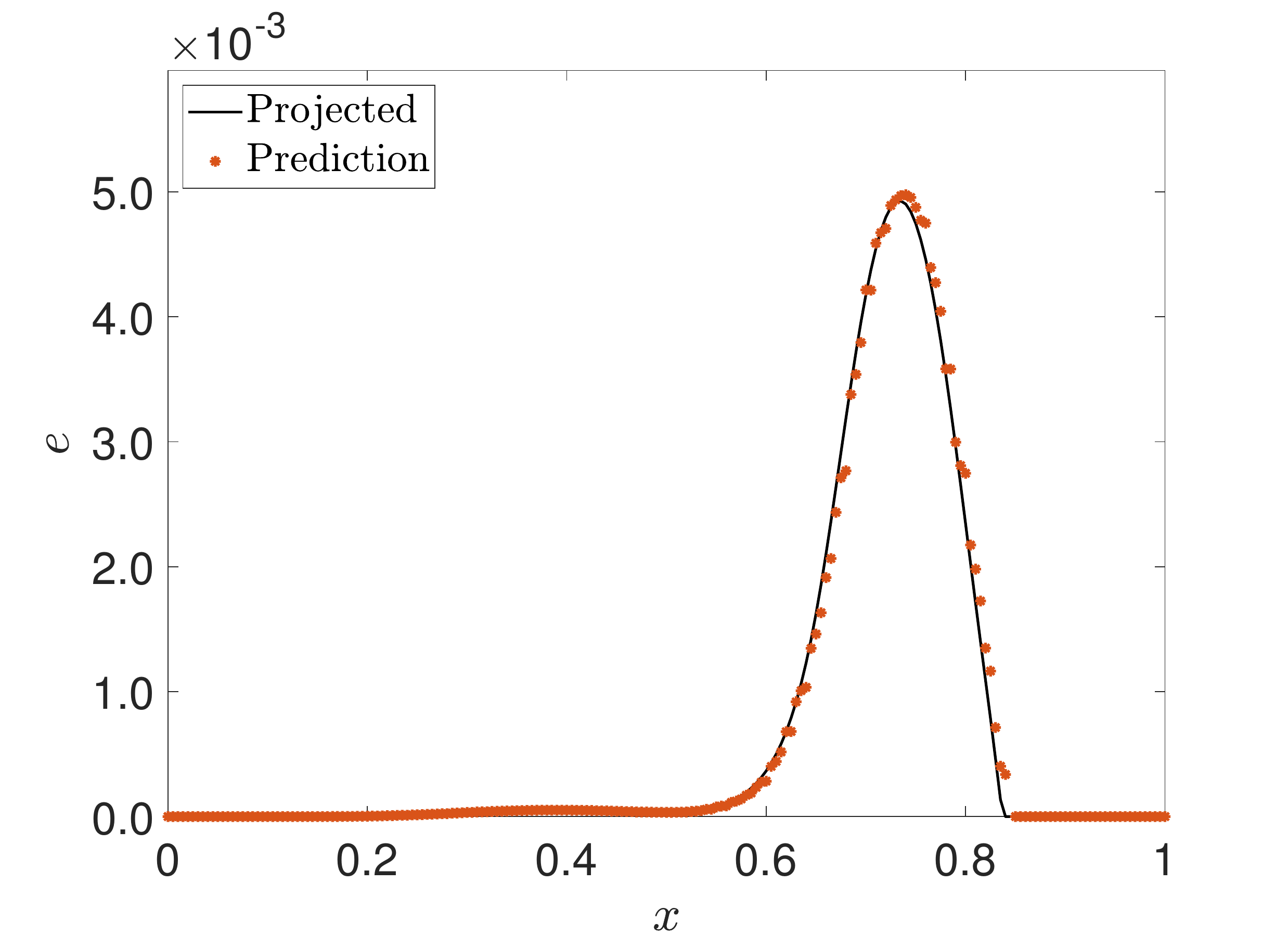}}
		\subfigure{\includegraphics[width = 0.49\linewidth]{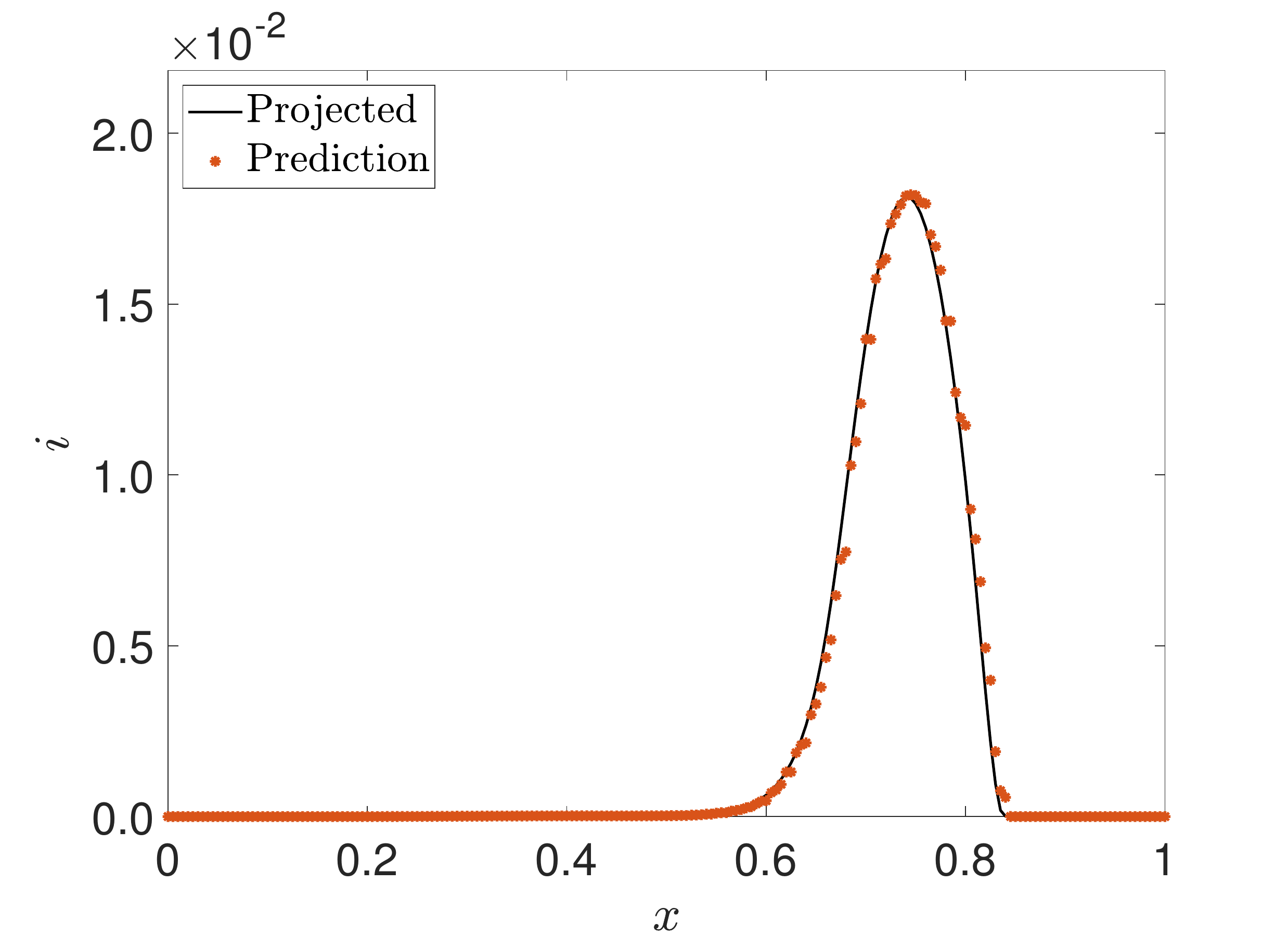}}
		\subfigure{\includegraphics[width = 0.49\linewidth]{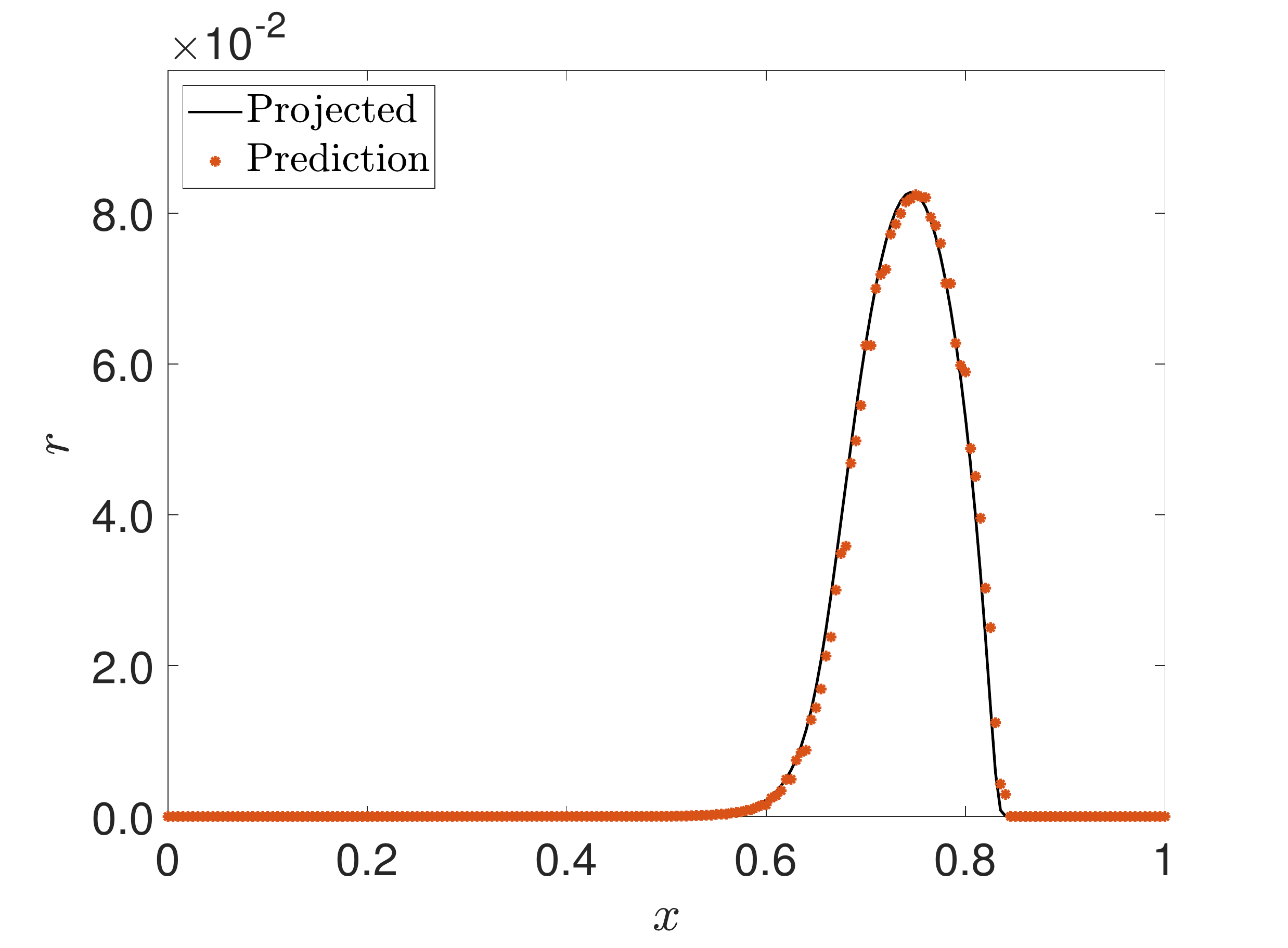}}
		\subfigure{\includegraphics[width = 0.49\linewidth]{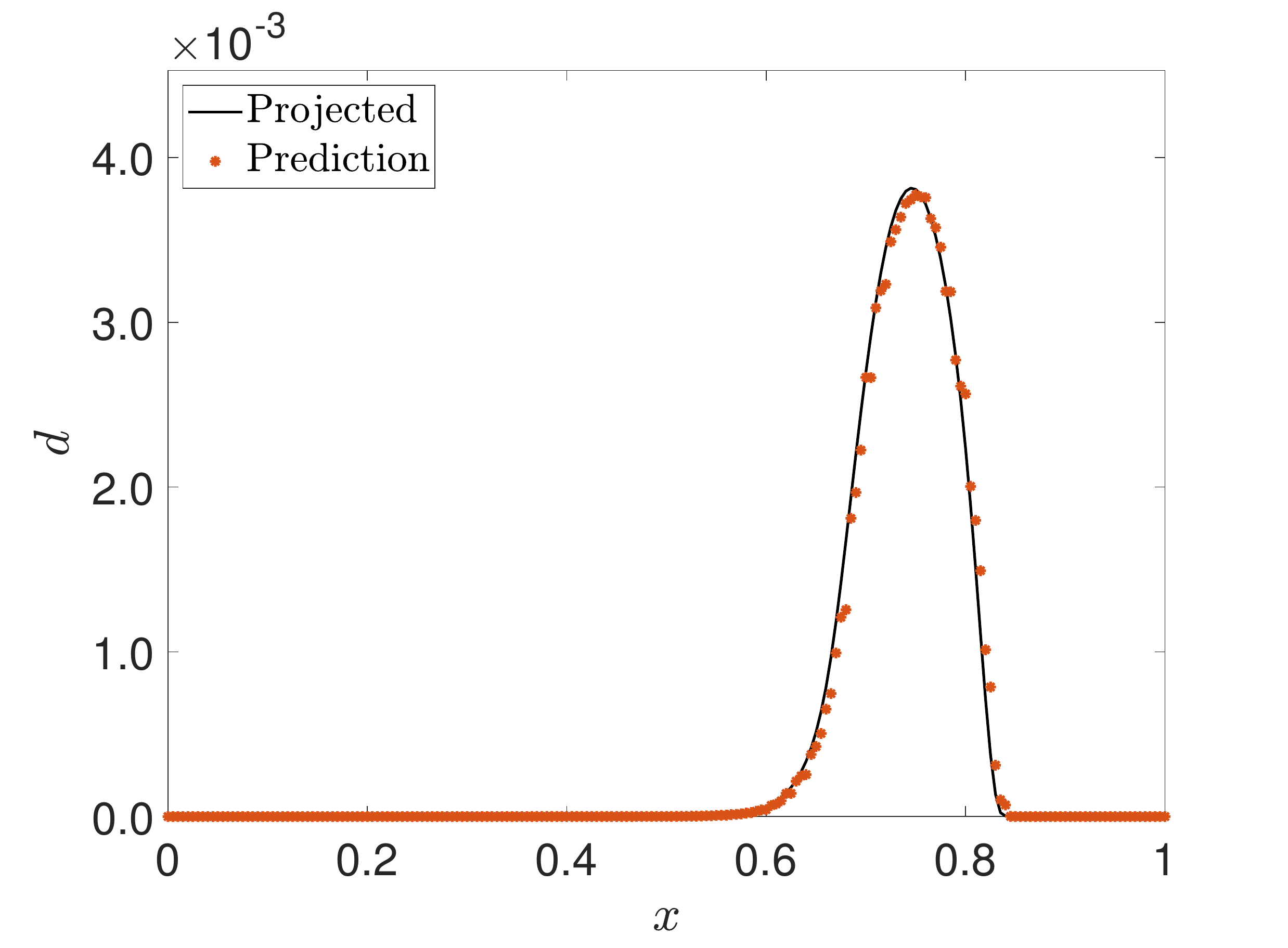}}
		\subfigure{\includegraphics[width = 0.49\linewidth]{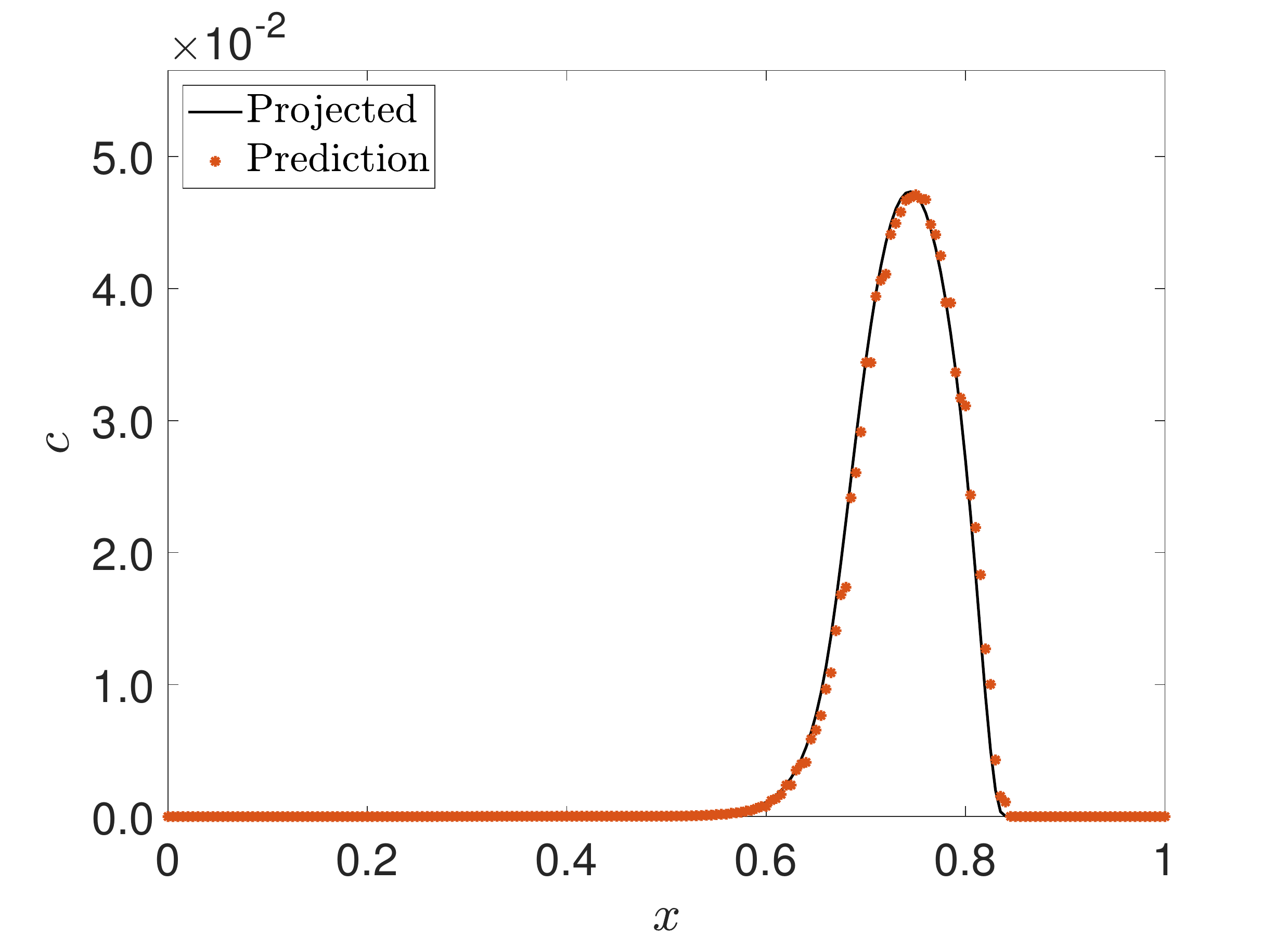}}
		\vspace{12pt}
		\caption{Solution at $t = 44$ days for the AMR simulation solution and the $14$ days projection using DMD for the 1D SEIRD example.}
		\label{fig:1D_seird_pred}
	\end{center}
\end{figure}

After some initial experiments, we set $r = 15$ for all compartments. In this example, the finite element simulation computes the results for $44$ days, that is, $176$ time steps. However, we only consider the first $30$ days in the snapshot matrix for reconstruction. DMD is set to approximate the results for a further $14$ days so the results can be compared with the $44$ days results obtained in the finite element simulation. In other words, we want to evaluate the DMD ability to predict the COVID-19 scenario two weeks in the future given the data of the last $30$ days. For numerical reasons, we considered an initial $3$ days shift in the snapshots since some compartments are initialized with zeros, affecting how DMD captures the dynamics. The $44$th day solution for the adaptive simulation and the $44$th day prediction considering DMD are seen in Figure \ref{fig:1D_seird_pred}. We observe good agreement between predictions and the numerical solutions of the simulations for most compartments. The speedup and overall relative error between the DMD approximation and the snapshots are seen in Table \ref{tab:efficiency_1D}.

\begin{table}[ht!]
    \centering
    \caption{Relative error between reconstructed (and predicted) data and the projected snapshots.}
    \begin{tabular}{|c|c|c|}
    \hline
    Compartments & Relative Error   &   Speedup \\
    \hline
    $s$ & $1.590 \times 10^{-3}$    &   $1.079\times10^3$\\
    \hline
    $e$ & $2.574 \times 10^{-2}$    &   $1.316\times10^3$\\
    \hline
    $i$ & $1.162 \times 10^{-2}$    &   $1.225\times10^3$\\
    \hline
    $r$ & $1.439 \times 10^{-2}$    &   $1.264\times10^3$\\
    \hline
    $d$ & $2.001 \times 10^{-2}$    &   $1.331\times10^3$\\
    \hline
    $c$ & $1.286 \times 10^{-2}$    &   $1.341\times10^3$\\
    \hline
    \end{tabular}
    \label{tab:efficiency_1D}
\end{table}

\subsubsection{The Lombardy region}

We extend our analysis by solving the continuous SEIRD model and applying DMD to a 2D real world domain that is the Lombardy region in Italy. The spread of the COVID-19 has been studied in this region using the continuous SEIRD model with accurate results \cite{Viguerie2021, viguerie2020diffusion}. Here, we reproduce this simulation with the solver developed in \cite{grave2020adaptive} which invokes adaptive mesh refinement every 4 time steps. We use the same parameters as shown in \cite{Viguerie2021, viguerie2020diffusion}. It is important to point that, in this simulation, the transmission rates and diffusion parameters vary with time in order to reproduce the effects of restrictions during the simulated period.
\begin{figure}[ht!]
    \centering
    \includegraphics[width=0.5\linewidth]{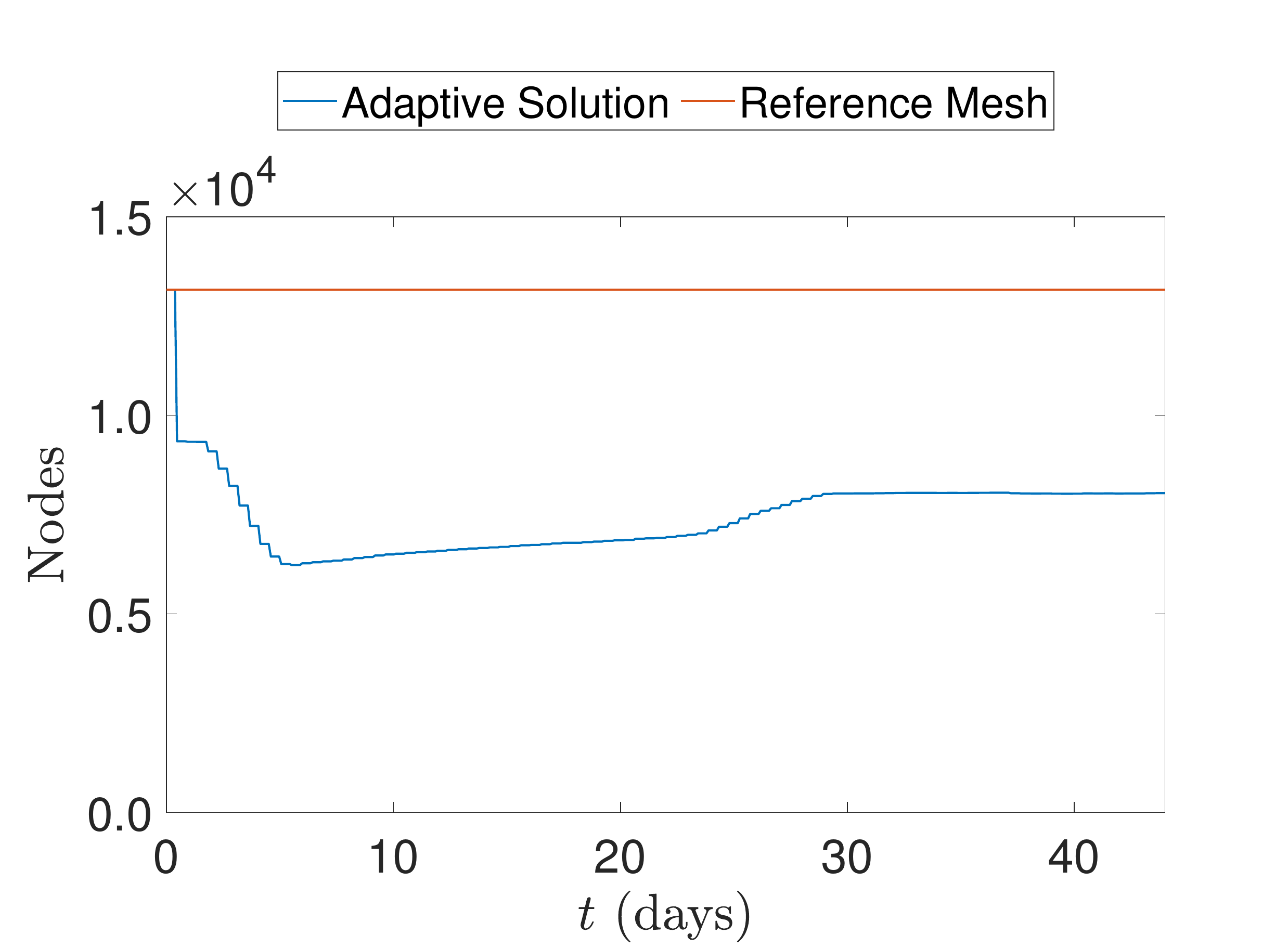}
    \caption{Number of mesh nodes in time for the adaptive solution and the proposed reference mesh for the 2D SEIRD example.}
    \label{fig:lombardy_dofs}
\end{figure}

\begin{figure}[ht!]
    \centering
    \includegraphics[width=\linewidth]{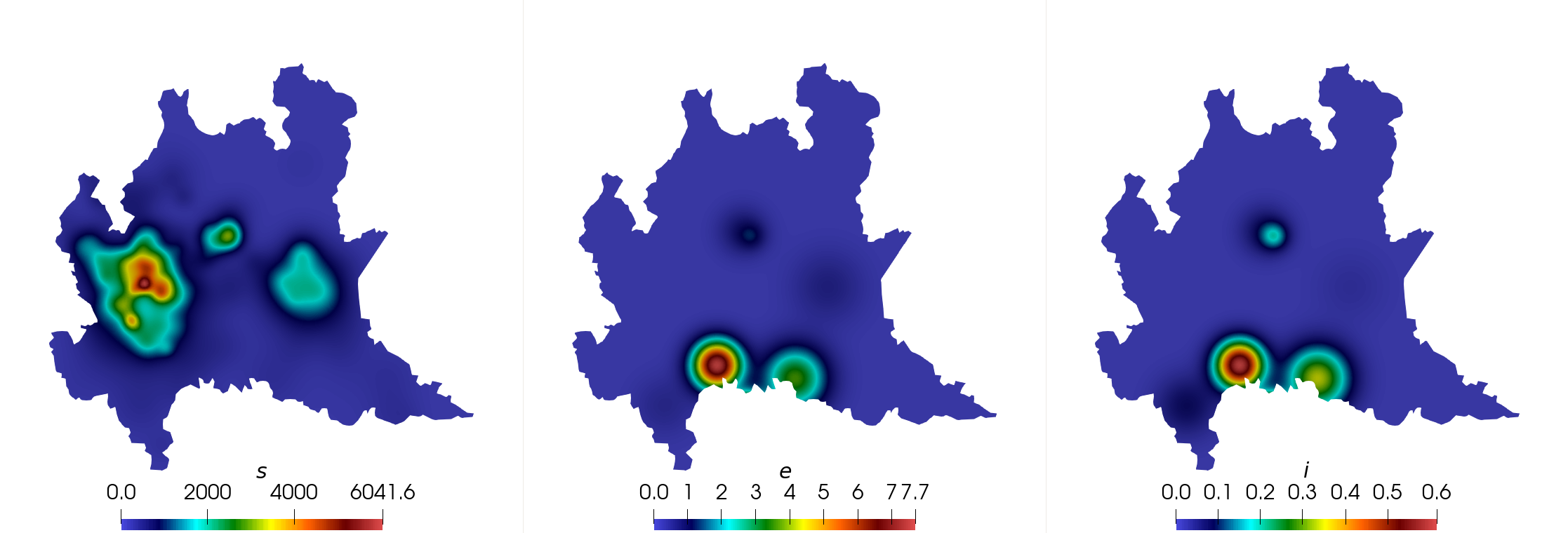}
    \caption{Initial conditions for the SEIRD model in the Lombardy case.}
    \label{fig:lombardy_IC}
\end{figure}

\begin{figure}[!ht]
	\begin{center}
		\subfigure[Fixed mesh solution]{\label{fig:s_adapt}\includegraphics[width = 0.7\linewidth]{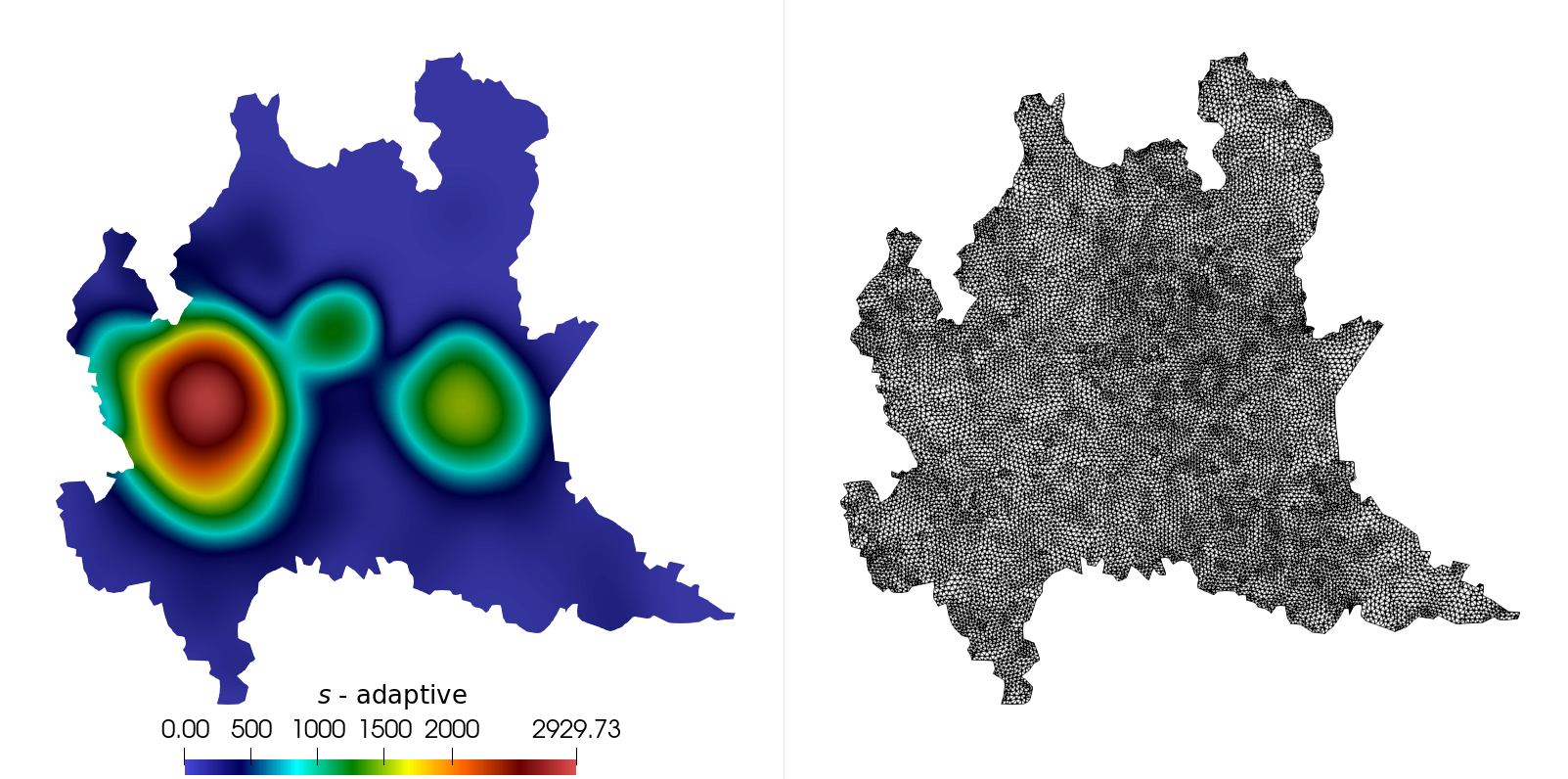}}
		\subfigure[Adaptive solution]{\label{fig:s_adapt}\includegraphics[width = 0.7\linewidth]{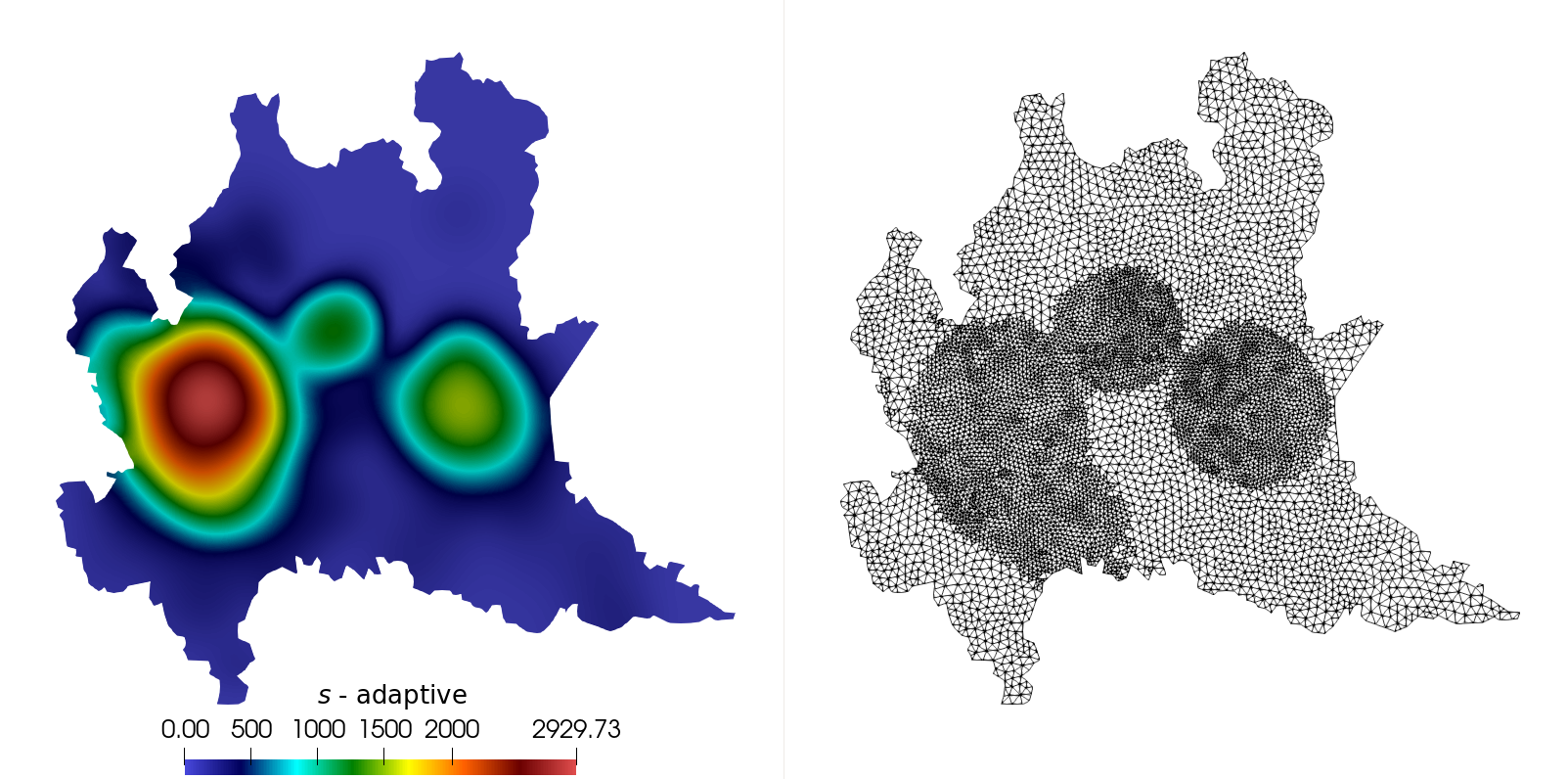}}
		\subfigure[Projection onto reference mesh]{\label{fig:s_projected}\includegraphics[width = 0.7\linewidth]{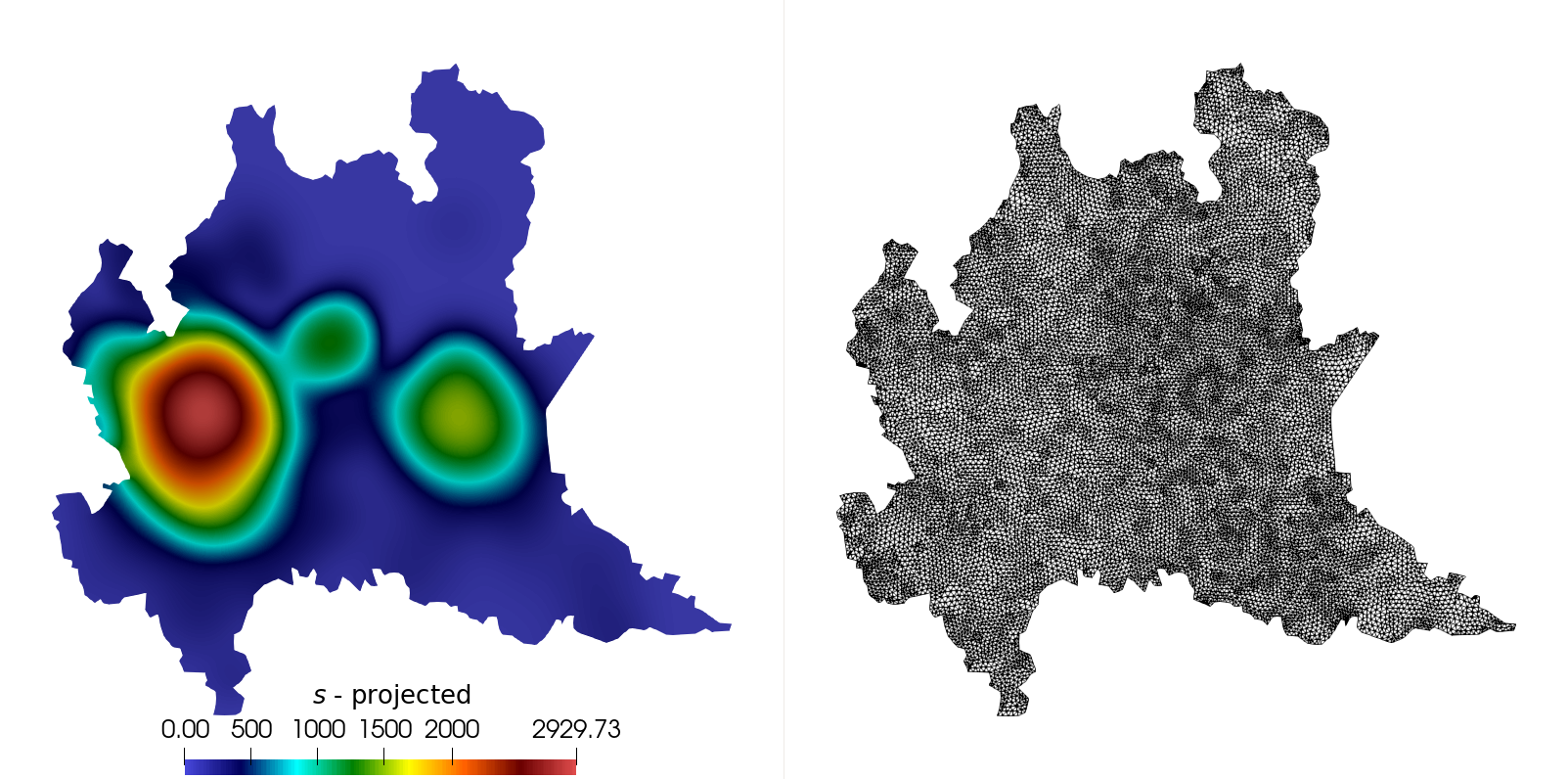}}
		\vspace{12pt}
		\caption{Solution for the susceptible compartment at $t = 46$ days obtained using an adaptive mesh and its respective projection onto a fixed reference mesh.}
		\label{fig:projected_mesh_lombardy}
	\end{center}
\end{figure}

For this simulation, an unstructured mesh is considered due to the complex geometry imposed by the domain. The mesh is generated using \texttt{Gmsh} and is uniformly refined as the simulation starts. After refining the whole mesh in one level, the mesh presents a minimum spatial resolution of approximately 1 kilometer. This procedure allows the solver to coarsen the regions where no significant dynamics are observed while preserving the scales of the regions of interest. Figure \ref{fig:lombardy_dofs} shows the variation of the number of nodes in time for the AMR/C strategy. The coarsening approach improves the simulation performance significantly since the average number of nodes (and respectively, the number of equations $n_{eq}$) used in the adaptive simulation is approximately half with respect to the case of a fixed mesh with the same spatial resolution considered for the entire domain. In this example, the reference target mesh is the uniformly refined unstructured mesh considered in the early stages of the simulation, presenting $13158$ nodes and $25340$ elements. 
The simulation considers a time step size of $\Delta t = 0.25$ days for the numerical integration and $\Delta t_o = \Delta t = 0.25$ days for the observations. Initial conditions for the Lombardy domain are the same presented on \cite{Viguerie2021,viguerie2020diffusion} and are seen in Figure \ref{fig:lombardy_IC}, while compartments $r$ and $d$ are initialized to zero.
We then proceed to run both adaptive and fixed mesh simulations and, to apply DMD in the adaptive mesh results, we consider the proposed projection scheme.  Figure \ref{fig:projected_mesh_lombardy} shows the $s$ compartment solution at $t=46$ days for both simulations and the projected adaptive solution onto the reference mesh, revealing that the results are in good agreement. In terms of efficiency, Table \ref{tab:projection_SEIRD} shows results for the computational time required for the projection compared to the finite element code. That said, the adaptive snapshots can now be assembled into a snapshot matrix for the DMD reconstruction and prediction.  
\begin{table}[ht!]
    \centering
    \begin{tabular}{|c|c|c|}
    \hline
        Code Part               &   Absolute Time (s)   &   Relative Time ($\%$) \\
        \hline
        AMR/C FEM               &   $2107.24$          &   $98.52$ \\
        \hline
        Mesh Projection         &   $31.60$            &   $1.48$ \\
        \hline 
    \end{tabular}
    \caption{Absolute and relative time required for the projection routine in comparison with the adaptive finite element simulation code (AMR/C FEM) for the SEIRD model in the Lombardy case.}
    \label{tab:projection_SEIRD}
\end{table}

\begin{figure}[!ht]
	\begin{center}
		\subfigure{\label{fig:s_last_step}\includegraphics[width = \linewidth]{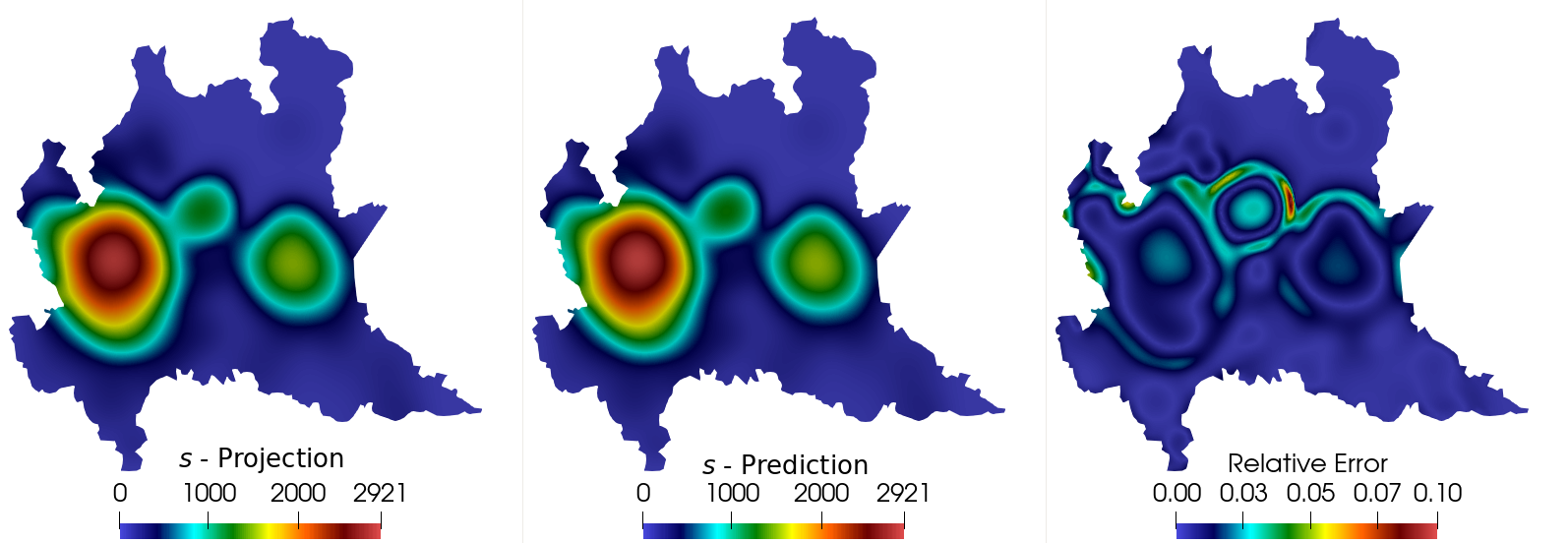}}
		\subfigure{\label{fig:e_last_step}\includegraphics[width = \linewidth]{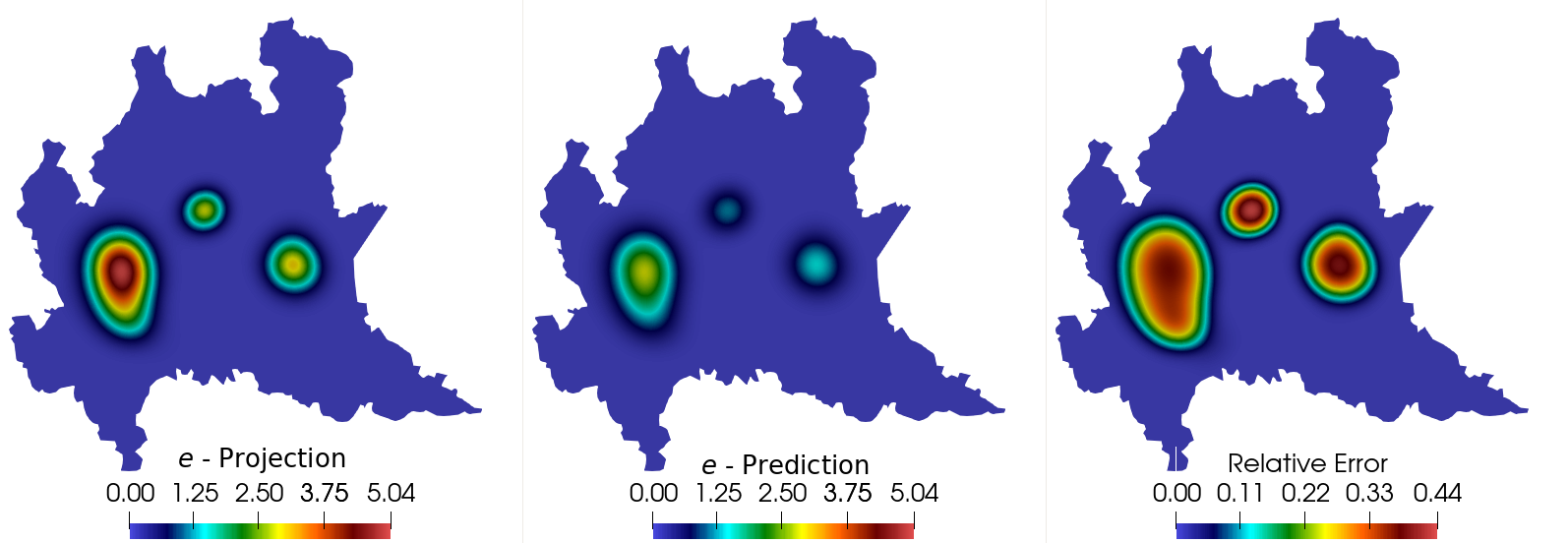}}
		\subfigure{\label{fig:i_last_step}\includegraphics[width = \linewidth]{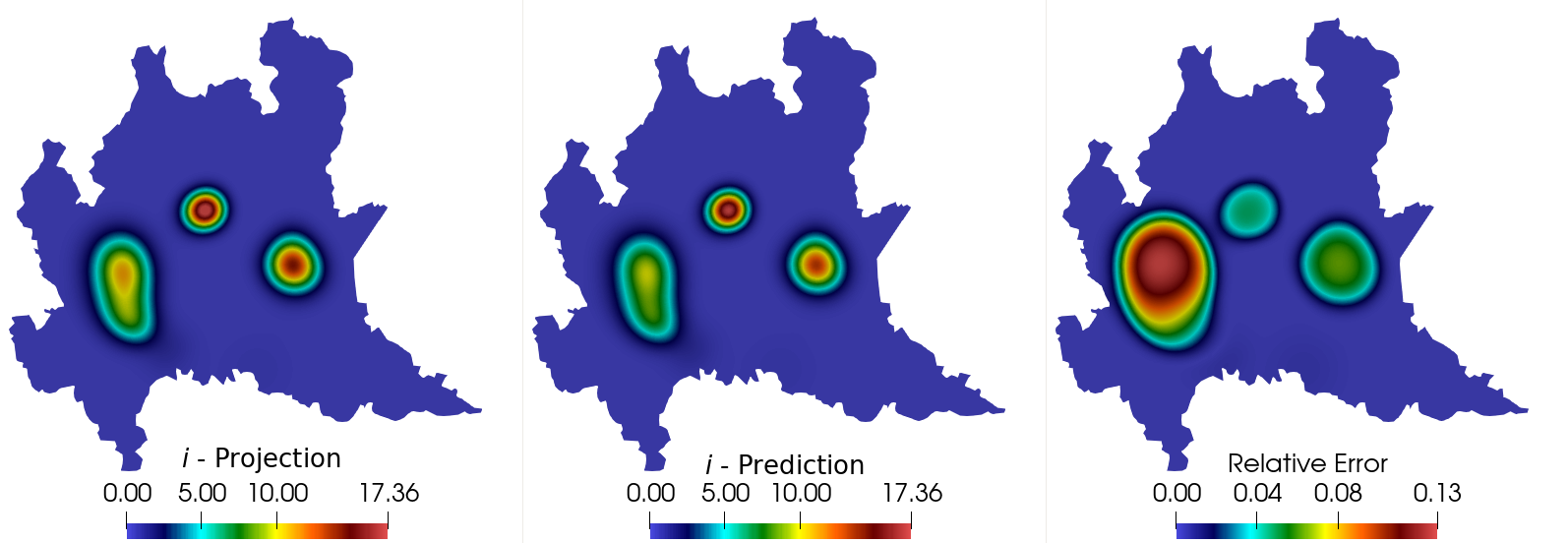}}
		\vspace{12pt}
		\caption{Comparison between computed and predicted solutions at $t=60$ days for the susceptible, exposed, and infected compartments.}
		\label{fig:last_step_lombardy}
	\end{center}
\end{figure}

\begin{figure}[!ht]
	\begin{center}
		\subfigure{\label{fig:r_last_step}\includegraphics[width = \linewidth]{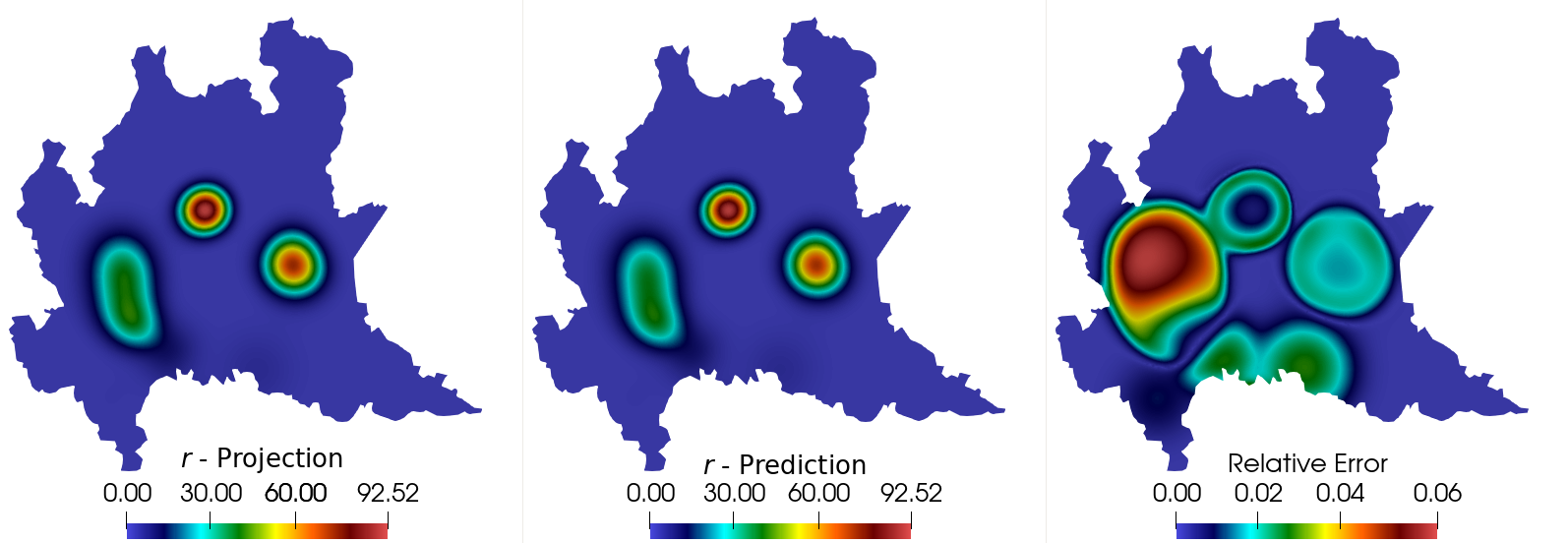}}
		\subfigure{\label{fig:d_last_step}\includegraphics[width = \linewidth]{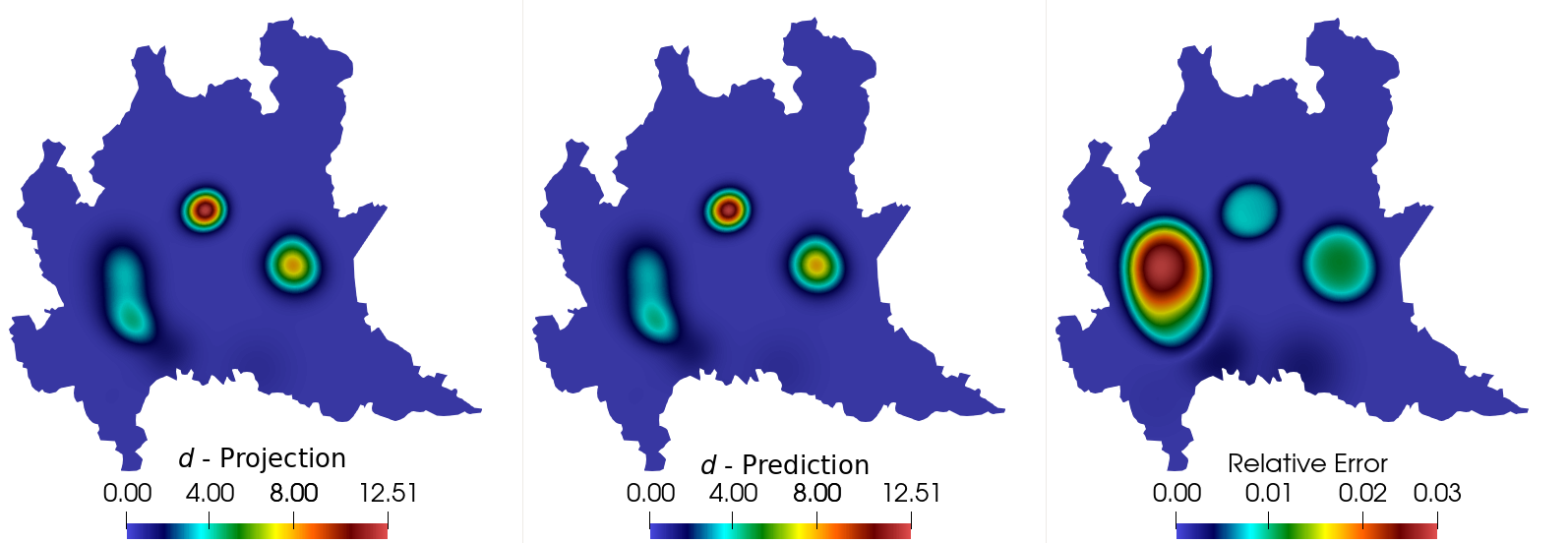}}
		\subfigure{\label{fig:c_last_step}\includegraphics[width = \linewidth]{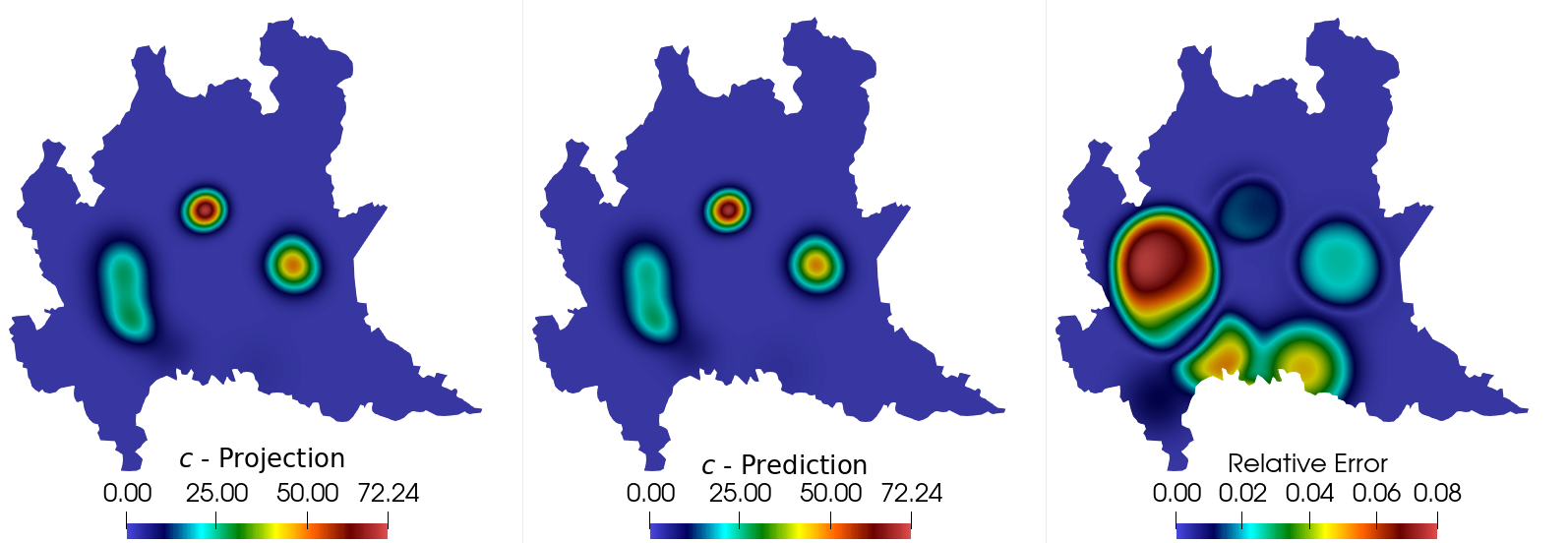}}
		\vspace{12pt}
		\caption{Comparison between computed and predicted solutions at $t=60$ days for the recovered, deceased, and cumulative infected compartments.}
		\label{fig:last_step_lombardy2}
	\end{center}
\end{figure}

\begin{figure}[!ht]
	\begin{center}
		\includegraphics[width = 0.5\linewidth]{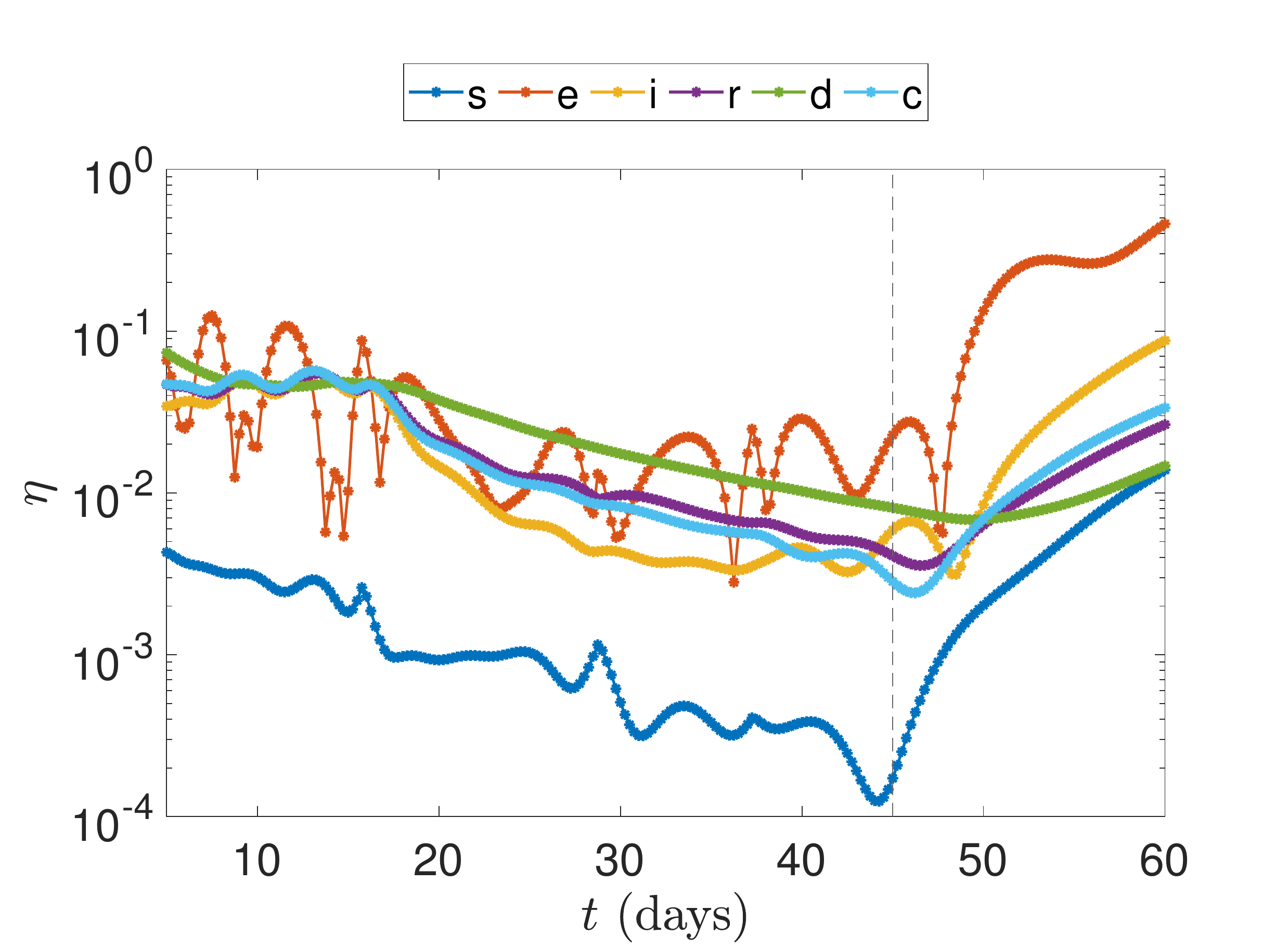}
		\vspace{12pt}
		\caption{Relative error for all compartments between numerical simulation snapshots and DMD reconstruction and prediction. The dashed line represents the beginning of the DMD prediction stage.}
		\label{fig:relative_error_lombardy}
	\end{center}
\end{figure}

The DMD analysis is made in the same way as presented in the 1D case: the simulation outputs the projected snapshots for the first $60$ days ($240$ snapshots). The snapshot matrix assembles the information regarding $46$ days of simulations, while DMD approximates the results for $60$ days. The idea is to predict two weeks in the future, given the data observed in the past $46$ days. For this example, we consider the SVD truncation at $r = 20$ for all compartments. Again, an initial $3$ days shift in the data is considered to avoid issues with the compartments initialized to zero. Results for the $60$th day comparing the computed numerical solutions and the DMD predictions are seen in Figures \ref{fig:last_step_lombardy} and \ref{fig:last_step_lombardy2}. We present the projected solutions, the DMD prediction, and the relative error in space between the two results from left to right. We can note that most compartments show results in agreement with the simulations, while the exposed compartment reveals more pronounced differences than the other compartments.
Figure \ref{fig:relative_error_lombardy} shows the relative error in time between the DMD results and the projected snapshots. The first thing we notice is that the curves are different for each compartment. This discrepancy occurs due to the different parameters for each equation in the SEIRD model, which largely affects the dynamics of the system. The dynamics for each compartment are different since each compartment presents different coupling,  diffusion, and reaction parameters. Also, regarding this issue, since the parameters are time and space-dependent, sudden changes in their values can affect the dynamics of the system as well as DMD's dynamics mapping ability. Some sudden changes in the $s$ and $e$ compartments related to stricter public policies considered to reduce the transmission rates (parameters $\beta_i$ and $\beta_e$) are incorporated into the model. Since the variation in the parameters is not introduced smoothly, DMD's ability to map sudden changes in the dynamics of the system is reflected by the existence of some spikes on the curves of the relative errors in time. Comparing the reconstruction and prediction stages, we observe that the errors tend to grow as soon as the prediction stage starts (dashed line). The exposed compartment, which yielded most of the oscillations due to parameter changing on the reconstruction stage, presented the same behavior on the prediction phase around day $49$. We also note that the exposed compartment yields a larger relative error for the $60$th day in comparison with the other compartments. Table \ref{tab:efficiency_lombardy} shows the overall relative error and the speedups for the six compartments approximations. Comparing these results with the results presented in Figures \ref{fig:last_step_lombardy}, \ref{fig:last_step_lombardy2}, and \ref{fig:relative_error_lombardy}, we can conclude that the predictions are reasonably accurate in comparison with the numerical solutions, specially when considering the time required for calculation. 

\begin{table}[ht!]
	\centering
	\caption{Relative error between reconstructed (and predicted) data and the computed snapshots and speedup between DMD and the numerical simulation.}
	\begin{tabular}{|c|c|c|}
		\hline
		Compartments  & Relative Error ($\eta_F$) & Speedup\\
		\hline
		$s$ & $1.345 \times 10^{-3}$   & $822.61$ \\
		\hline
		$e$  & $5.187 \times 10^{-2}$  & $938.09$  \\
		\hline
		$i$  & $1.304 \times 10^{-2}$  & $755.97$ \\
		\hline
		$r$  & $7.316 \times 10^{-3}$  & $1036.93$ \\
		\hline
		$d$  & $1.097 \times 10^{-2}$  & $977.91$  \\
		\hline
		$c$  & $1.251 \times 10^{-2}$  & $977.03$ \\
		\hline
	\end{tabular}
	\label{tab:efficiency_lombardy}
\end{table}

\begin{figure}[!ht]
	\begin{center}
		\includegraphics[width = 0.5\linewidth]{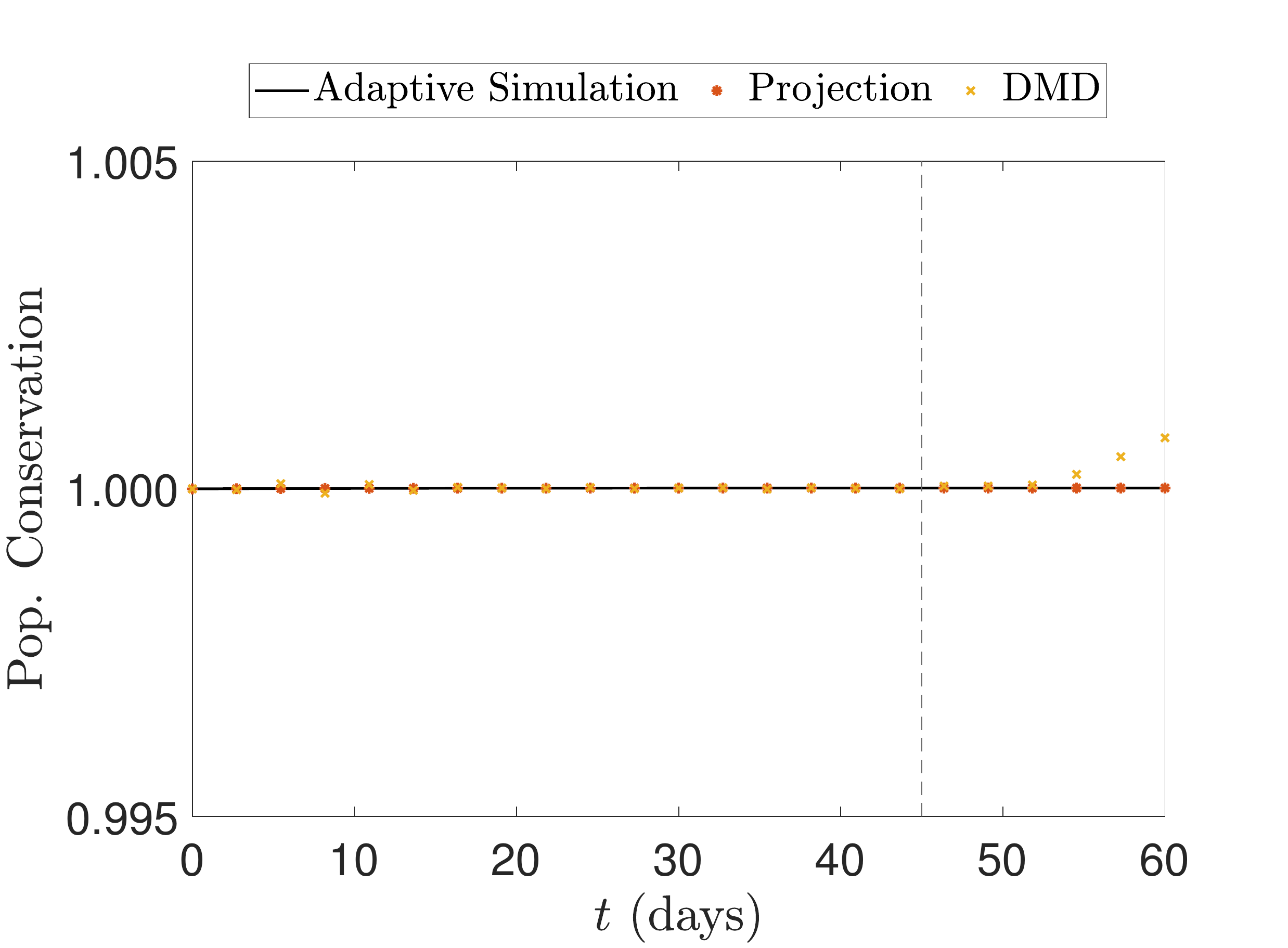}
		\caption{Population conservation for both adaptive and projected results.}
		\label{fig:mass_lombardy}
	\end{center}
\end{figure}

Another important analysis to be done is the conservation property of the continuous SEIRD model. As mentioned before, the standard $\mathcal{L}^2-$projection does not guarantee conservation among the projections. Figure \ref{fig:mass_lombardy} shows the total population during the simulation, normalized by the total population modeled in the initial conditions. The total population is computed as the sum of the integral of the compartments (excluding $c$) divided by the sum of the integral of the elements of the mesh. Since the SEIRD model does not consider any population growth, the value must be theoretically constant for all the simulations. From the figure, we observe that the population is kept constant during all the adaptive simulation, and this was preserved by the projected solutions and the DMD reconstruction stage. That is, we can note that the $\mathcal{L}^2-$projection does not yield conservation issues in this example. For the prediction phase, DMD preserves the total population for several days in the future. However, it presents a slight increase (around $0.1\%$) for predictions over $10$ days, which does not affect the results significantly. This increase can be explained by the relative errors behavior, observed in Figure \ref{fig:relative_error_lombardy}, as DMD computes future estimates. 

\subsection{Fluid dynamics}
This section evaluates the DMD use on two cases involving AMR/C in computational fluid dynamics: the reconstruction of a 2D density-driven gravity current simulation and the temporal prediction on a 3D rising bubble. Different from the previous cases, the test cases presented in this section are advection-dominant. To approximate the governing equations, we use a finite element RBVMS formulation \cite{hughes,rasthofer,ahmed2017,codina2018,Guerra2013}. In the first case, we reconstruct the solution and evaluate important quantities of interest regarding density-driven gravity flows. On the bubble rising simulation, we show how our strategy works on a 3D mesh, and we evaluate the DMD ability to predict the quantities of interest of the rising bubble in time.

\subsubsection{Density-driven gravity flow}
\begin{figure}
    \centering
    \includegraphics[width=0.7\linewidth]{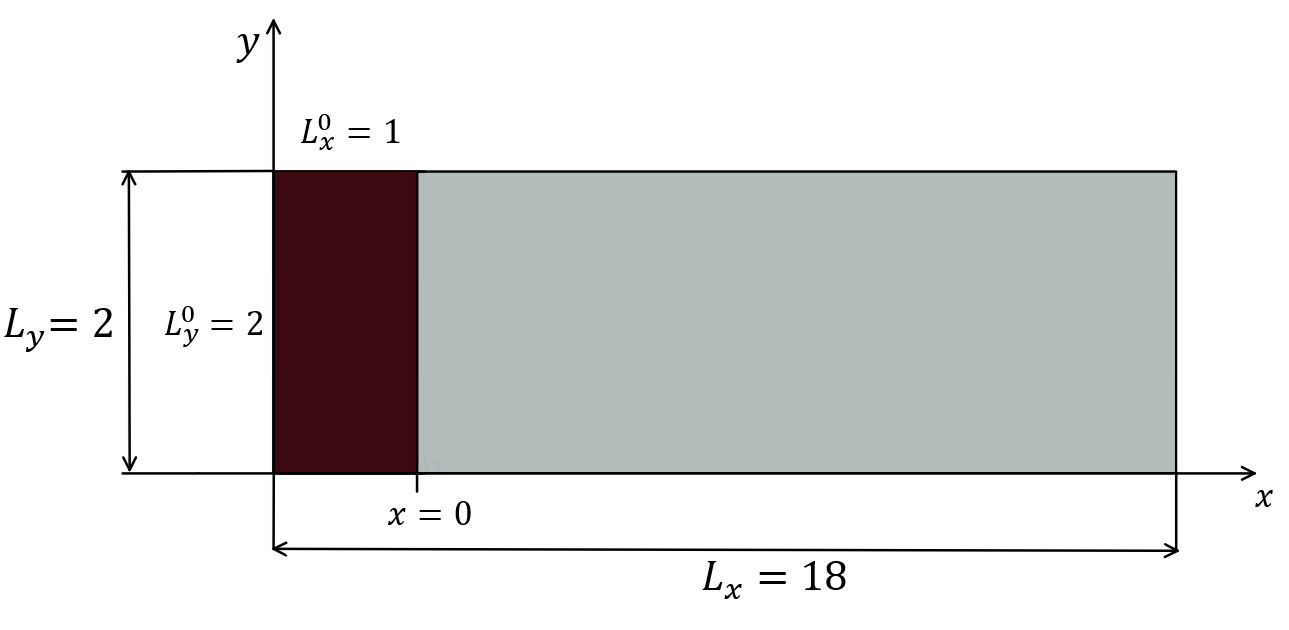}
    \caption{Scheme illustrating the initial conditions for the density-driven gravity flow example.}
    \label{fig:tank}
\end{figure}

In this section, we consider a long numerical simulation that consists of a lock-exchange between two fluids, the heavy fluid, A, and the lighter fluid, B, based on the numerical example in \cite{Necker2002}. The difference between their densities is such that the Boussinesq hypothesis is considered valid. Moreover, particles in the heavy fluid have negligible inertia and are much smaller than the smallest length scales of the buoyancy-induced fluid motion. Thus, the dimensionless governing equations are,
\vspace{6pt}
\begin{center}
	\begin{equation}
	\begin{aligned}
	\nabla\cdot \mathbf{u} = 0,\\
	\dfrac{\partial \mathbf{u}}{\partial t} + \mathbf{u}\cdot \nabla\mathbf{u} + \nabla p - \dfrac{1}{\sqrt{Gr}}\Delta \mathbf{u} - \phi\mathbf{e}^g = 0,\\
	\dfrac{\partial \phi}{\partial t} + \mathbf{u}\cdot \nabla \phi - \dfrac{1}{Sc\sqrt{Gr}}\Delta \phi = 0,
	\end{aligned}
	\label{eq:LE}
	\end{equation}
\end{center}
\vspace{6pt}
where $\mathbf{u}$ is the fluid velocity, $\phi$ is the concentration field, $p$ is the pressure, $\mathbf{e}^g = (0,-1)$ is the vector pointing in the direction of gravity, $Sc = 1.0$ is the Schmidt number and $Gr = 5 \times 10^{6}$ is the Grashof number, two dimensionless numbers that relate viscous effects with diffusion and buoyancy effects, respectively. A Grashof number of this magnitude indicates a turbulent flow. The field $\phi = \phi_A/\phi_B$ is the concentration and is responsible for mapping the evolution of fluid interactions. The time step size considered for this simulation consists on $\Delta t = 0.01$s for a total simulation time of $T = 30$s with an output frequency of $\Delta t_o = \Delta t = 0.01$s.
We consider a tank, that is, a rectangular domain with length $L_1 = 18$m, height $L_2 = 2$m. The boundary conditions for this case are no-slip for the velocity and no-flux for the transport equation, and the initial conditions are such that the heavy fluid is represented as a column with dimensions $L_x^0 \times L_y^0 = 1$m $\times$  $2$m located at the left border of the tank and the light fluid fills the rest of the domain. Figure \ref{fig:tank} illustrates the domain and the initial conditions.
\par
\begin{figure}[!ht]
	\begin{center}
		\subfigure[Adaptive solution]{\label{fig:le_adapt_solution}\includegraphics[width=0.49\linewidth]{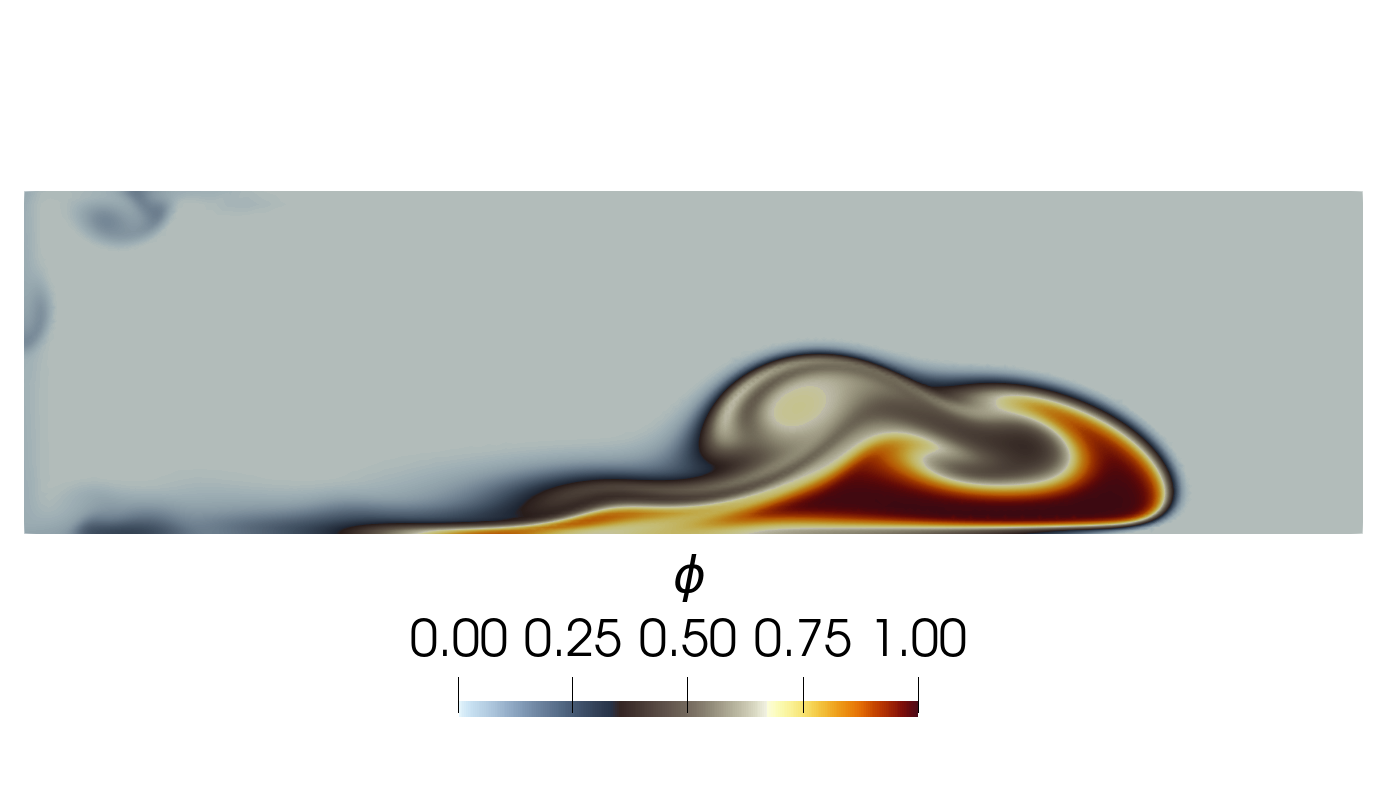}}
		\subfigure[Adaptive mesh]{\label{fig:le_adapt_mesh}\includegraphics[width=0.49\linewidth]{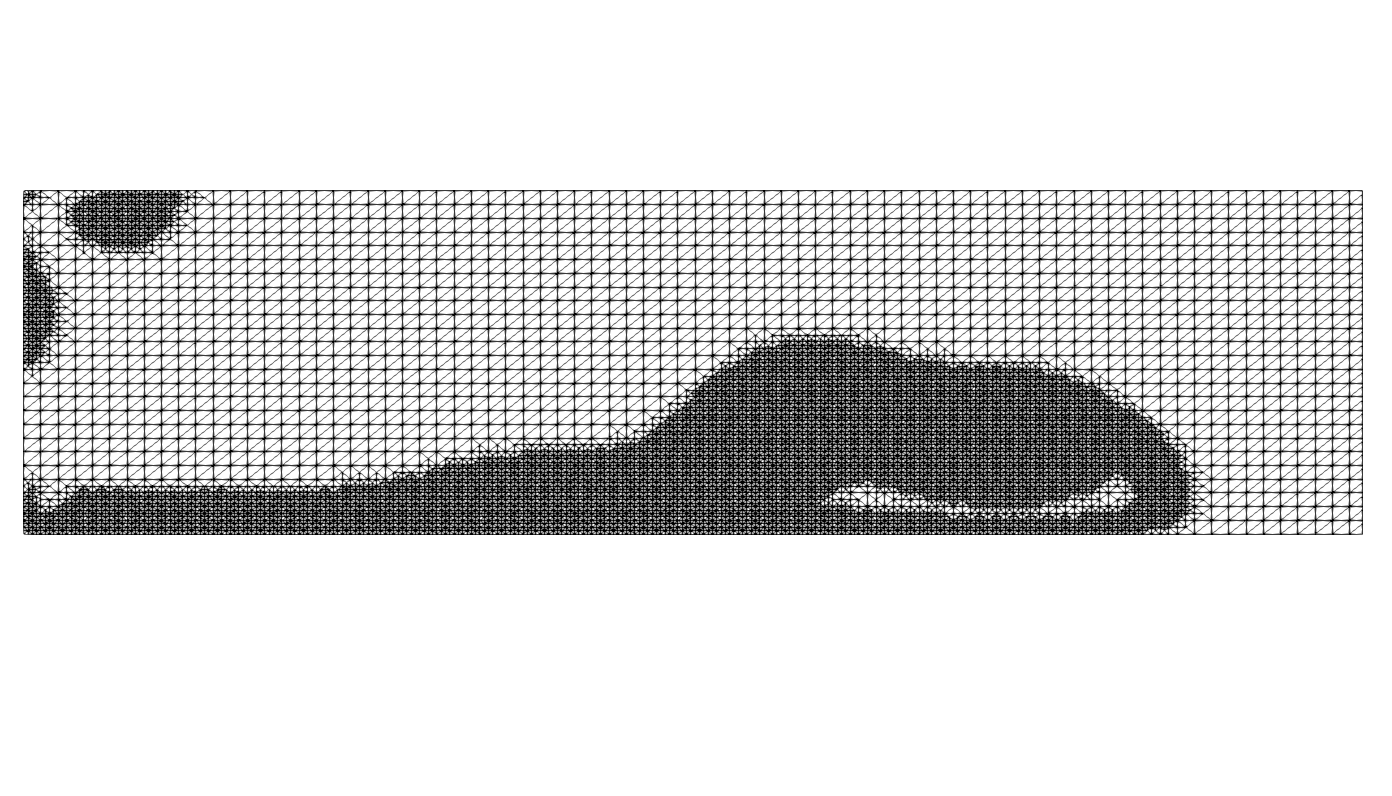}}
		\subfigure[Projected solution in the structured mesh]{\label{fig:le_proj_mesh}\includegraphics[width=0.49\linewidth]{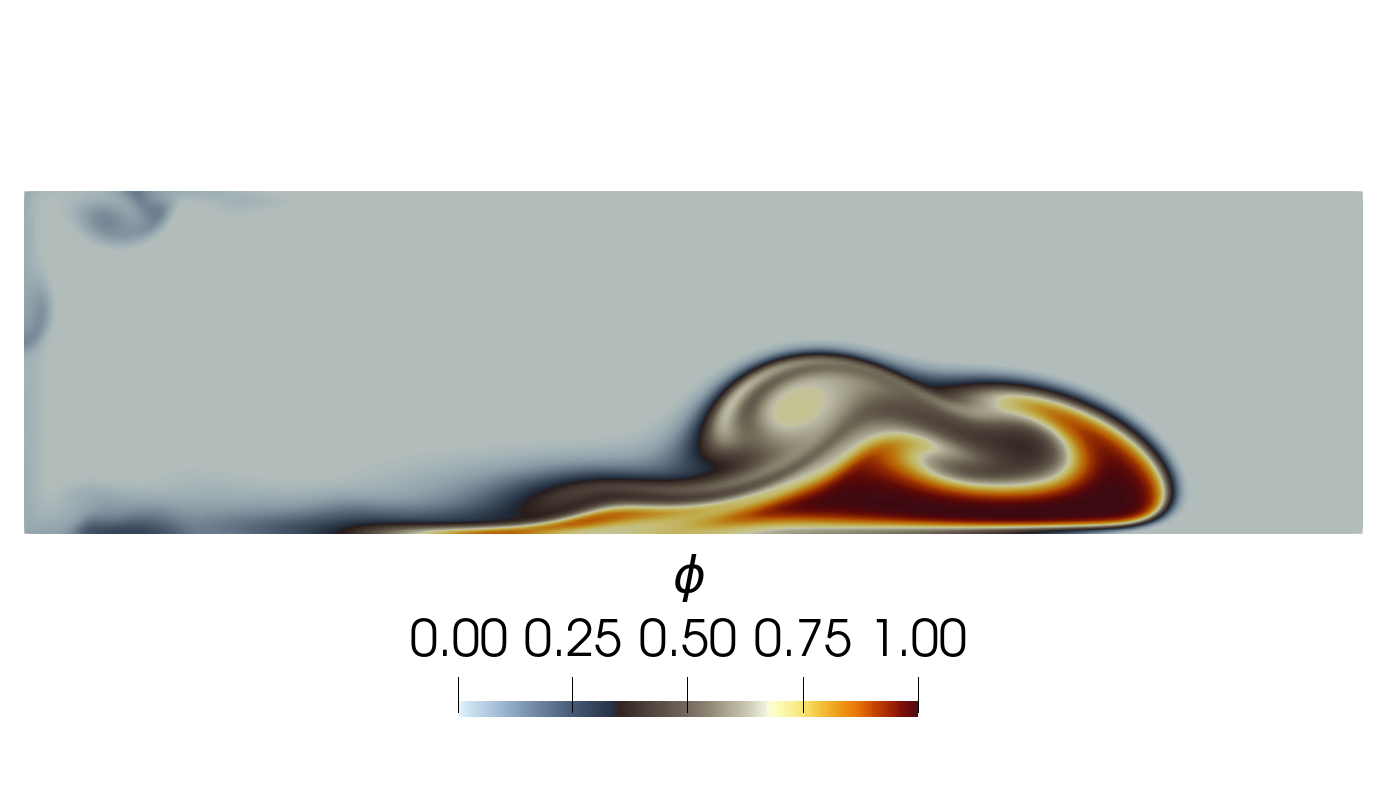}}
		\subfigure[Fixed structured mesh solution]{\label{fig:le_fixed_mesh}\includegraphics[width=0.49\linewidth]{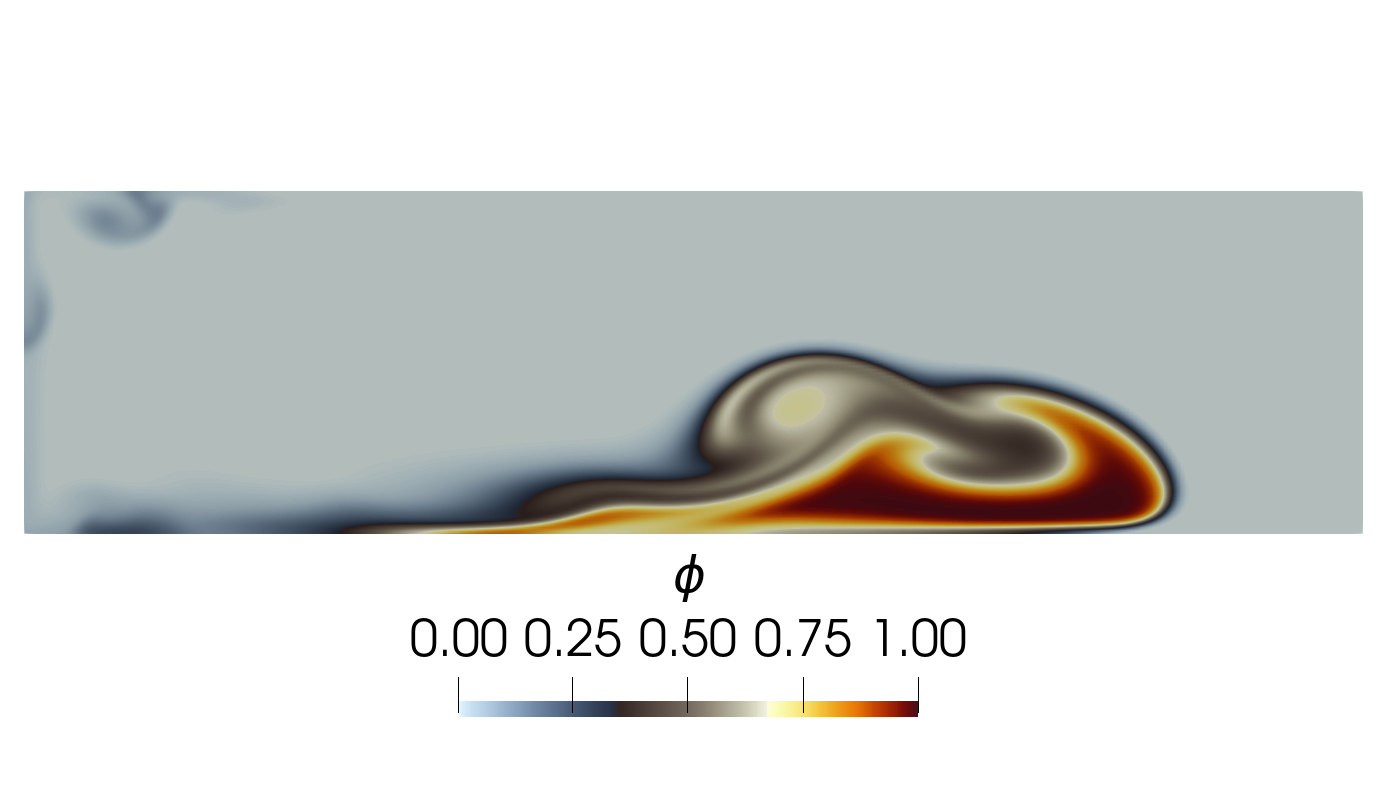}}
		\vspace{12pt}
		\caption{ Results and mesh for the first 8 meters of the domain at $t = 10$s.}
		\label{fig:le}
	\end{center}
\end{figure}

\begin{figure}[ht!]
    \centering
    \includegraphics[width=0.5\linewidth]{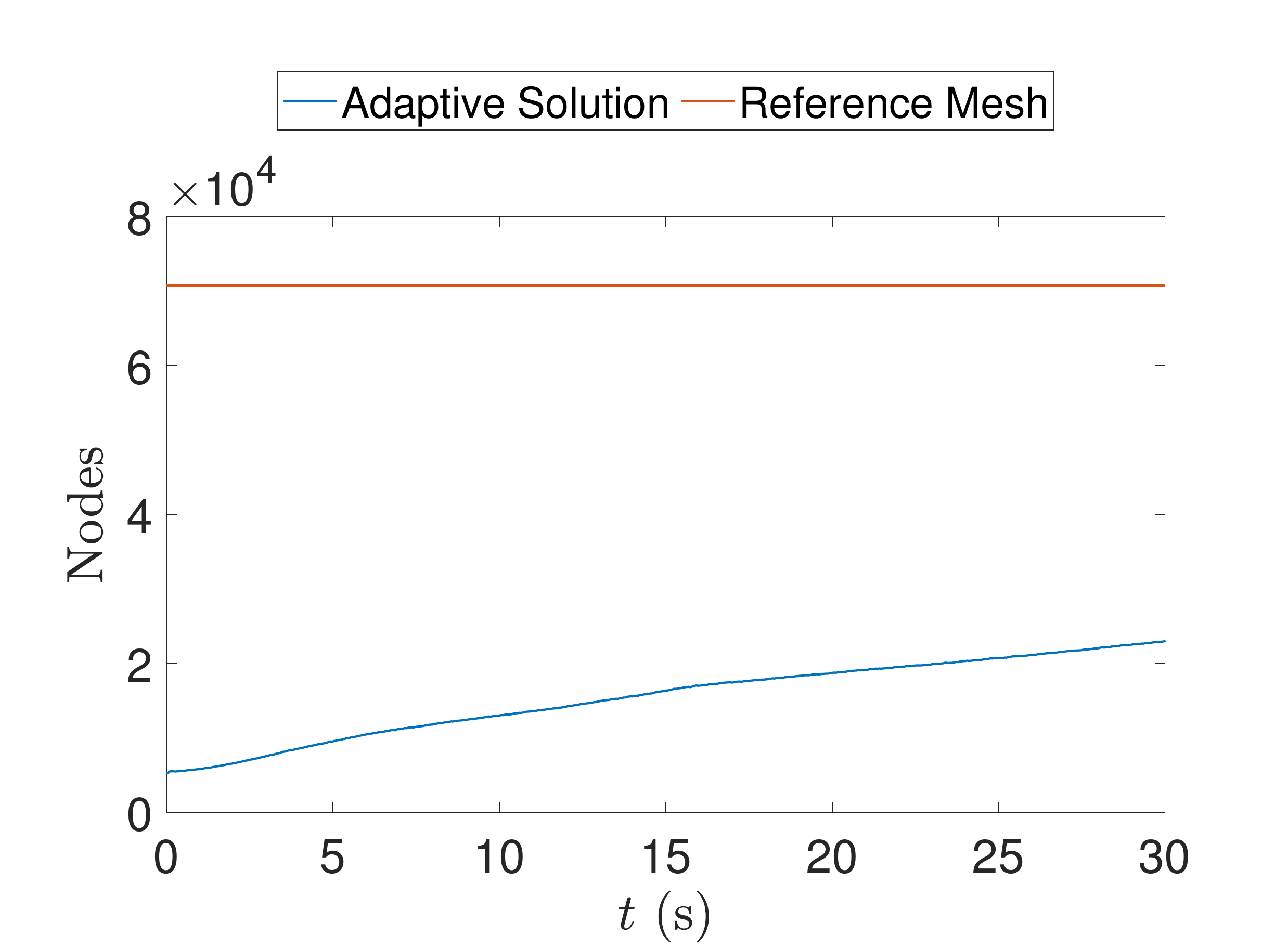}
    \caption{Number of mesh nodes in time for the adaptive solution and the proposed reference mesh for the density-driven gravity current example.}
    \label{fig:lockexchange_dofs}
\end{figure}

\begin{figure}[!ht]
	\begin{center}
		\includegraphics[width=0.5\linewidth]{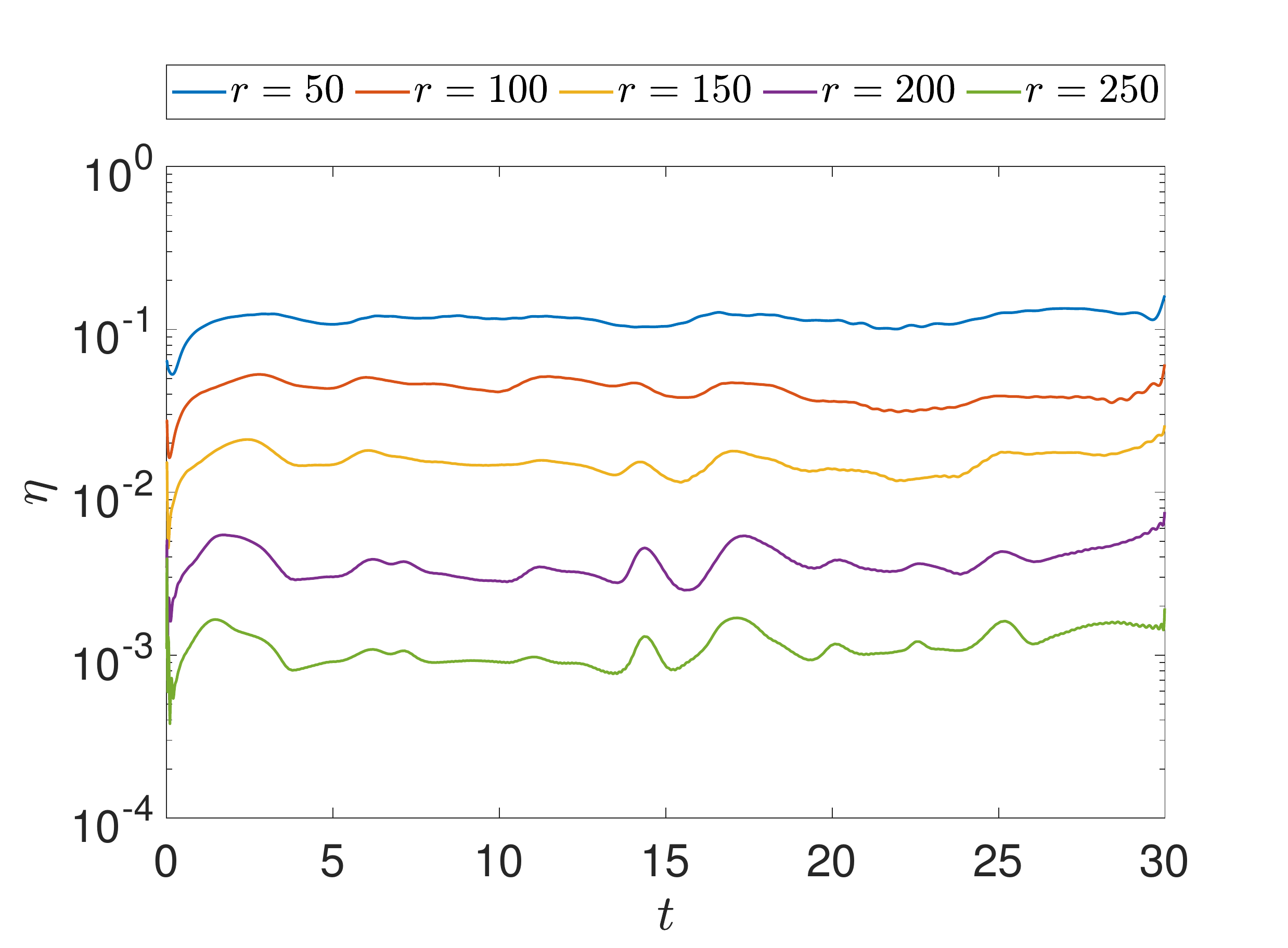}
		\vspace{12pt}
		\caption{Relative error for the reconstruction considering different values of the rank $r$.}
		\label{fig:le_rel_error}
	\end{center}
\end{figure}

\begin{figure}[!ht]
	\begin{center}
		\subfigure[Front position]{\label{fig:le_fixed_sol}\includegraphics[width=0.6\linewidth]{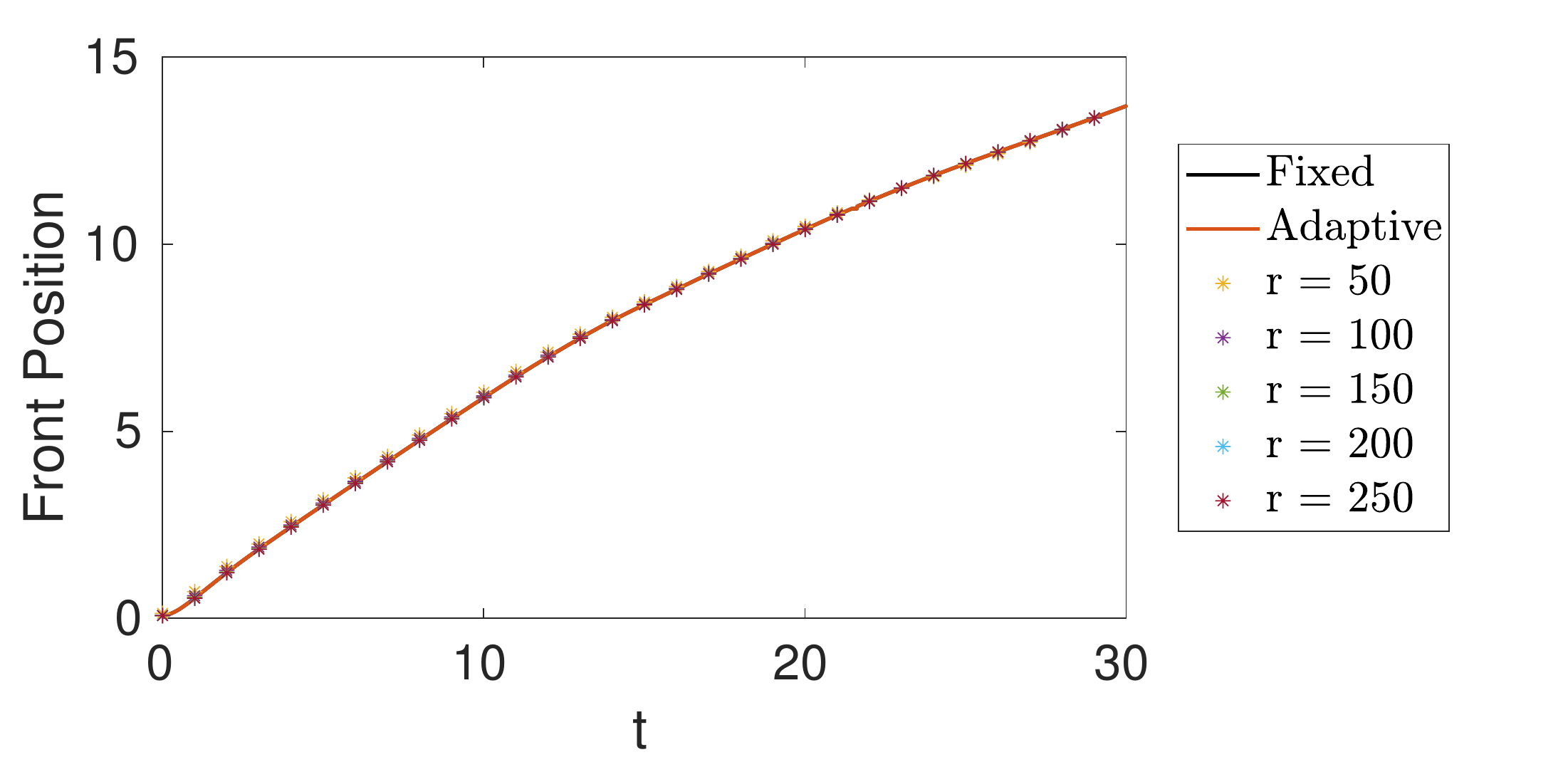}}\\
		\subfigure[Mass conservation]{\label{fig:le_adapt_sol}\includegraphics[width=0.6\linewidth]{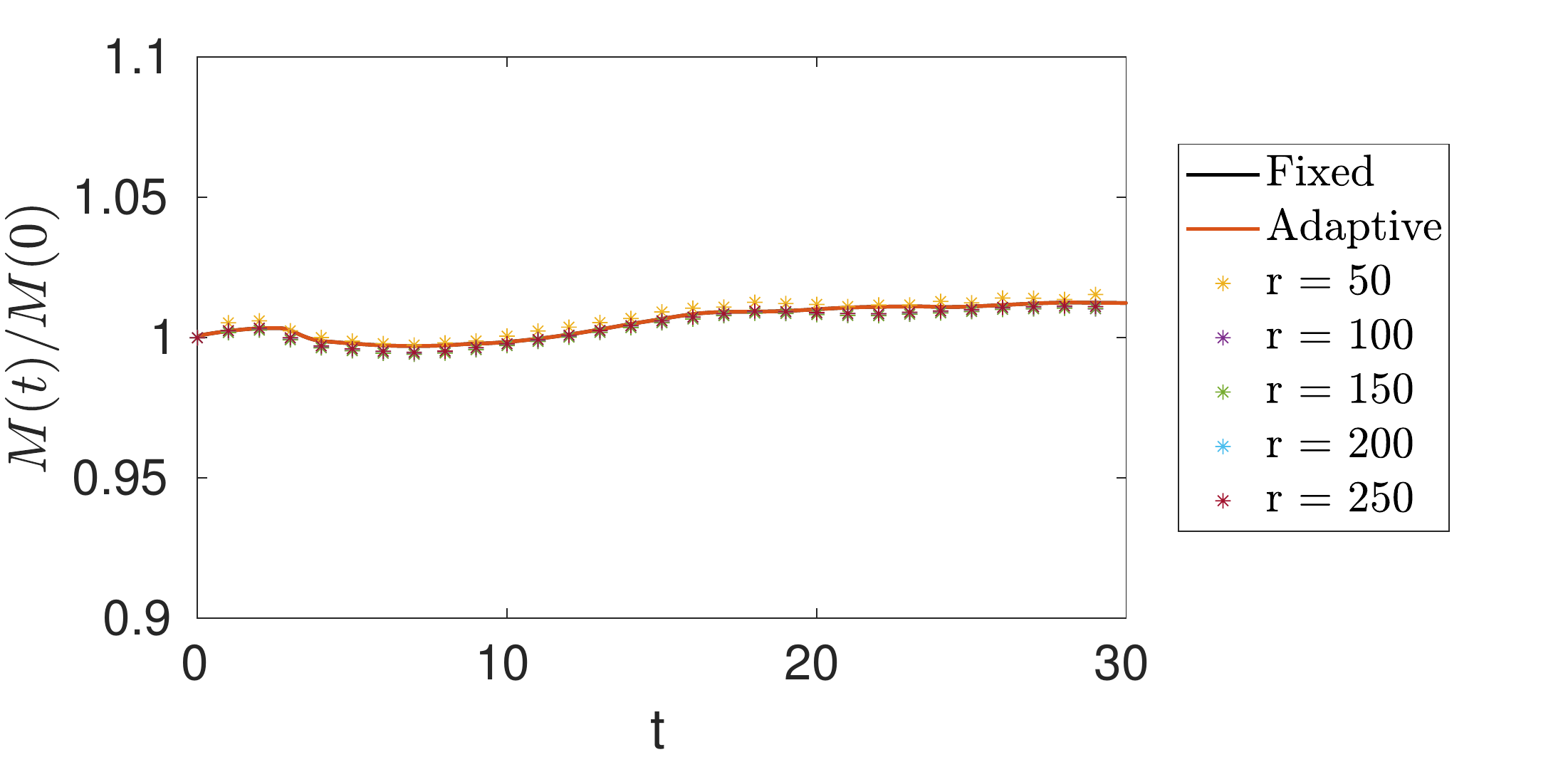}}
		\vspace{12pt}
		\caption{Front position and mass conservation for the fixed mesh and adaptive mesh simulations and reconstructions with the target mesh.}
		\label{fig:le_qoi_reconst}
	\end{center}
\end{figure}

 
To solve the governing equations, we implement the RBVMS formulation \cite{Guerra2013} for Eq. (\ref{eq:LE}) using the FEniCS 2019.1 \cite{AlnaesBlechta2015a} framework to generate the snapshots for this example. The adaptive mesh refinement procedure returns the solution for $\mathbf{u}, p,$ and $\phi$. For this example, we only consider the snapshots of $\phi$ for our calculations, that is, an overall data reduction of  $75\%$. Details of the formulation of the problem can be found in \cite{Guerra2013, Barros2020}. We consider a fixed mesh simulation with $701\times101$ nodes and $700 \times 100$ cells, where each cell is divided into two linear triangles. We consider an interface-tracking adaptive mesh error indicator that flags and refines the mesh where the two fluids interact for the adaptive mesh simulation. The error indicator for the mesh refinement is $|\nabla\phi|$ being larger than a given tolerance. For this purpose, a mesh containing $175 \times 25$ cells is considered, the interfaces are refined considering two levels of refinement, and the mesh refinement is invoked at every time step. Figure \ref{fig:le} shows the results at $t = 10$s for both fixed and adaptive mesh simulations and the projection of the adaptive solution onto the same mesh used in the fixed mesh simulation. 
AMR/C in this problem is advisable since the dynamics are predominant on the interface between the fluids. Most of the domain is not affected in the early stages of the simulation, and the use of fine meshes outside these regions may represent unnecessary computational effort. Figure \ref{fig:lockexchange_dofs} shows the number of nodes in the mesh during the simulation time for the adaptive mesh compared with the fixed $701 \times 101$ nodes mesh.  
We observe that the number of nodes in the adaptive simulation is smaller than the fixed mesh for all the simulation time. Table \ref{tab:projection_LE} shows the time required for the simulation to run in comparison with the time spent on projecting the solutions onto the reference target mesh. 
\begin{table}[ht!]
    \centering
    \begin{tabular}{|c|c|c|}
    \hline
        Code Part               &   Absolute Time (s)   &   Relative Time ($\%$) \\
        \hline
        AMR/C FEM               &   $15748.11$          &   $94.94$ \\
        \hline
        Mesh Projection         &   $839.42$            &   $5.06$ \\
        \hline 
    \end{tabular}
    \caption{Absolute and relative time required for the projection routine in comparison with the adaptive finite element simulation code (AMR/C FEM) for the lock-exchange example.}
    \label{tab:projection_LE}
\end{table}
Now we proceed applying DMD to the projected solution and reconstructing the solutions. Figure \ref{fig:le_rel_error} shows the relative error for the reconstruction using different values of the rank $r$. The results are also confirmed in Table \ref{tab:efficiency_le}, where the relative error between the reconstructed and projected snapshot matrices is shown and the speedup computed for each case. We observe that, for increasing values of $r$, the relative error decreases for all steps and affects the overall relative error of the matrices. That is, inserting more structures in the DMD basis yields better accuracy in terms of overall relative error. As for the speedup, we notice that it decreases for increasing values of $r$. Such a decrease occurs because larger values of $r$ directly affect the rSVD algorithm performance \cite{Barros2020, Erichson2019} and increase the dimensions of the matrices for the computation of the DMD basis.
\begin{table}[ht!]
	\centering
	\caption{Relative error between reconstructed data and the snapshots and speedup between DMD and the numerical simulation.}
	\begin{tabular}{|c|c|c|}
		\hline
		Rank $r$ & Relative Error ($\eta_F$) & Speedup \\
		\hline
		$50$ & $1.081 \times 10^{-1}$ & $558.53$   \\
		\hline
		$100$ & $3.472 \times 10^{-2}$ & $445.91$  \\
		\hline
		$150$ & $6.207 \times 10^{-3}$ & $230.67$  \\
		\hline
		$200$ & $2.024 \times 10^{-3}$ & $204.74$ \\
		\hline
		$250$ & $9.604 \times 10^{-4}$ & $189.17$ \\
		\hline
	\end{tabular}
	\label{tab:efficiency_le}
\end{table}
Most importantly, we can evaluate the quantities of interest common to density-driven gravity currents. Figure \ref{fig:le_qoi_reconst} shows the front position and mass conservation for both simulations and the respective reconstructions. We note that, for these quantities all reconstructions are in extremely good agreement with those computed with the fixed and adaptive meshes, approaching them, as expected, for higher values $r$.  

\subsubsection{Bubble rising problem}

\begin{figure}[ht!]
\begin{center}
\includegraphics[width=0.3\textwidth]{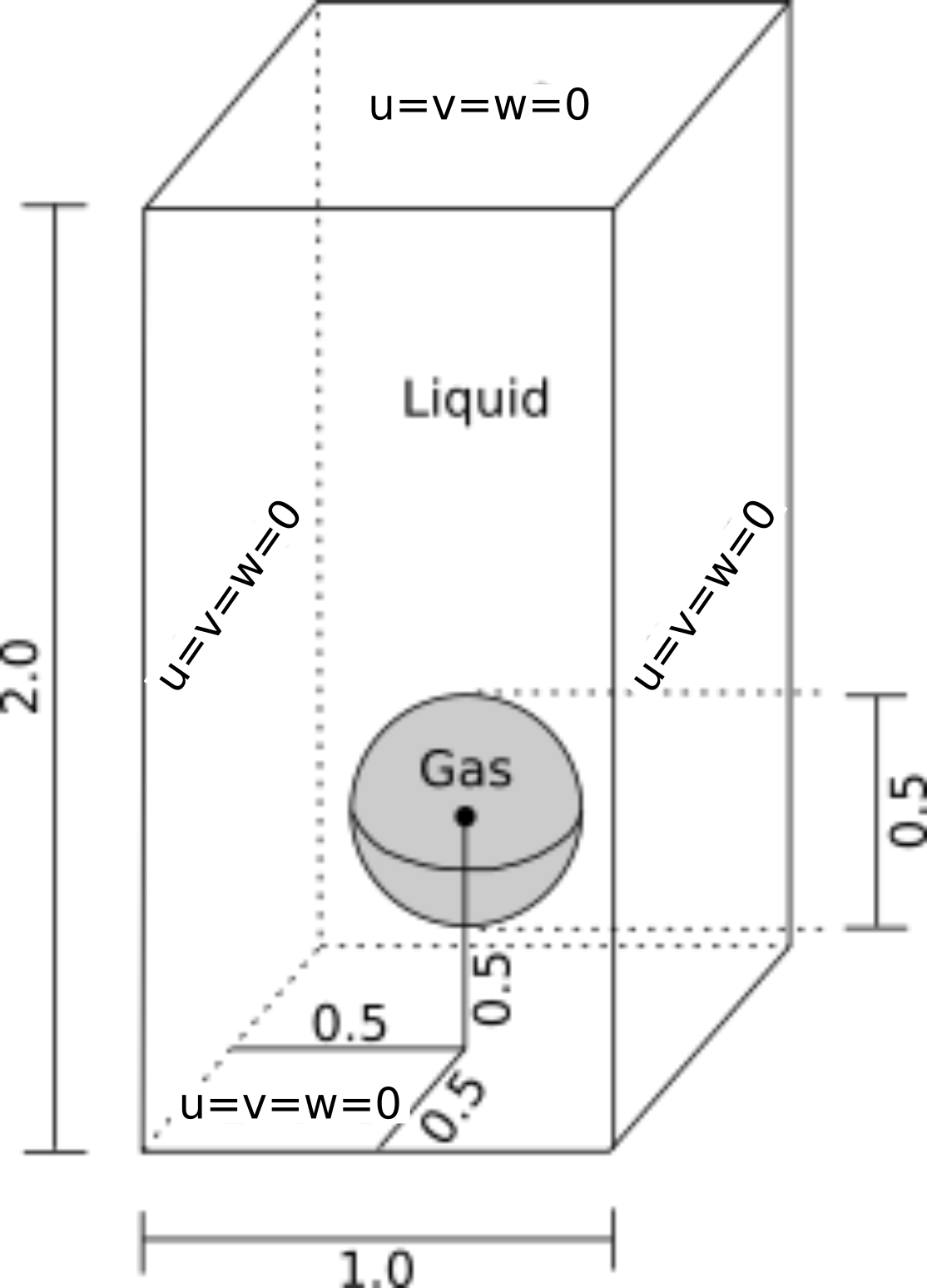}
\end{center}
\caption{Initial configuration and boundary conditions for the bubble rising problem.}
\label{bubble_bench3d}
\end{figure}

We now study a bubble rising 3D benchmark, whose task is to track the evolution of a three-dimensional bubble rising in a liquid column. The initial configuration is described in Figure \ref{bubble_bench3d}.

For this problem, we couple the Navier-Stokes equations with an interface capturing method called convected level-set \cite{coupez2007convection,ville2011,grave2020new}. The convected level-set is a  method used to represent the interface between two phases and, by a convection equation, to move the interface as the flow evolves. A force that has an important role in bubble problems is the surface tension $\mathbf{F_{st}}$, which is applied using the Continuum Surface Model (CSF) \cite{brackbill1992continuum}.

We write the governing equations in their dimensional form as


\vspace{6pt}
\begin{center}
	\begin{equation}
	\begin{aligned}
	\nabla\cdot \mathbf{u} = 0,\\
	\rho\frac{\partial \mathbf{u}}{\partial t} +\rho\mathbf{u} \cdotp \nabla \mathbf{u} + \nabla p - \mu \nabla^2\mathbf{u} - \rho\mathbf{g} - \mathbf{F_{st}} = \mathbf{0},\\
	\frac{\partial \alpha}{\partial t} + (\mathbf{u} + \lambda \mathbf{U})\cdot \nabla \alpha - \lambda \sgn(\alpha) S = 0,
	\end{aligned}
	\label{level-set}
	\end{equation}
\end{center}
\vspace{6pt}

\noindent where $\rho$ is the density, $\mu$ is the dynamic viscosity, $\mathbf{g}$ is the acceleration of gravity vector, $\alpha$ is the level-set function, $\lambda$ is a penalty constant, $\mathbf{U} = \sgn(\alpha) \dfrac{\nabla \alpha}{||\nabla \alpha||}$, and $S$ a function related to the level-set signed distance function. 

For the temporal integration, we apply the Backward-Euler method to the Navier-Stokes equations, while for the convected level-set, we use the BDF2 method. One may find more details about the governing equations and methods in \cite{grave2020new}.

The initial configuration consists of a spherical bubble of radius $R = 0.25$ m centered at $[0.5, 0.5,0.5]$ m in a $[1 \times 1 \times 2]$ m domain. The no-slip boundary condition is applied to all boundaries. Table \ref{tab:bubble} lists the parameters used for this simulation. 

\begin{table}[ht!]
  \begin{center}
  \caption{Rising bubble data.}
   \label{tab:bubble}
    \begin{tabular}{c c c} 
      \hline
Computational domain & $1\times 1 \times 2$ & (m)
\\
Grid sizes & $0.100$ to $0.025$ & (m)
\\
Number of time steps & $240$ & (-)
\\
Time step & $0.0125$ & s
\\
Bubble radius & $0.25$ & m
\\
Initial bubble position & $(x,y) = (0.5, 0.5, 0.5)$ & m
\\
Liquid density & $1000$ & kg/$\text{m}^3$
\\
Liquid viscosity & $10$  & kg/(ms)
\\
Gas density &  $100$ & kg/$\text{m}^3$
\\
Gas viscosity & $1$ & kg/(ms)
\\
Surface tension &  $24.5$  & N/m
\\
Gravity &  $0.98$ &  m/$\text{s}^2$
\\
     \hline
    \end{tabular}
  \end{center}
  \end{table}

\begin{figure}[!ht]
	\begin{center}
		\subfigure[Adaptive mesh solution]{\label{fig:bubble_adaptive}\includegraphics[width = 0.49\linewidth]{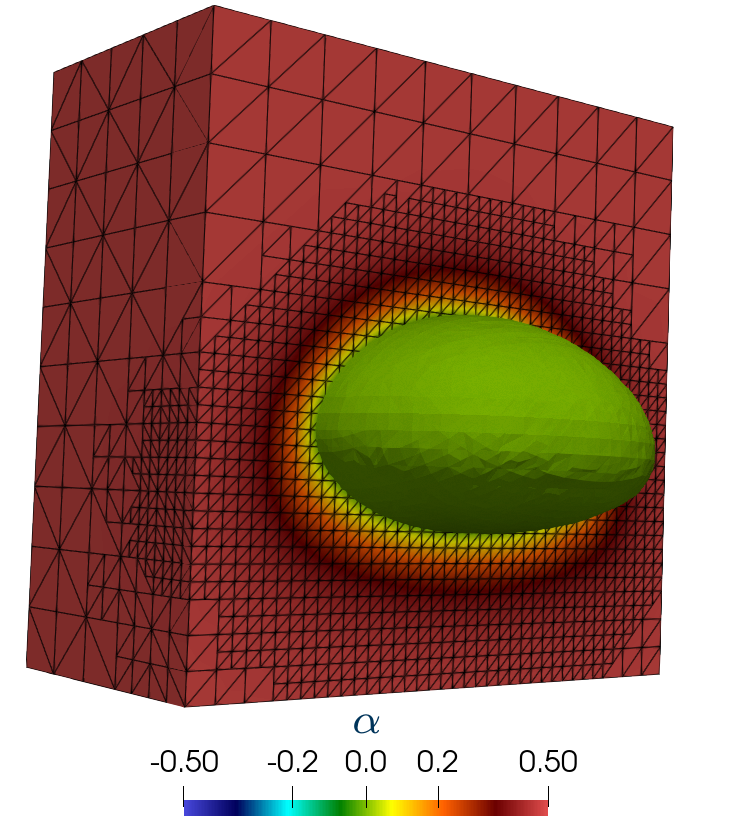}}
		\subfigure[Coarse mesh projection]{\label{fig:bubble_supercoarse}\includegraphics[width = 0.49\linewidth]{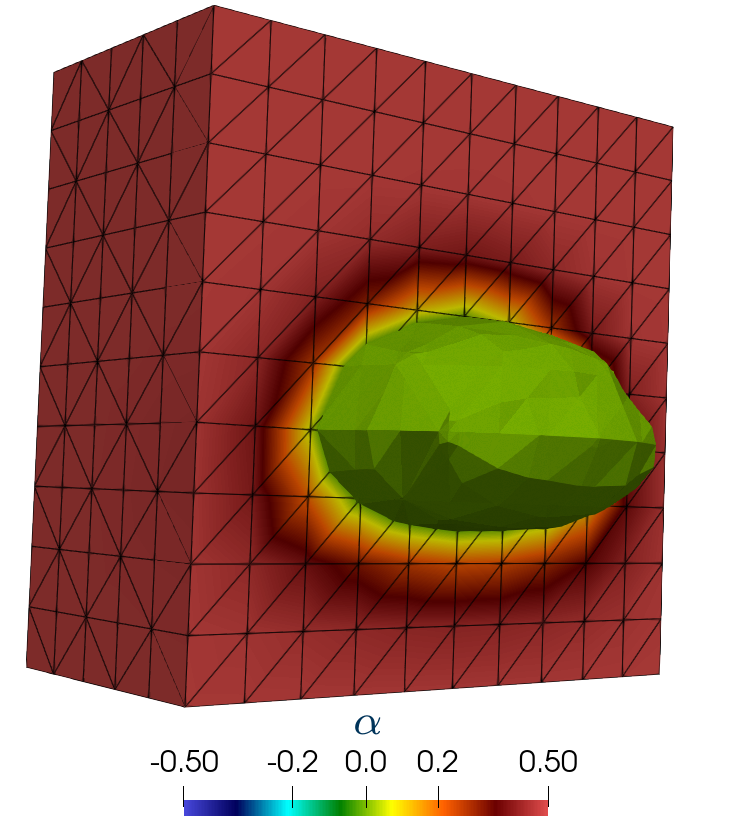}}
		\subfigure[Intermediate mesh projection]{\label{fig:bubble_coarse}\includegraphics[width = 0.49\linewidth]{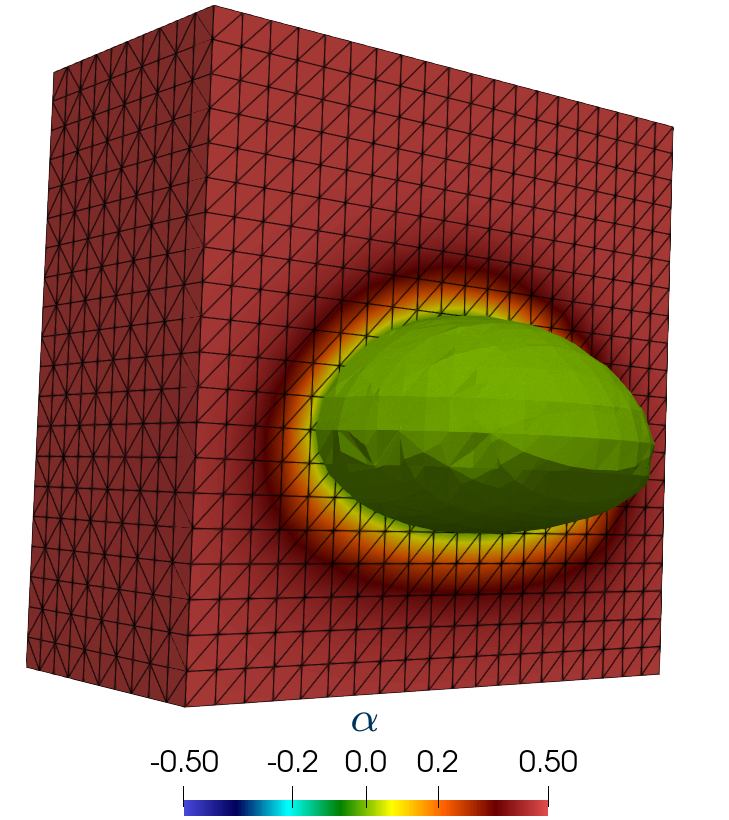}}	
		\subfigure[Fine mesh projection]{\label{fig:bubble_medium}\includegraphics[width = 0.49\linewidth]{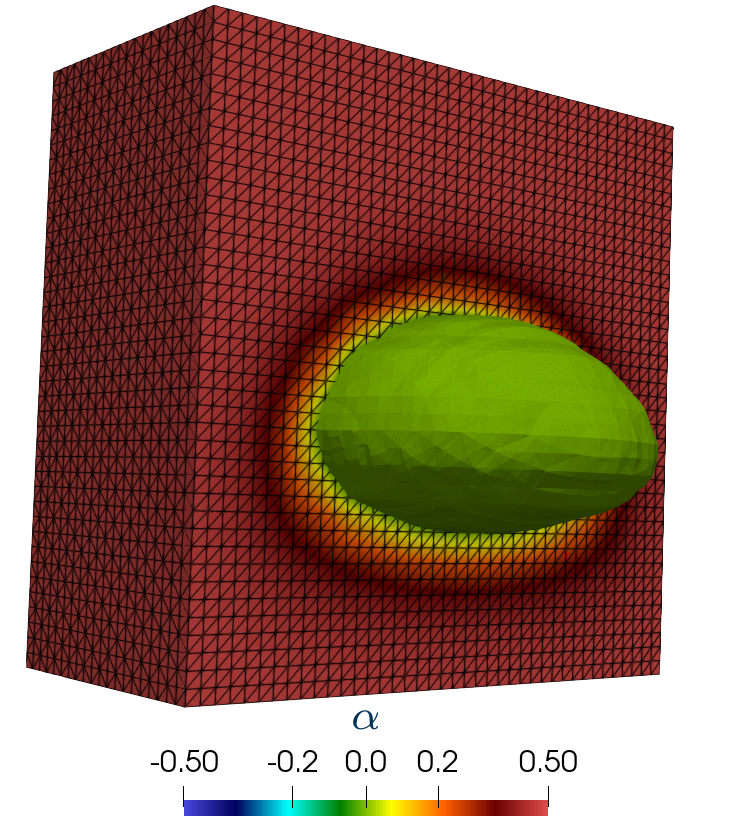}}	
		\vspace{12pt}
		\caption{Level-set solution detail at $t = 3.0$s and projection to the coarse, medium and fine meshes.}
		\label{fig:bubble}
	\end{center}
\end{figure}
\begin{figure}[ht!]
  \centering
  \subfigure[Volume.]{\includegraphics[width=0.49\linewidth]{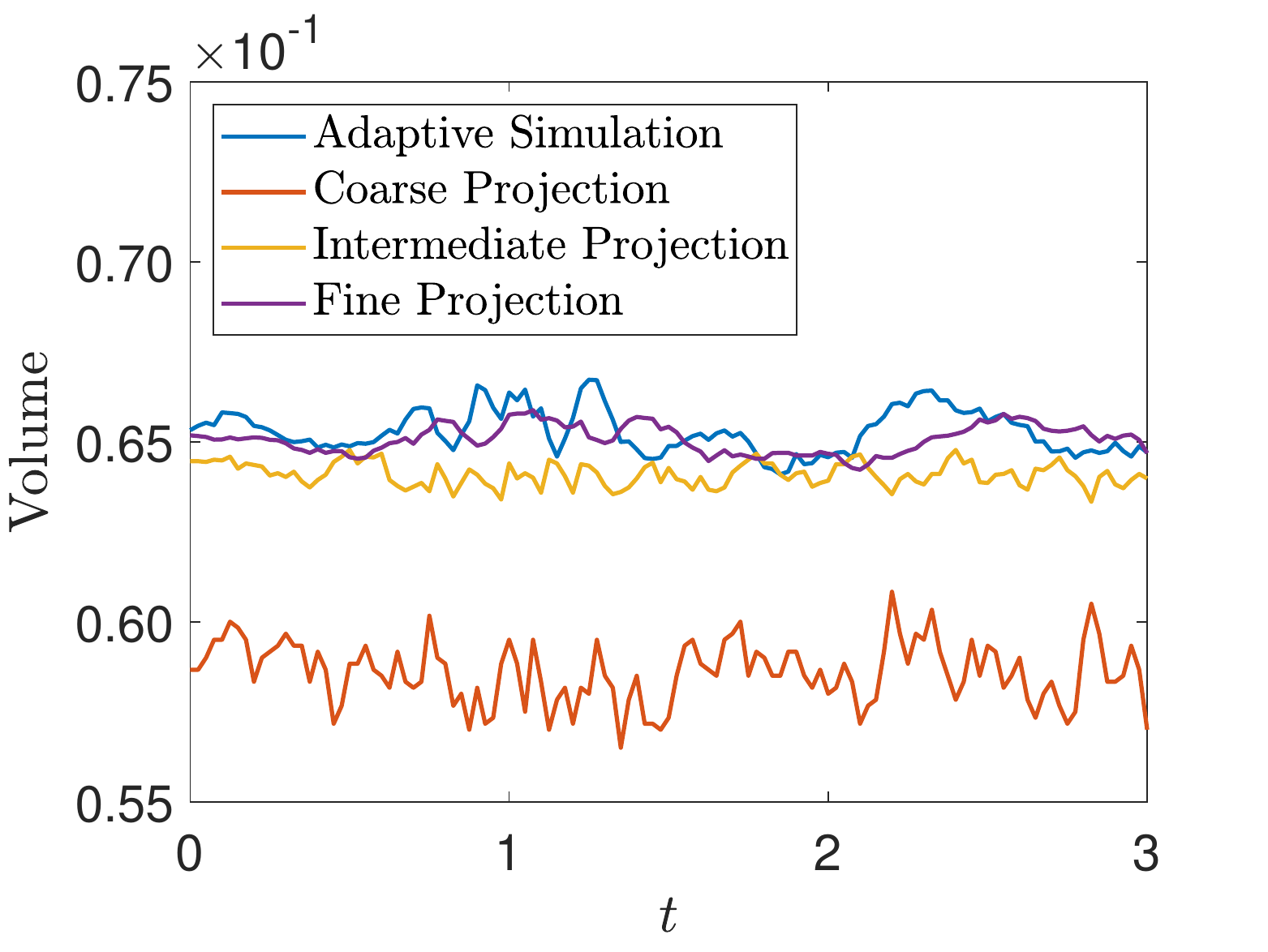}\label{volume1-3d}}
    \hfill
  \subfigure[Sphericity.]{\includegraphics[width=0.49\linewidth]{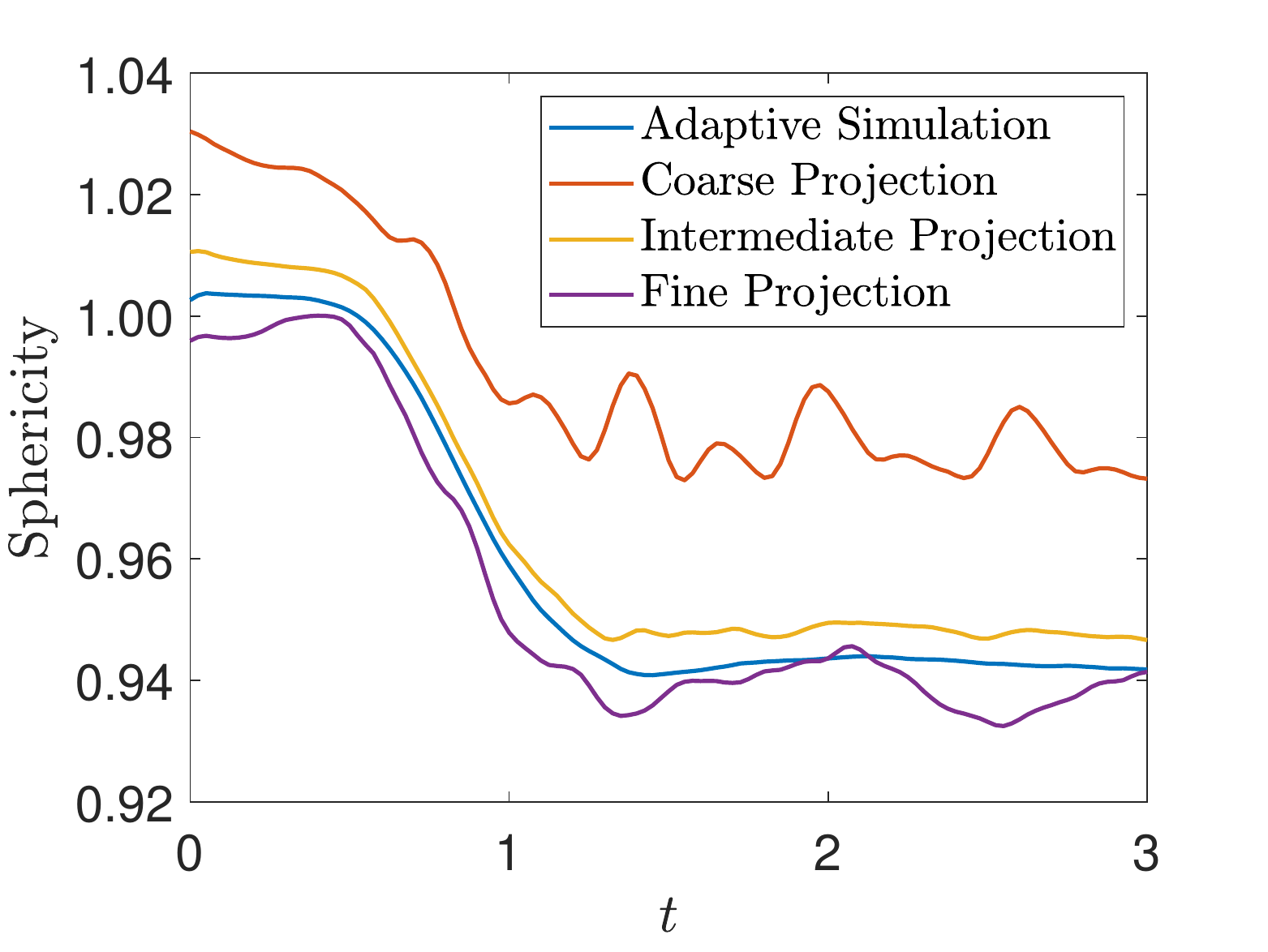}\label{circularity1-3d}}
    \hfill
  \subfigure[Center of mass.]{\includegraphics[width=0.49\linewidth]{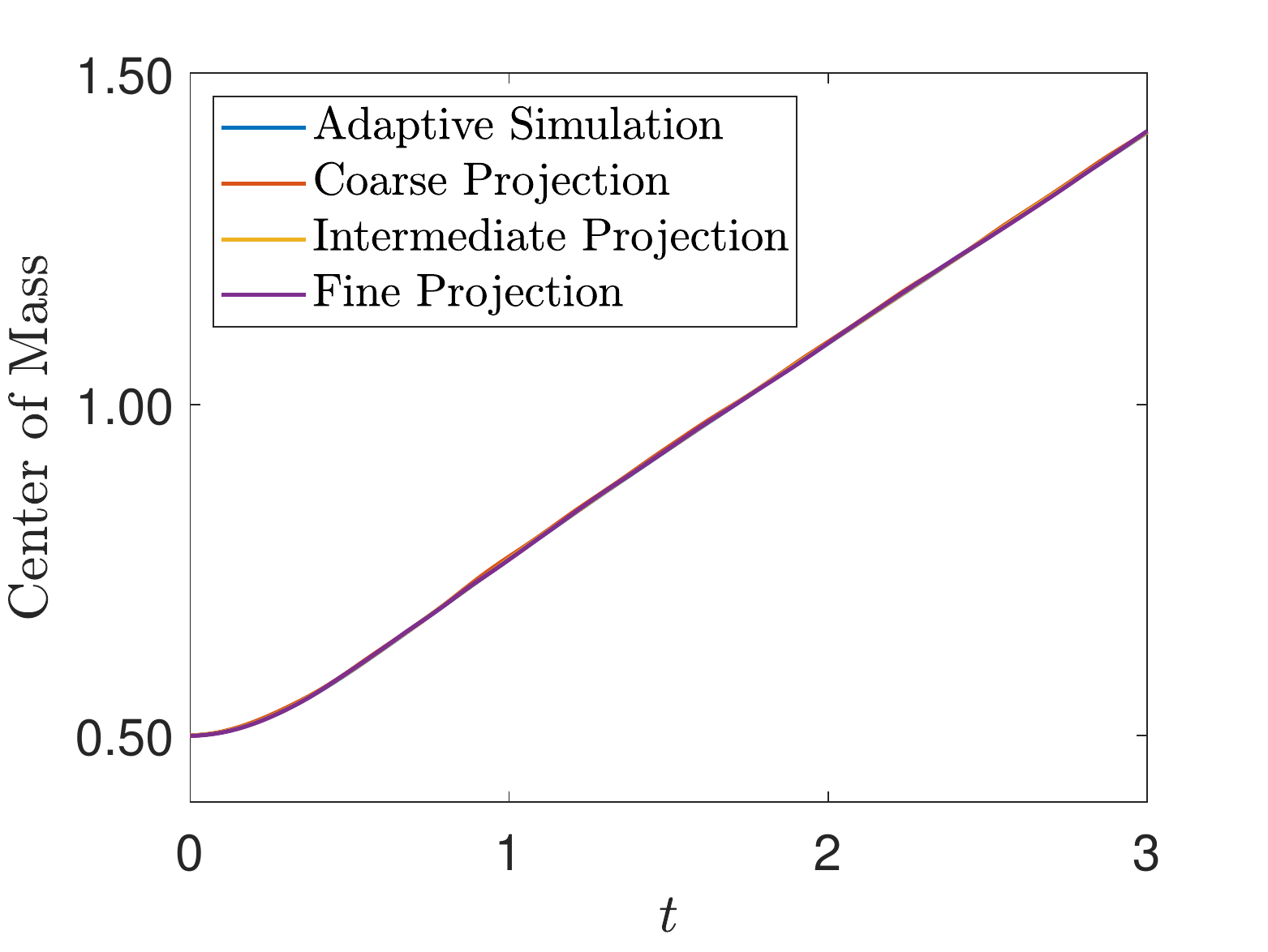}\label{mass1-3d}}
    \hfill
  \subfigure[Rise velocity.]{\includegraphics[width=0.49\linewidth]{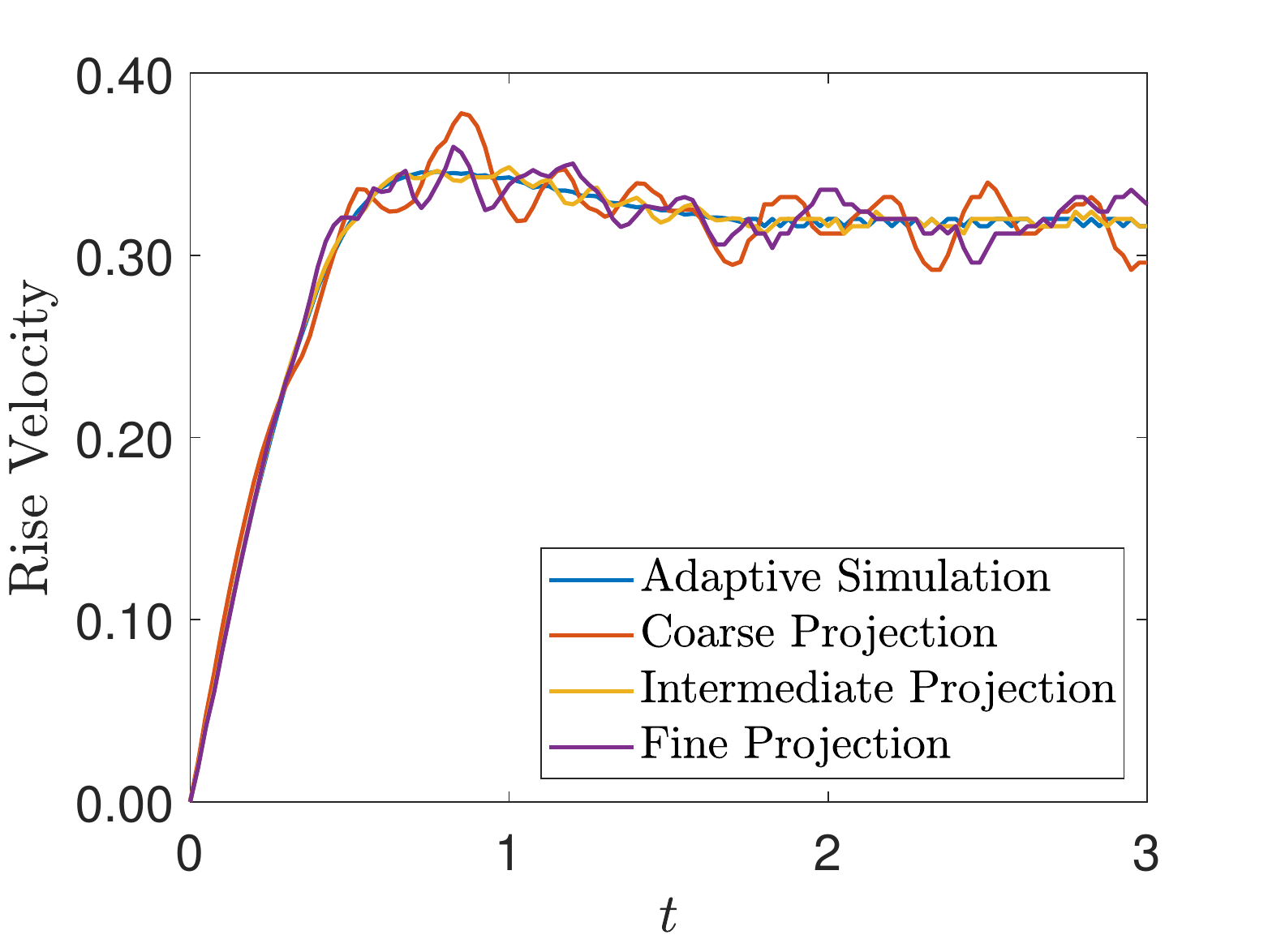}\label{risevel1-3d}}
  \caption{Comparison between the simulation and projection of the 3D rising bubble quantities of interest.}
  \label{projection_bubble}
\end{figure}

We use an adapted mesh, initially with $10 \times 10 \times 20$ cells, with each cell divided into $6$ linear tetrahedra.  We refine the initial region where the bubble is located into two levels and, after the refinement, the smallest element has a size of $0.025$ m. The adaptive mesh refinement is based on the flux jump of the level-set function error, in which $h_{max}= 2$. We apply the adaptive mesh refinement every four time steps. The interface is modeled with $E = 0.05$, and the time step size is defined as $\Delta t = 0.0125$s. We output the projected solutions at every $2$ time steps such that $\Delta t_o = 0.025$s. In this example, we consider three tetrahedral meshes for our projection strategy, presented in Figure \ref{fig:bubble}. The three meshes named coarse, intermediate and fine, present characteristic lengths similar to the three scales existent in the refinement levels of the adaptive mesh simulation. 
The coarse mesh represents the initial mesh on the adaptive simulation, with 12000 elements and 2541 nodes. The intermediate mesh presents smaller elements equivalent to the generated elements after the first refinement level on the AMR/C simulation, totalizing 96000 elements and 18081 nodes and the fine mesh contains the smallest scales presented on the adaptive numerical solution with 768000 elements and 136161 nodes. The figures show the level-set solution on half of the domain and at $t = 3$s for the adaptive mesh solution and the respective projections. We also present on Table \ref{tab:projection_bubble} the time required for the projection of the solutions of $\alpha$ on the three meshes in terms of absolute and relative time. We can see that the projection time is very small. 
\begin{table}[ht!]
    \centering
    \begin{tabular}{|c|c|c|c|}
    \hline
        Target Mesh                     & Code Part             &   Absolute Time (s)   &   Relative Time ($\%$)  \\
        \hline
        \multirow{2}{*}{Coarse}         & AMR/C FEM   &   $17345.92$          &   $99.97$ \\
                                        & Mesh Projection       &   $4.41$              &   $0.03$ \\
        \hline 
        \multirow{2}{*}{Intermediate}   & AMR/C FEM   &   $17390.03$          &   $99.85$ \\
                                        & Mesh Projection       &   $24.56$             &   $0.14$ \\
        \hline 
        \multirow{2}{*}{Fine}           & AMR/C FEM   &   $17653.70$          &   $98.91$ \\
                                        & Mesh Projection       &   $194.87$            &   $1.09$ \\
        \hline 
    \end{tabular}
    \caption{Absolute and relative time required for the projection routine in comparison with the adaptive finite element simulation code (AMR/C FEM) for the bubble rising example.}
    \label{tab:projection_bubble}
\end{table}
To verify if the bubble geometry and dynamics are being preserved during the mesh projections, we evaluate the quantities of interest such as the bubble volume, sphericity, center of mass, and rise velocity. Bubble volume and sphericity are related to the bubble geometry, while the position of the bubble center of mass and rise velocity regard the bubble dynamics. The results are compared for the AMR/C simulation output and the projections on Figure \ref{projection_bubble}. The quantities of interest are computed using the adaptive snapshots as well as the projected snapshots and compared with each other. Regarding the geometry quantities of interest, we observe that the coarse mesh projection solution affects the bubble's geometry, leading to bad results regarding the bubble volume and sphericity compared with the AMR/C simulation. For the intermediate and the fine meshes, the bubble's geometry is not largely affected, and the results are compatible with the AMR/C results. As for the quantities of interest related to the bubble dynamics, no significant difference is observed on the projection of the three meshes. We observe that the center of mass is not affected by the projections compared with the simulation results. As for the rise velocity, we observe some minor differences in all projection cases. 
\par
\begin{figure}
    \centering
    \includegraphics[width=0.55\linewidth]{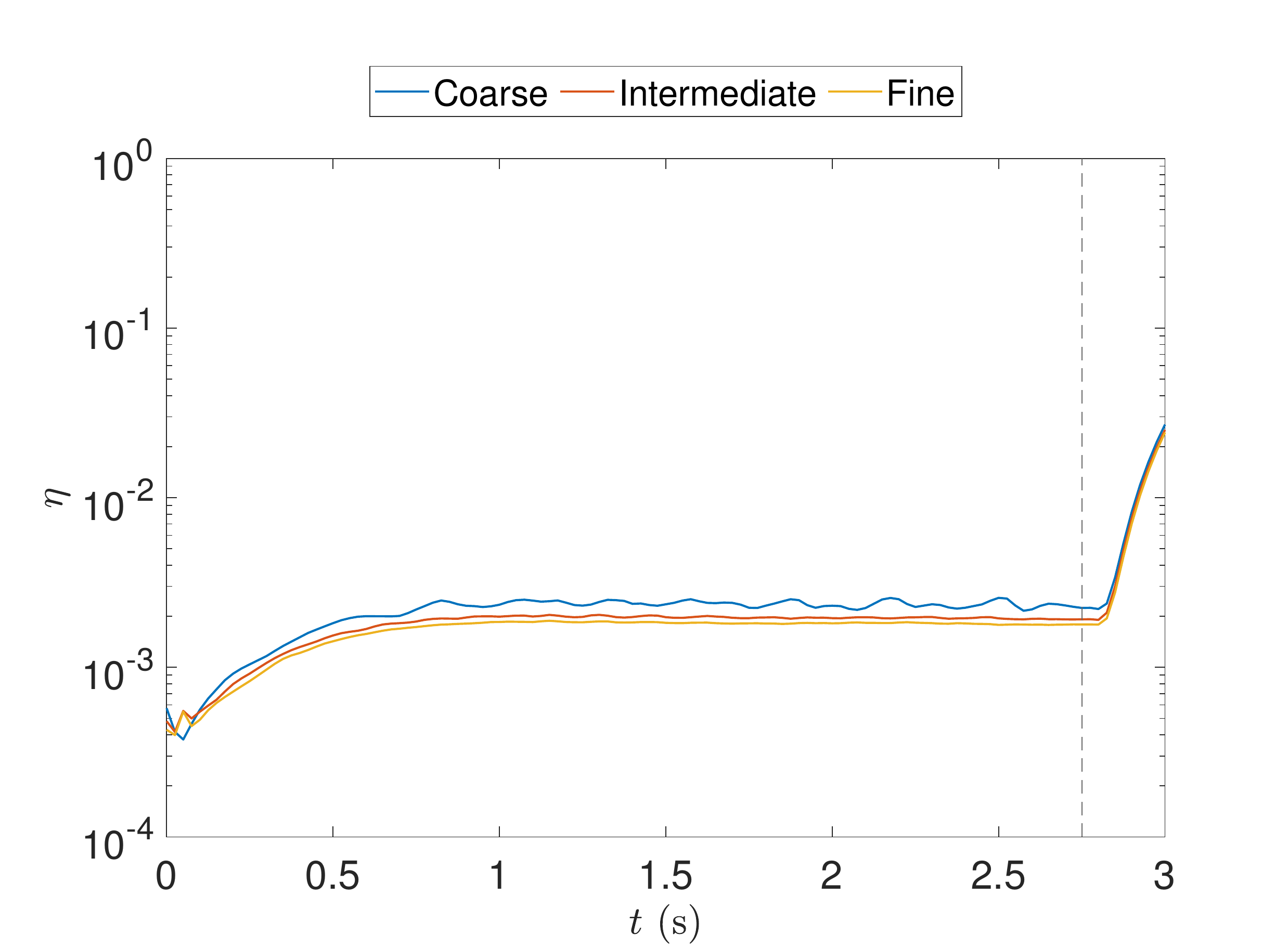}
    \caption{Relative error for the rising bubble example for the coarse, intermediate, and fine mesh solutions. The dashed line defines the start of the prediction phase.}
    \label{fig:bubble_rel_error}
\end{figure}

\begin{figure}[ht!]
  \centering
  \subfigure[$t = 2.75$s.]{\includegraphics[width=0.49\linewidth]{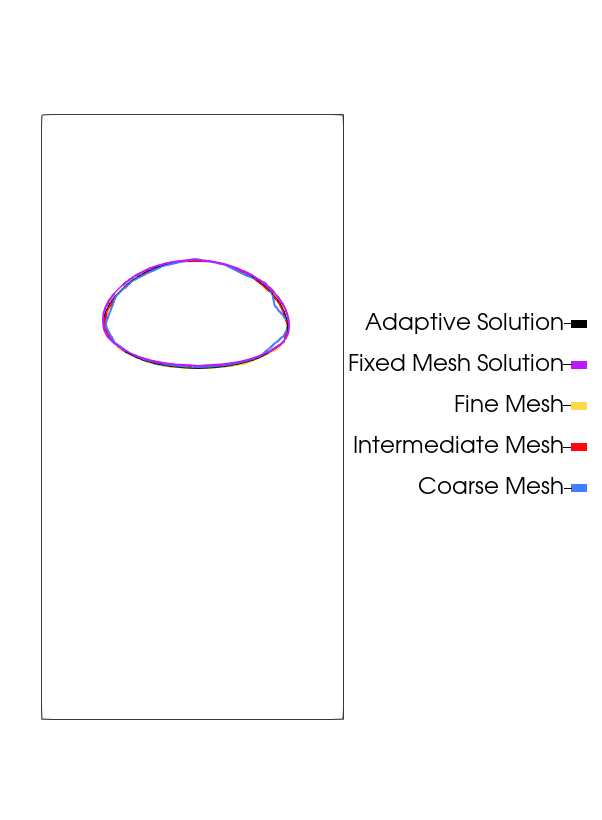}}
    \hfill
  \subfigure[$t = 3.00$s.]{\includegraphics[width=0.49\linewidth]{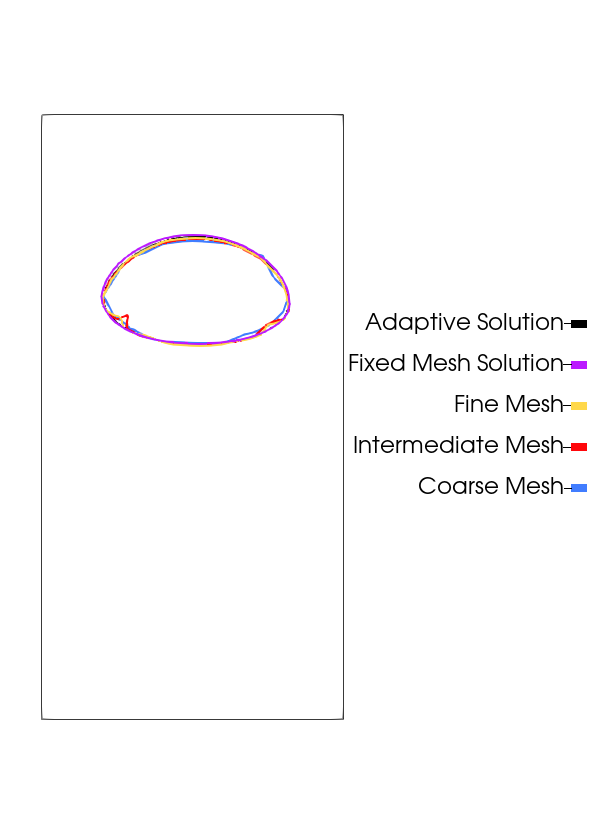}}
  \caption{Bubble contour at the vertical mid plane for the signal reconstruction ($t=2.75$s) and prediction ($t=3.0$s) last steps.}
  \label{fig:bubble_contour}
\end{figure}

\begin{figure}[ht!]
  \centering
  \subfigure[Volume.]{\includegraphics[width=0.49\linewidth]{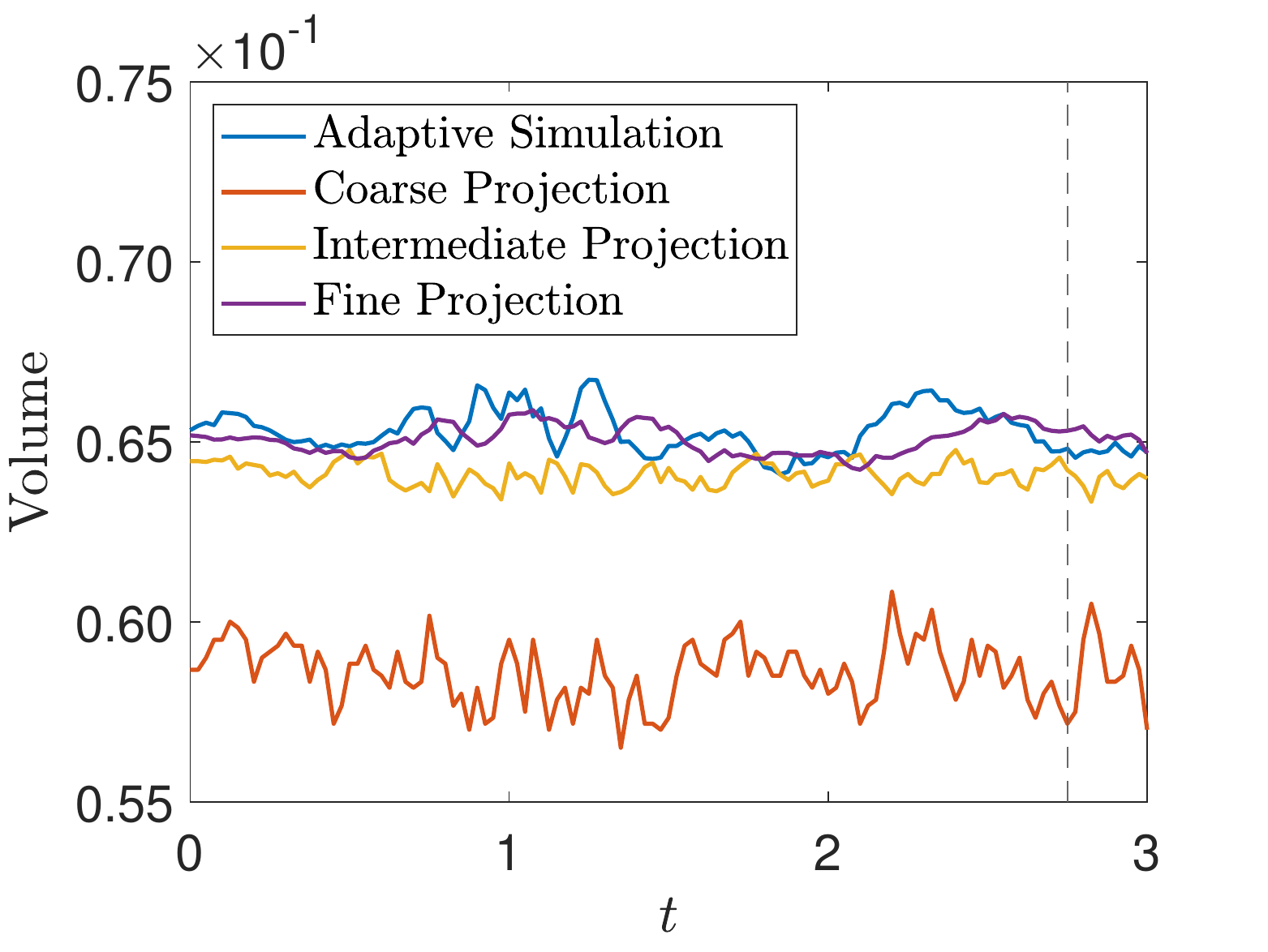}\label{volume2-3d}}
    \hfill
  \subfigure[Sphericity.]{\includegraphics[width=0.49\linewidth]{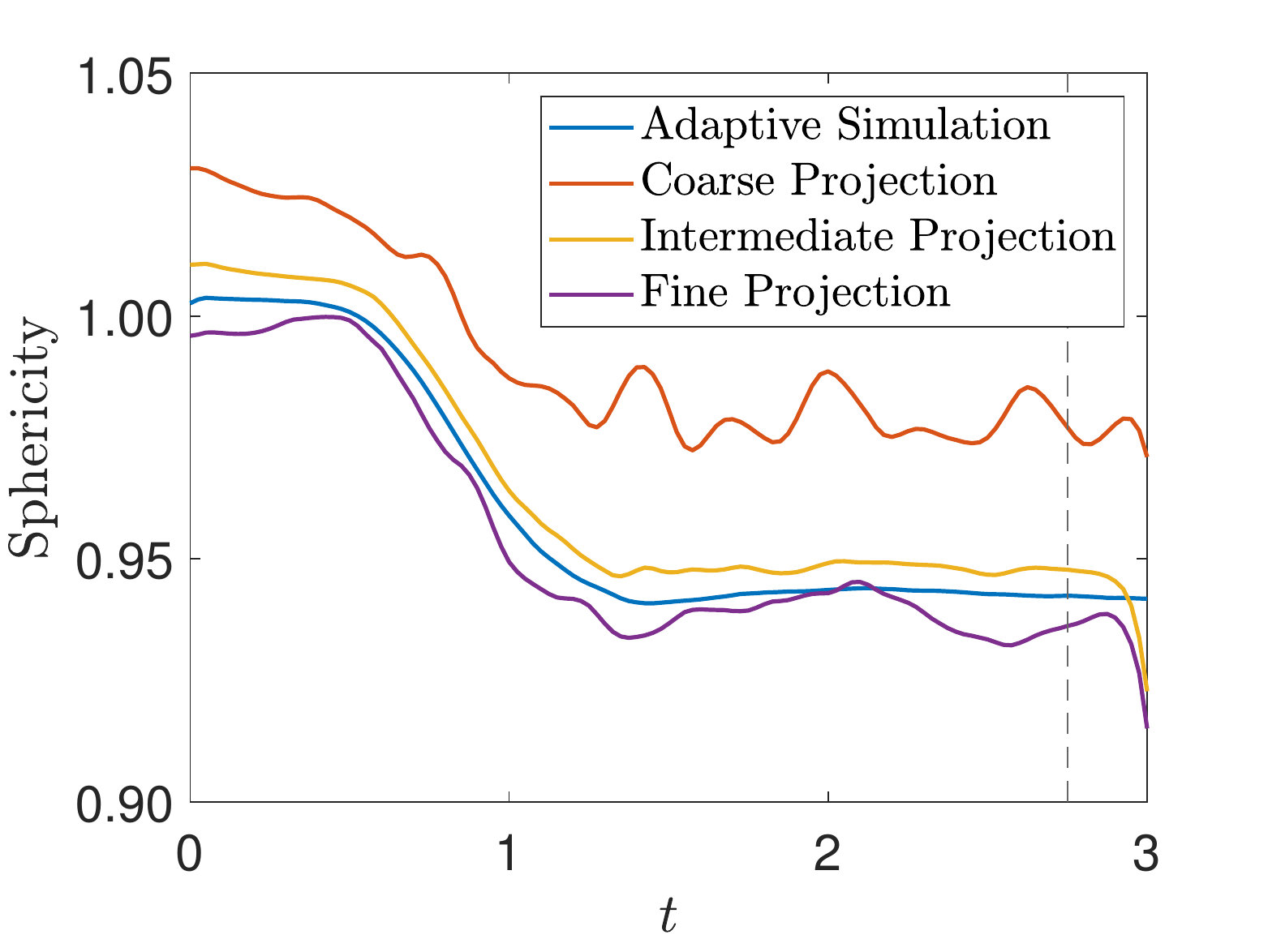}\label{circularity2-3d}}
    \hfill
  \subfigure[Center of mass.]{\includegraphics[width=0.49\linewidth]{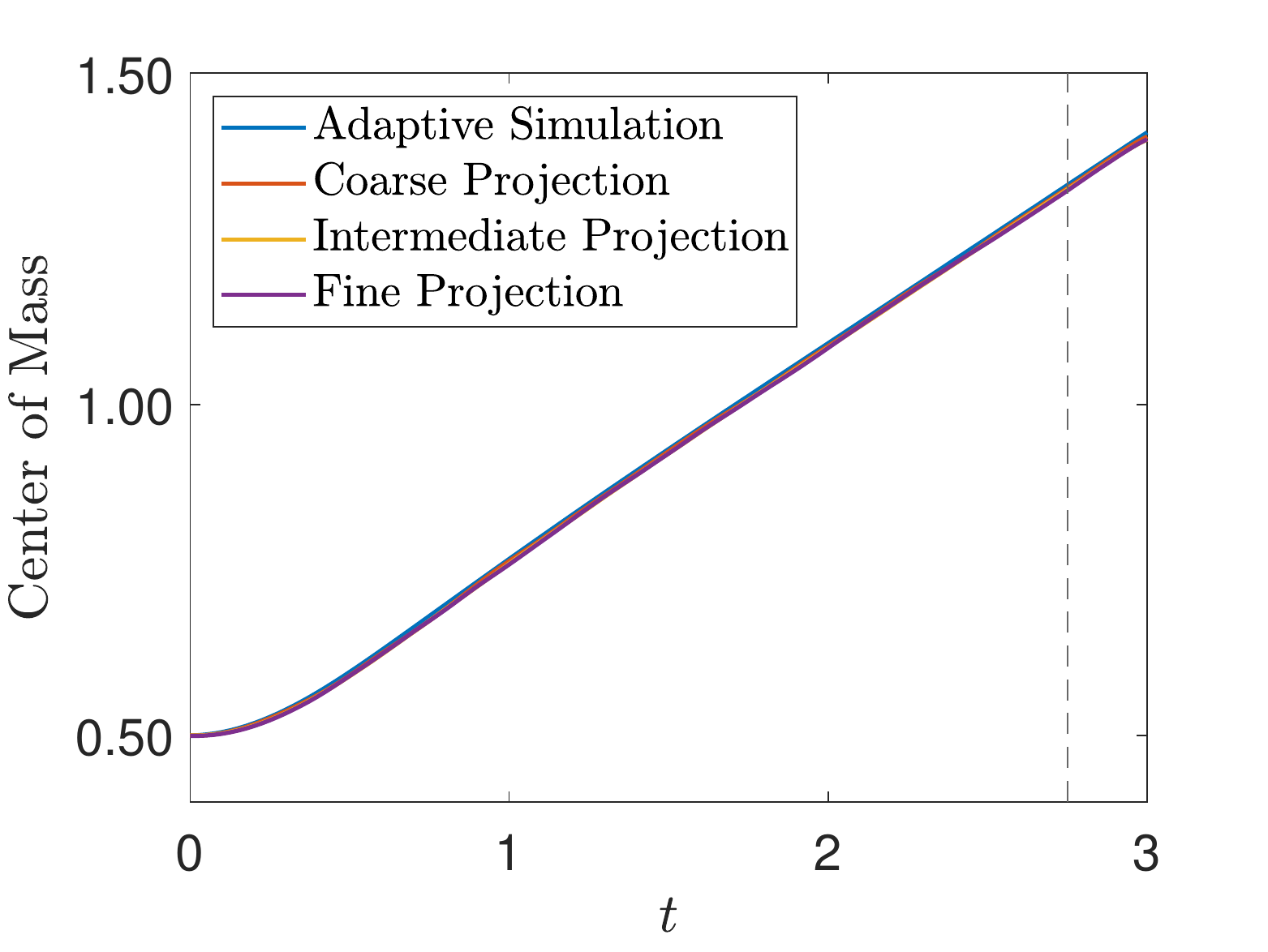}\label{mass2-3d}}
    \hfill
  \subfigure[Rise velocity.]{\includegraphics[width=0.49\linewidth]{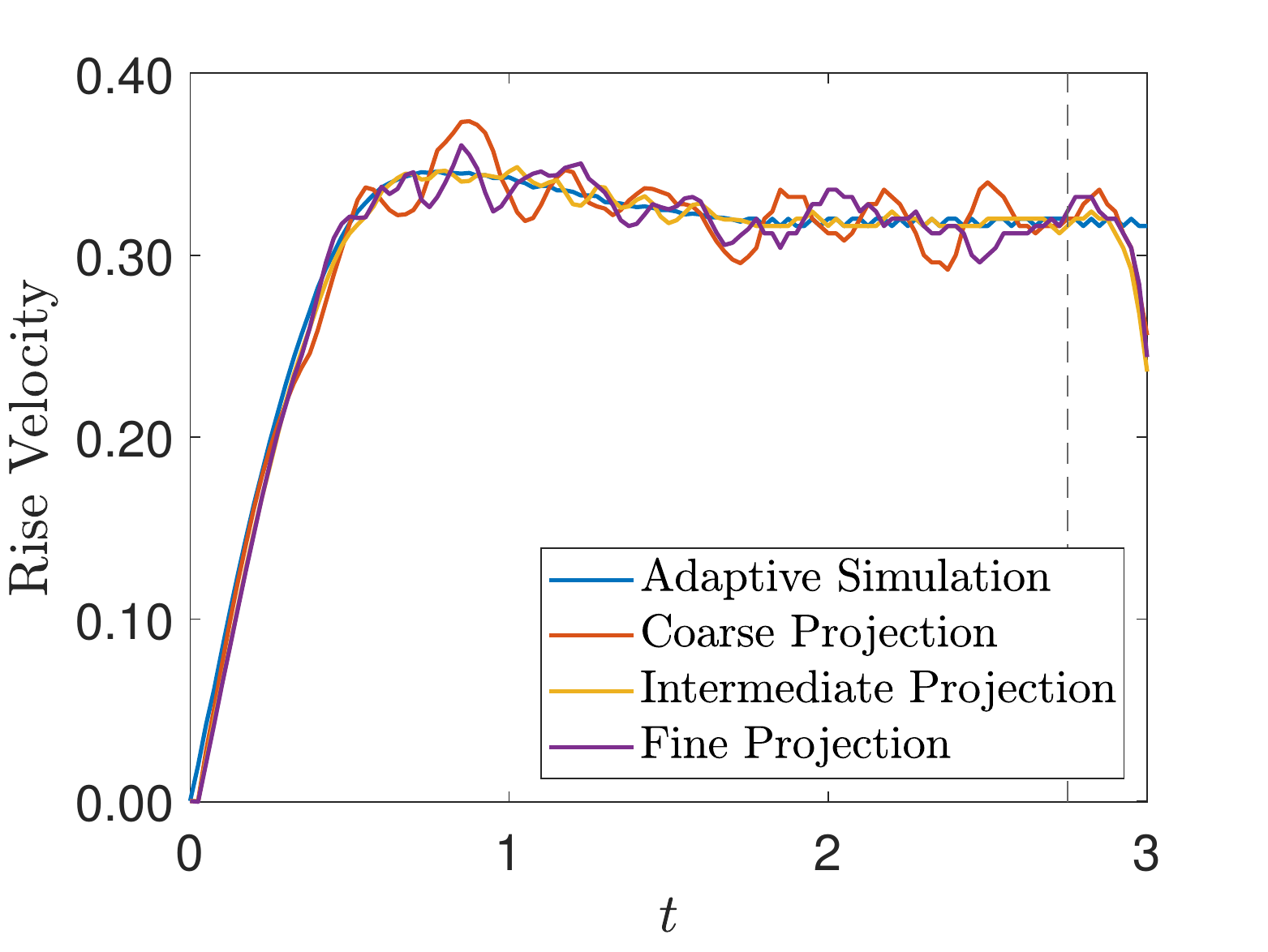}\label{risevel2-3d}}
  \caption{Comparison between the simulation and DMD signal plus prediction of the 3D rising bubble quantities of interest. The dashed line marks the beginning of the prediction regime for the DMD.}
  \label{DMD_bubble}
\end{figure}

 
The simulation proceeds until $T = 3.0$s, yielding a dataset containing $240$ snapshots regarding the solution of $\alpha$ for each target mesh from the projections. We do not consider the use of velocities and pressure in the DMD analysis, reducing the required data in $80\%$. We consider the results for the first $2.75$s to construct the basis and predict the last $0.25$ seconds.  We compare the DMD results in terms of relative error between the snapshot matrix and the obtained solutions for each case. The DMD solution for the coarse mesh is compared to the projected adaptive solution onto the coarse mesh and so forth. Results are evaluated for multiple values of $r$, such that $r = \{5,10,15,30,45,60\}$. The results regarding accuracy and performance for multiples values of $r$ and the three meshes are presented in Table \ref{tab:efficiency_bubble}.  
\begin{table}[ht!]
	\centering
	\caption{Relative error between reconstructed data and the projected snapshots and speedup between DMD and the numerical simulation. Results presented for multiple values of $r$.}
	\begin{tabular}{|c|c|c|c|}
		\hline
		Rank $r$ & Mesh & Rel. Error ($\eta_F$) & Speedup  \\
		\hline
		\multirow{3}{*}{5} 	& Coarse 		& $6.222\times 10 ^{-2}$ & $1.33 \times 10^5$ \\
							& Intermediate 	& $6.655\times 10 ^{-2}$ & $7.95 \times 10^4$ \\
							& Fine 			& $6.865\times 10 ^{-2}$ & $1.53 \times 10^4$ \\
		\hline
		\multirow{3}{*}{10} & Coarse 		& $2.384\times 10 ^{-2}$ & $1.21 \times 10^5$  \\
							& Intermediate 	& $2.461\times 10 ^{-2}$ & $3.60 \times 10^4$ \\
							& Fine 			& $2.490\times 10 ^{-2}$ & $1.22 \times 10^4$ \\
		\hline
		\multirow{3}{*}{15} & Coarse 		& $1.415\times 10 ^{-2}$ & $1.15 \times 10^5$ \\
							& Intermediate 	& $1.402\times 10 ^{-2}$ & $2.46 \times 10^4$ \\
							& Fine 			& $1.429\times 10 ^{-2}$ & $7.55 \times 10^3$ \\
		\hline
		\multirow{3}{*}{30} & Coarse 		& $8.900\times 10 ^{-3}$ & $1.09 \times 10^5$   \\
							& Intermediate 	& $8.125\times 10 ^{-3}$ & $3.29 \times 10^4$   \\
							& Fine 			& $7.687\times 10 ^{-3}$ & $6.66 \times 10^3$   \\
		\hline
		\multirow{3}{*}{45} & Coarse 		& $6.382\times 10 ^{-3}$ & $1.02 \times 10^5$   \\
							& Intermediate 	& $5.820\times 10 ^{-3}$ & $5.35 \times 10^4$   \\
							& Fine 			& $5.776\times 10 ^{-3}$ & $5.80 \times 10^3$   \\
		\hline
		\multirow{3}{*}{60} & Coarse 		& $5.659\times 10 ^{-3}$ & $8.74 \times 10^4$   \\
							& Intermediate 	& $5.444\times 10 ^{-3}$ & $1.45 \times 10^4$   \\
							& Fine 			& $5.518\times 10 ^{-3}$ & $5.53 \times 10^3$   \\
		\hline
	\end{tabular}
	\label{tab:efficiency_bubble}
\end{table}
The results for $r = 60$ are shown in Figure \ref{fig:bubble_rel_error}. We observe that the errors are stable for the reconstruction case, that is, the DMD solution before $t = 2.75$s. From that point, shown as a dashed line on the figure, the errors begin to grow exponentially for each predicted time step, while still remaining below 1\% until around 2.9 seconds. We observe that the errors in the reconstruction case are different regarding the mesh used. That is, the errors are larger with respect to the minimum characteristic length of the projection meshes. However, we observe that the errors grow at the same rate on the prediction phase independently of spatial discretization. 
We also show the results considering $r = 60$ for reconstruction at $t = 2.75$s and prediction at $t = 3.0$s in Figure \ref{fig:bubble_contour}. We compare the DMD results for the three projected meshes, the adaptive mesh solution, and a fixed mesh solution. The fixed mesh solution is obtained by running the simulation with a mesh of $768000$ elements and $136161$ nodes, as the fine mesh used in the projection.
We observe initially that the bubble geometry is better defined on the reconstruction than on the prediction figure. This better definition is directly related to the errors observed in Figure \ref{fig:bubble_rel_error}. We observe that the coarse mesh results do not capture the bubble geometry with the same accuracy as the intermediate and fine meshes for the reconstruction results. As for the intermediate and fine meshes, they present similar results in comparison with the projected solutions. However, when we observe the prediction figure, we observe that instabilities inherent to DMD arise on the bubble contour, affecting the bubble geometry for the intermediate and fine mesh.
We now proceed comparing the results in terms of quantities of interest for the DMD results. Figure \ref{DMD_bubble} shows the bubble volume, sphericity, center of mass, and rise velocity for the DMD results compared to the adaptive solution results. The same issue regarding sphericity on the coarse mesh projection is observed on the coarse mesh DMD results. However, for the intermediate and fine meshes, the values match the results observed for the projection. We observe results in conformity for the center of mass and rise velocity as well. In terms of prediction, we observe that DMD accurately predicts the volume and the center of mass evolution. As for the other quantities of interest, the increasing exponential errors in the DMD prediction structure affect the quantities of interest for long time future predictions. 



\section{Conclusions}
\label{sec:conclusions}
\par 
In this work, we propose a strategy to enable data-driven snapshot-based methods on finite element solutions obtained by AMR/C simulations. The use of AMR/C algorithms in finite element approximations of PDEs is known to reduce memory usage and to increase the efficiency of the simulations without compromising the accuracy. The simulation adapts the mesh with the evolution of the solution, refining regions of interest and coarsening regions that are not of interest. The adaptation process leads to different meshes during the simulations, and the solutions have different dimensions and topologies (or different connectivities and nodal coordinates) whenever the AMR/C algorithm is invoked. Snapshots with different dimensions prevent the use of snapshot-based algorithms such as DMD, where the snapshot matrix is built by stacking the solutions for each observation in columns. In this study, we considered a strategy to project the adaptive solutions on reference meshes such that all snapshots present the same dimensions and nodal indices, enabling the construction of snapshot matrices. The method employed to project the solutions onto the target reference mesh is the $\mathcal{L}^2-$projection, a simple, fast and versatile approach. The $\mathcal{L}^2-$projection is a common strategy present in several finite element libraries and frameworks and consists basically of solving a linear system, where the matrix is obtained by a self-adjoint operator enabling the use of efficient solvers. Despite presenting drawbacks, especially regarding properties' conservation, we investigate the use of the $\mathcal{L}^2-$projection as a postprocessing tool without any relevant issues. By postprocessing, we mean that the $\mathcal{L}^2-$projection algorithm is invoked only to insert the solutions on a reference function space for each time step and output the files. This strategy does not yield significant additional computational effort as we observe that the $\mathcal{L}^2-$projection routine required around $1\%$ of the computational time required for the adaptive finite element code to run in most cases and $5\%$ in the worst case. When the source code is not available to invoke the $\mathcal{L}^2-$projection, one can construct the solutions from the output files and project them onto the target reference mesh in a complete non-intrusive workflow.
\par 
We test the algorithm on several models presenting different dynamics and underlying physics. First, we present the results for DMD on a continuous SEIRD model for COVID-19 for fictional data in 1D and real data for Lombardy, Italy, in 2D. The idea of considering short-time future estimates on COVID-19 models could improve the decision-making of public policies to avoid further contamination, and the use of AMR/C in the simulations, coupled with the presented projection scheme and DMD, can lead to fast and reliable predictions. The simulations are implemented using the \texttt{libMesh} library using its refinement built-in functions. We compare the solution, its projection in the reference mesh, and the adaptive solution and notice that no relevant errors are found for the projection strategy. In this example, the $\mathcal{L}^2$-projection consists of $1.47\%$ and $1.48\%$ of the total time required for the code to run for the 1D and 2D cases, respectively. As for the use of short-time future estimates, we consider DMD for predicting two weeks in the future given a set of snapshots. For the 1D case, we feed DMD with snapshots covering $30$ days, while for the 2D case, the snapshot matrix comprises $46$ days of simulation results. The DMD results are presented in terms of efficiency and accuracy compared to the projected solutions, and we observe that the predictions are mostly in good agreement except for the exposed compartment. To test the approach's agnosticism regarding the dynamics of the systems, we consider two different fluid dynamics problems: a 2D density-driven gravity current and a 3D bubble rising problem. 
We use a $3000$-snapshot simulation to generate a basis containing the spatio-temporal coherent structures for the density-driven gravity flow. The lock-exchange simulation is solved using the \texttt{FEniCS} framework. The computational effort regarding the projection, in this case, corresponds to $5.06\%$ of the overall computational time. As for the DMD analysis, we evaluate the results for different values of the rank $r$ and observe that, as we increase it, in this case, the overall relative errors decrease significantly. However, when evaluating the quantities of interest such as mass conservation and front position, even the worst case yields good approximations.  
For the 3D example, we use the \texttt{libMesh} library, testing three different reference meshes considering the three element sizes existent in the adaptive simulation. We investigate the occurrence of projecting the adaptive solution onto meshes that do not necessarily guarantee that all the scales obtained in the solution are preserved in terms of quantities of interest related to the bubble geometry and dynamics. We observe that the use of $\mathcal{L}^2$-projection on coarser meshes leads to issues regarding the shape of the bubble, affecting quantities of interest such as bubble volume and sphericity. When considering the $\mathcal{L}^2-$projection on meshes with approximately the same spatial resolution as the finest scale in the adaptive meshes, these effects are mitigated. As for the bubble dynamics, even coarser meshes reveal similar results for the bubble center of mass and rise velocity in comparison with finer mesh projections and the adaptive solution itself. In terms of computational effort, the $\mathcal{L}^2-$projection corresponds to $0.03\%$, $0.14\%$ and $1.09\%$ of the total simulation time for the coarse, intermediate, and fine meshes, respectively. Proceeding to the DMD analysis, we present the results for a $220$ time step reconstructions and a $20$ time step prediction for the bubble rising problem. We test the approximations for multiple values of $r$ being $r = 60$ our best approximations. The results computed by DMD are presented in terms of efficiency and accuracy compared to the projected solutions. We also reevaluate the quantities of interest, now with the DMD approximation. For the reconstruction, results are practically identical to those observed for the projection, while for the prediction, the center of mass and bubble volume yield good results while sphericity and rise velocity are not in agreement with those obtained in the simulation.
\par 
Summarizing our findings, we highlight that the present approach, despite its simplicity:
\begin{itemize}
    \item is versatile since every finite element library that presents AMR/C is often equipped with efficient projection algorithms;
    \item is fast in the sense that the $\mathcal{L}^2-$projection only requires solving a sparse, well-conditioned linear system $\mathbf{A}\mathbf{x} = \mathbf{b}$ where $\mathbf{A}$ is the mass matrix, obtained by a self-adjoint operator, enabling the use of efficient solvers;
    \item is framework-agnostic, as noted in the results of our simulations implemented on \verb|libMesh| and \verb|FEniCS|, requiring only the standard procedures already in place in both libraries;
    \item is problem-agnostic, as observed on meshes with different spatial dimensions (1D, 2D, and 3D), topologies (structured and unstructured), and equations (convection-dominated and reaction-diffusion systems);
    \item preserves the dynamics computed on the adaptive snapshots for all the tested examples.
\end{itemize}
\par 
The results observed in this study reveal that the use of mesh projections for adaptive solutions preserves the dynamics inherent to the finite element solutions. However, this approach on large problems could be inefficient or prohibitive since the mesh projection on a sufficiently fine target reference mesh can lead to unfeasible DMD storage and computation times. As a future work, aiming to circumvent this issue, we plan to explore the use of data compression on DMD, once used for streaming purposes \cite{Erichson2019}, in the mesh projection context.

\section*{Acknowledgements}
This research was financed in part by the Coordena\c{c}\~ao de Aperfei\c{c}oamento de Pessoal de N\'ivel Superior - Brasil (CAPES) - Finance Code 001. This research has also received funding from CNPq and FAPERJ. Computer time in Lobo Carneiro supercomputer was provided by the High Performance Computer Center at COPPE/Federal University of Rio de Janeiro, Brazil. A. Reali was partially supported  by the Italian Ministry of University and Research (MIUR) through the PRIN project XFAST-SIMS (No. 20173C478N).

\section*{Conflict of interest}
The authors declare that they have no conflict of interest.

\section*{Appendix}
\label{sec:appendix}
\par In this appendix we provide a more detailed discussion of several points raised in section 3 regarding the use of DMD on adapted meshes. As discussed in the main text, given the way DMD works by acting on the snapshot matrix, it is therefore essential for its proper application that all snapshots are of equal dimension; if the original dataset has been produced using different meshes, this may not be the case. Thus, it is  crucial in such a case to use some sort of post-processing technique to provide a uniform snapshot size across the time series. 
\par
\begin{figure}[ht!]
  \centering
 \includegraphics[width=\linewidth]{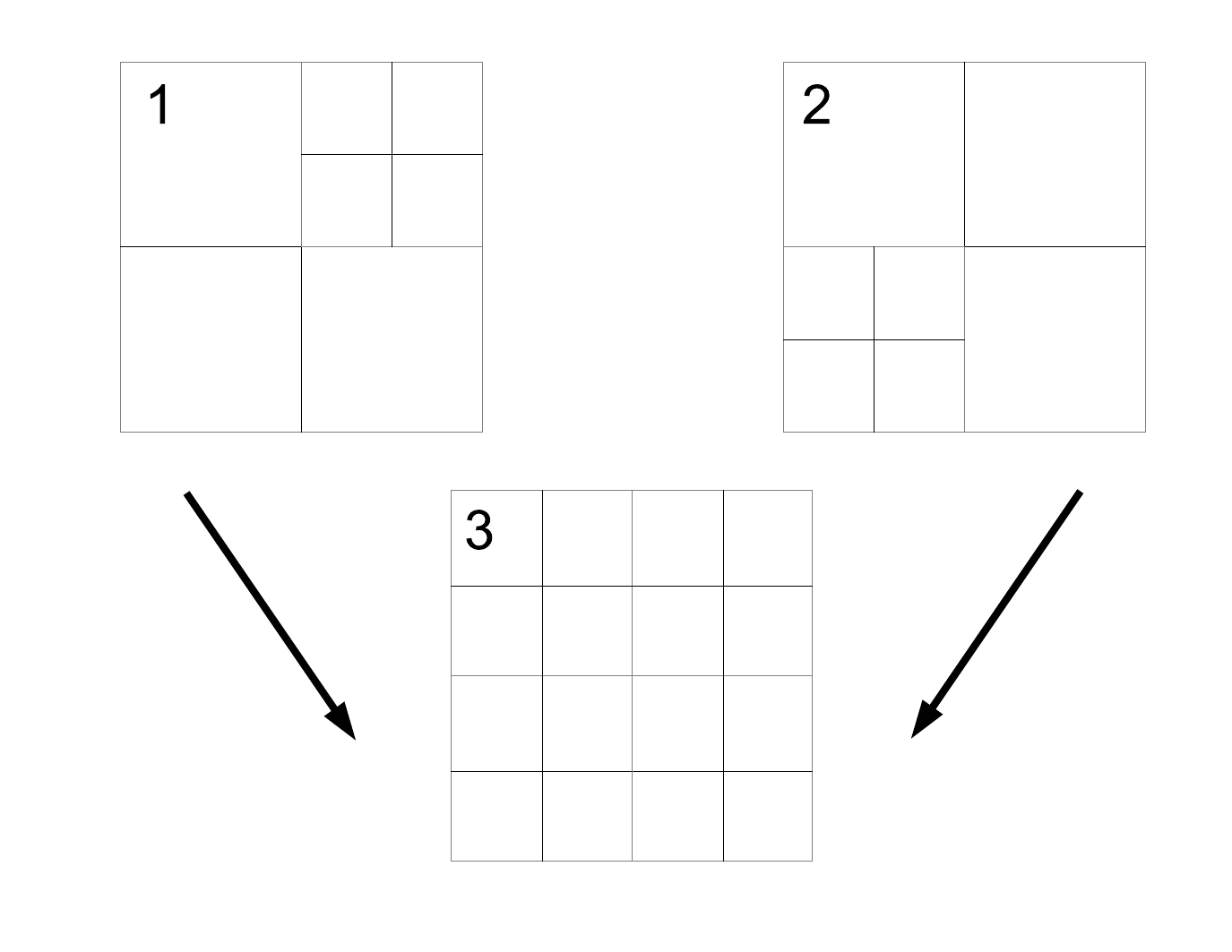}
  \caption{Depiction demonstrating the need for projection. Even though Mesh 1 and Mesh 2 have the same number of degrees of freedom, the different topologies require that, for a proper application of DMD, we must project both meshes onto a reference mesh 3 (Mesh 3). As shown, this reference mesh should be sufficiently fine to resolve all necessary in all areas of the domain.}
  \label{fig:Mesh_Projection}
\end{figure}

 For practical purposes, however, ensuring simply that all snapshots of uniform dimension is a necessary but not sufficient condition to usefully apply DMD. As illustrated in Figure \ref{fig:Mesh_Projection}, we show an example of two meshes (labelled as 1 and 2) that both an have equal number of nodes. Assuming the same order of finite element approximation is used on each mesh, the resulting vectors $\mathbf{u}_1^h$ and $\mathbf{u}_2^h$ will have the same dimension. However, while the number of degrees of freedom is the same on each mesh, they are topologically distinct. Hence, to properly perform DMD in this instance, we must have a third mesh, which we call a \textit{reference mesh} (labeled as mesh 3), on which we project \textit{both} $\mathbf{u}_1^h$ and $\mathbf{u}_2^h$. We note that this mesh is refined everywhere, compared to the locally-refined mesh 1 and mesh 2. In this way, we guarantee that the fine-scale frequencies captured by all the different mesh levels are properly represented by the DMD modes.
\par In the current work, we have performed this by using $\mathcal{L}^2$-projection of all snapshots onto a reference (or projected) mesh. Let $\mathbf{u}^h$ denote the finite element solution on the original mesh. Then we define the $\mathcal{L}^2$-projection onto the reference mesh as:
\begin{align}\label{l2Prjn}
    (\mathbf{u}^{proj},\mathbf{v}^{proj}) &= (\mathbf{u}^h,\mathbf{v}^{proj}),
\end{align}
for all $\mathbf{v}^{proj}$ in $V^{proj}$. One may quickly verify that this satisfies the \textit{Galerkin orthogonality} property on the reference mesh, i.e., for all $\mathbf{v}^{proj}$ in $V^{proj}$, the relation:
$$ (\mathbf{u}^{proj}-\mathbf{u}^h,\mathbf{v}^{proj})=0 $$
is satisfied. At the algebraic level, the problem \eqref{l2Prjn} may be expressed as:
\begin{align}\label{l2PrjnAlg}
\mathbf{M} \mathbf{u}^{proj} &= \mathbf{P} \mathbf{u}^h,    
\end{align}
where $\mathbf{M}$ is the mass matrix on the FEM space $V_h^{proj}$ and $\mathbf{P}$ is a rectangular matrix consisting of the projection from $V^h$ onto $V^{proj}$. Letting $N_i^{proj}$ correspond to the shape functions of $V^{proj}$ and $N_i^h$ to $V^h$, the matrices have the following definitions:
\begin{align}\begin{split}\label{matrixDefs}
    (\mathbf{M})_{i,j} &= \int_{\Omega} N_i^{proj} N_j^{proj} \\
    (\mathbf{P})_{i,j} &= \int_{\Omega} N_i^{proj} N_j^h.
\end{split}
\end{align}

We then may view $\mathbf{u}^{proj}$ as:
\begin{align}\label{uhRefDefin}
\mathbf{u}^{proj} &= \mathbf{M}^{-1} \mathbf{P} \mathbf{u}^h.    
\end{align}
Consider now DMD on Mesh 1:
\begin{align}\label{DMDMesh1}
\mathbf{Y}_2^h &= \mathbf{A}^1 \mathbf{Y}_1^h, \qquad \text{where } \mathbf{Y}_1^h = \left[
                \begin{array}{cccc}
                \vrule & \vrule &        & \vrule \\
                \mathbf{u}^h(t_0)  & \mathbf{u}^h(t_1)  & \ldots & \mathbf{u}^h(t_{m})    \\
                \vrule & \vrule &        & \vrule
                \end{array}
                \right]
\end{align}
Similarly, DMD defined on the reference mesh reads:
\begin{align}\begin{split}\label{DMDMesh1}
\mathbf{Y}_2^{proj} &= \mathbf{A}^{proj} \mathbf{Y}_1^{proj},\, \text{where } \mathbf{Y}_1^{proj}= \left[
                \begin{array}{cccc}
                \vrule & \vrule &        & \vrule \\
               \mathbf{M}^{-1}\mathbf{P} \mathbf{u}^h(t_0)  & \mathbf{M}^{-1}\mathbf{P} \mathbf{u}^h(t_1)  & \ldots & \mathbf{M}^{-1}\mathbf{P} \mathbf{u}^h(t_{m})    \\
                \vrule & \vrule &        & \vrule
                \end{array}
                \right] \\
\mathbf{Y}_2^{proj} &= \mathbf{A}^{proj} \mathbf{M}^{-1} \mathbf{P} \mathbf{Y}_1^h.
\end{split}\end{align}
Similarly, 
\begin{align}\label{Y2Proj}
    \mathbf{Y}_2^{proj} &= \mathbf{M}^{-1} \mathbf{P} \mathbf{Y}_2^h.
\end{align}
From \eqref{DMDMesh1}, \eqref{Y2Proj}:
\begin{align}\begin{split}\label{IdentityofProj}
    \mathbf{M}^{-1} \mathbf{P} \mathbf{Y}_2^h &= \mathbf{A}^{proj} \mathbf{M}^{-1} \mathbf{P} \mathbf{Y}_1^h, \text{ implying: } \\
    \mathbf{Y}_2^h &= \mathbf{P}^{-T} \mathbf{M} \mathbf{A}^{proj}\mathbf{M}^{-1} \mathbf{P} \mathbf{Y}_1^h,
\end{split}\end{align}
where $\mathbf{P}^{-T}$ denotes the left-sided pseudoinverse of $\mathbf{P}$ (see e.g. \cite{golub2013matrix}). Thus, rank($\mathbf{P}$) = dim($\mathbf{u}_h$), DMD may properly represent all relevant dynamics on the reference (projected) mesh.
\newline \textbf{Theorem.} \textit{If every degree of freedom of the original mesh is also a degree of freedom of the reference mesh, then the matrix $\mathbf{P}$ is full-rank.}
\newline \textbf{Proof.} By assumption, for every $N_i^h$, the corresponding $N_i^{proj}$ (assumed to be more fine and hence greater in number) is defined such that:
$$ \text{supp}(N_i^{proj}) \subset \text{supp}(N_i^h).$$
This idea may be alternatively expressed as: 
$$V^1 + V^2 + ... V^n = V^{proj},$$
where the different $V^i$ are finite element spaces corresponding to the different adapted meshes. Such a condition was also used in \cite{ullmann2016pod}. We therefore have that:
\begin{align}\begin{split}\label{integralReln}
    \int_{\Omega} N_i^h N_j^{proj} &= \kappa_{ij} \int_{\Omega} N_i^h N_j^h,
\end{split}
\end{align}
where $0 < \kappa_{ij} \leq 1 $ for all $i,j$ is a volumetric constant relating the measures of $\text{supp}(N_j^{proj})$ and $\text{supp}(N_j^h)$. By \eqref{integralReln}, the matrix $\widetilde{\mathbf{M}}$, defined such that:
$$ (\widetilde{\mathbf{M}})_{i,j} = \kappa_{ij} \int_{\Omega} N_i^h N_j^h,$$
which has the same rank (that is, dim($V^h$)) as $\mathbf{M}^h$ by construction. We then observe that, for $i,j \leq \text{dim}(V^h)$: 
$$ (\mathbf{P})_{i,j} = (\widetilde{\mathbf{M}})_{i,j}.$$ 
Thus we can express $\mathbf{P}$ as:
$$ \mathbf{P} = \begin{pmatrix} \widetilde{\mathbf{M}} \\ \mathbf{0} \end{pmatrix} +\begin{pmatrix} \mathbf{0} \\ \widetilde{\mathbf{P}} \end{pmatrix}. $$
Assume now for the sake of contradiction that $\mathbf{P}$ is not full-rank. This implies that there exists a $\mathbf{b} \in \mathbb{R}^{\text{dim}(V^h)}$ such that $\mathbf{P}\mathbf{b}=\mathbf{0}.$ Then:
\begin{align}
     \begin{pmatrix} \widetilde{\mathbf{M}}\mathbf{b} \\ \mathbf{0} \end{pmatrix} +\begin{pmatrix} \mathbf{0} \\ \widetilde{\mathbf{P}}\mathbf{b} \end{pmatrix} &=  \begin{pmatrix} \mathbf{0} \\ \mathbf{0} \end{pmatrix},
\end{align}
implying that $\widetilde{\mathbf{M}}\mathbf{b}=0$, contradicting the full-rank of $\widetilde{\mathbf{M}}$. Hence we can conclude $\mathbf{P}$ is full-rank.

\par The above theorem requires an assumption that, while often the case in practice, is not strictly necessary in our experience to obtain acceptable results using DMD. As mentioned, a similar condition was required for the related POD method in \cite{ullmann2016pod}, in which a common finite element space $V^n$ was implicitly constructed from the different spaces from each adapted mesh. Issues regarding projection between meshes were also studied in \cite{maddison2017optimal,dickopf2011study, bolten2015generalized}, in which the interested reader may find more theoretical results regarding projection between meshes.

\bibliographystyle{abbrv}  
\bibliography{references} 
\end{document}